%% file: main.tex
\pdfoutput=1
\documentclass[lettersize,journal]{IEEEtran}
\usepackage{amsmath,amsfonts}
\usepackage{lineno}
\usepackage{array}
\usepackage{tabularx}
\usepackage[caption=false,font=normalsize,labelfont=sf,textfont=sf]{subfig}
\usepackage{textcomp}
\usepackage{stfloats}
\usepackage{url}
\usepackage{verbatim}
\usepackage{graphicx}
\usepackage{hyperref}
\usepackage[affil-it]{authblk}
\usepackage{balance}

\usepackage{tikz}
\usepackage[edges]{forest} 
\definecolor{hidden-draw}{RGB}{106,142,189} 
\definecolor{hidden-blue}{RGB}{194,232,247} 
\definecolor{hidden-orange}{RGB}{217, 232, 252} 

\usepackage{caption}

\usepackage{booktabs}
\usepackage{mathabx}
\usepackage{algorithm}
\usepackage[noend]{algpseudocode}
\usepackage{multirow}
\usepackage{wrapfig}

\begin{document}
\title{Parameter-Efficient Fine-Tuning in Large Models: A Survey of Methodologies}

\author{Luping Wang\(^1\), Sheng Chen\(^1\), Linnan Jiang\(^1\), Shu Pan\(^1\), Runze Cai\(^1\), Sen Yang\(^1\) and Fei Yang*\thanks{* Corresponding author}\(^1\)\\
\(^1\)Zhejiang Laboratory\\
{\tt\small \{wangluping,scucs,jianglinnan,shu.pan,cairz,yangsen,yangf\}@zhejianglab.org}
}

\maketitle

\begin{abstract}
The large language models, as predicted by scaling law forecasts, have made groundbreaking progress in many fields, particularly in natural language generation tasks, where they have approached or even surpassed human levels. However, the unprecedented scale of their parameters brings significant computational and storage costs. These large language models require substantial computational resources and GPU memory to operate. When adapting large language models to specific downstream tasks, their massive parameter scale poses a significant challenge in fine-tuning on hardware platforms with limited computational power and GPU memory. To address this issue, Parameter-Efficient Fine-Tuning (PEFT) offers a practical solution by efficiently adjusting the parameters of large pre-trained models to suit various downstream tasks. Specifically, PEFT adjusts the parameters of pre-trained large language models to adapt to specific tasks or domains, minimizing the introduction of additional parameters and the computational resources required. This review mainly introduces the preliminary knowledge of PEFT, the core ideas and principles of various PEFT algorithms, the applications of PEFT, and potential future research directions. By reading this review, we believe that interested parties can quickly grasp the PEFT methodology, thereby accelerating its development and innovation.
\end{abstract}

\section{Introduction}
\label{Introduction}
\input{Introduction}

\section{Preliminary}
\label{Preliminary}
\input{Preliminary}

\section{PEFT Taxonomy}
\label{PEFT_Taxonomy}
\input{PEFT_Taxonomy}

\section{Applications of PEFT}
\label{Applications_of_PEFT}
\input{Applications_of_PEFT}

\section{Future Directions}
\label{Future_Directions}
\input{Future_Directions}

\section{Conclusions}
\label{Conclusions}
\input{Conclusions}

\section{Acknowledgements}{This work was supported by the National Key Research and Development Program of China (Grant No. 2023YFE0108600), National Natural Science Foundation of China (Grant No. U22A6001), Shanghai Artificial Intelligence Laboratory (Grant No. P22KN00581) and "Pioneer" and "Leading Goose" Research and Development Program of Zhejiang (Grant No. 2024SSYS0002).}

\bibliographystyle{IEEEtran}
\bibliography{reference.bib}

\end{document}

%% file: Introduction.tex
In recent years, large pre-trained models, commonly referred to as ``large language models", have emerged as a significant advancement in the field of artificial intelligence. Due to their outstanding performance and versatility in various application contexts, these models have attracted plenty of attention and provoked much discussion. These models have impressive computing capabilities and extensive data resources, allowing them to excel in tackling intricate jobs. Within the field of natural language processing (NLP), notable interest is given to Large Language Models (LLMs). These models demonstrate remarkable ingenuity in text generation \cite{wu2023next, li2024pre}, machine translation \cite{zhu2023multilingual,wang2023document}, personalized chatbots \cite{zheng2023judging,kim2023chatgpt,dan2023educhat}, text summarization \cite{zhang2019pretraining}, sentiment analysis \cite{zhang2023enhancing}, and question-answering systems \cite{pan2024conv}. 

Nevertheless, the development of large language models faces significant challenges and controversies. These models require substantial computational resources and data support, which can potentially jeopardize the environment and compromise privacy protection \cite{yao2024survey}. Despite their impressive performance in specific tasks, these models still have limitations and error rates that need continuous optimization and improvement \cite{huang2024survey,huang2022towards,saparov2022language}. When directly using large language models for specific tasks, their performance often falls below desired levels. Consequently, fine-tuning large language models has become a crucial method for enhancing model performance.

Parameter-Efficient Fine-Tuning (PEFT) is a transfer learning method specifically developed to adapt the parameters of the large pre-trained models to suit new tasks and scenarios. This approach involves dynamically adjusting the model to enhance its effectiveness in performing certain tasks, taking into account the distinct features and requirements of the target task. The fine-tuning process typically entails improving the model architecture \cite{houlsby2019parameter}, optimizing parameters \cite{li2021prefix,lester2021power}, and adapting learning strategies \cite{chen2020recall}, among other considerations, to achieve better performance in new tasks. As the field of deep learning continues to evolve, techniques for optimizing and fine-tuning large language models have also made significant advancements. Notable PEFT approaches include LoRA \cite{hu2021lora}, adapter tuning \cite{zhang2023adalora}, prefix tuning \cite{li2021prefix}, prompt tuning \cite{lester2021power}, P tuning \cite{liu-etal-2022-p}, BitFit \cite{zaken2021bitfit}, and others. However, despite the significant achievements of large model fine-tuning techniques across several fields, there are always challenges and difficulties that need to be resolved. Overfitting mitigation, optimizing fine-tuning efficiency, and striking a learning balance between pre-training and fine-tuning tasks are a few examples of issues that need more investigation.

In recent years, hundreds of articles on PEFT have been published, with some studies offering informative overviews of the most prevalent approaches. A comparative analysis of these surveys in terms of taxonomy and application is shown in Table \ref{tab:survey_comp}.

\begin{table*}[htbp]
\centering
\setlength{\tabcolsep}{4pt}
\renewcommand{\arraystretch}{1.2}
\begin{tabularx}{\textwidth}{@{}>{\centering\arraybackslash}X *{9}{>{\centering\arraybackslash}X}@{}}
\toprule
\multirow{2}{*}{Survey} & \multicolumn{5}{c}{Taxonomy} &  \multicolumn{4}{c}{Application}\\
\cmidrule(lr){2-6}\cmidrule(lr){7-10}
 & Add. & Sel. & Rep. & Hybrid & Unified & NLP & Vision & Multi. & Diffusion \\
\midrule
\cite{ding2022delta} & \checkmark & \checkmark & \checkmark & & &\checkmark & & &\\
\cite{lialin2023scaling} & \checkmark & \checkmark &\checkmark & \checkmark & & & & &\\
\cite{xu2023parameter} & \checkmark & \checkmark &\checkmark &\checkmark &\checkmark &\checkmark & &\checkmark &\\
\cite{xin2024parameter} & & \checkmark &\checkmark & &\checkmark & &\checkmark & &\\
\cite{han2024parameter} & \checkmark & \checkmark &\checkmark &\checkmark &\checkmark &\checkmark & & & \checkmark\\
Ours & \checkmark & \checkmark &\checkmark &\checkmark & &\checkmark & \checkmark &\checkmark &\checkmark\\
\bottomrule
\end{tabularx}
\caption{A Comparative Analysis of Survey Methodologies: Taxonomy and Application Domains, with Abbreviations: Additive (Add.), Selective (Sel.), Reparameterized (Rep.), Multi-task (Multi.) and Diffusion Model (Diffusion)}
\label{tab:survey_comp}
\end{table*}

\cite{ding2022delta} introduce a theoretical abstraction for Delta Tuning, which is analyzed from the viewpoints of optimization and optimum control. This abstraction offers a unified approach to describe the current parameter-efficient fine-tuning methods which provides a distinct perspective for future investigations. Nonetheless, while the study predominantly concentrates on NLP applications, the generalizability and efficacy of these methods in diverse domains merit additional investigation.
\cite{lialin2023scaling} provide a comprehensive analysis and classification that covers a broad range of methods and compares approximately 30 approaches across five dimensions: storage efficiency, memory efficiency, computational efficiency, accuracy, and inference overhead. However, while the article primarily focuses on detailed methods with practical efficiency for fine-tuning multibillion-scale language models, the exploration of real-world application scenarios is relatively limited. \cite{xu2023parameter} provide a thorough evaluation and analysis of current PEFT approaches, assessing their performance, parameter efficiency, and memory utilization within a range of NLP tasks. Nonetheless, the paper does not fully expound on the practical applications of these methodologies in actual operational environments, nor does it deeply investigate their adaptability and the domain-specific challenges they might encounter. \cite{xin2024parameter} offer a comprehensive overview and future directions for visual PEFT, with a systematic review of the latest advancements. While the article spans multiple visual tasks, the experiments are primarily focused on several common tasks and do not fully encompass the broader range of potential application scenarios.  \cite{han2024parameter} provide a detailed classification of PEFT approaches and explores the application of PEFT techniques across various model architectures and downstream tasks, as well as the systematic design challenges of parameter-efficient fine-tuning methods. It offers researchers and engineers a comprehensive overview of PEFT approaches, but there is still room for improvement in terms of practical application coverage.

Our contributions are as follows:
\begin{itemize}
    \item This survey comprehensively reviews the latest literature PEFT, covering cutting-edge methods and related research. It establishes a theoretical framework and offers a solid knowledge base for future research.
    \item We make extensive use of intuitive schematic diagrams and structured tables to elaborate on PEFT methodologies. By means of visualization, we demonstrate the complex principles of these methods, carry out comparative analyses of different approaches, and organically combine intuitiveness with systematicness, which significantly enhances the readability and academic value of the research content.
    \item Breaking traditional boundaries, this survey explores PEFT in natural language processing, computer vision, multimodal fusion, and diffusion models. It uncovers application potential, offers practical guidelines, and broadens the application scope of fine-tuning technology.
\end{itemize}

This survey aims to comprehensively review the recent advancements in large model fine-tuning techniques. By conducting a thorough examination of existing research, our objective is to identify and fill the gaps in our current knowledge system. This will result in the development of a comprehensive and systematic framework of knowledge, which will provide researchers with a concise perspective on the topic and guide their future research. In conclusion, our work offers valuable resources and perspectives that can be utilized for both academic and practical purposes in related domains.
The remainer of this survey is structured in the following manner:

In Section \ref{Preliminary}, we offer a succinct summary of the fundamental components of large language models, including their past development, emerging capabilities, and the scaling laws that govern their size. Subsequently, we offer a brief overview of the dominant classifications of comprehensive language models and introduce the fundamental principles and framework of multi-modal comprehensive models. Furthermore, we investigate the primary methodologies employed in the fine-tuning domain of extensive language models, including instruction fine-tuning, alignment, and Reinforcement Learning from Human Feedback (RLHF). Ultimately, we present a brief summary of the most used benchmarks and assessment datasets in the field of big model fine-tuning.

In Section \ref{PEFT_Taxonomy}, we offer a comprehensive analysis and summary of PEFT approaches, presenting a cohesive framework for classifying current PEFT methodologies, encompassing over 100 research articles published from June 2019 to July 2024. Expanding on the conventional tripartite classification of additive, reparameterized, and subtractive PEFT, we incorporate summaries of hybrid, quantization, and multi-task categorization PEFT approaches.

In Section \ref{Applications_of_PEFT}, we present a comprehensive analysis and description of the prevailing PEFT approaches in the fields of multimodal, visual, and diffusion models. Our objective is to provide a deep understanding and recommendations for choosing and improving PEFT in different application scenarios.

In Section \ref{Future_Directions}, we encapsulate our extensive survey and put forward multiple promising avenues for future advancements, encompassing both algorithmic refinements and task scenarios, hoping to provide valuable insights for further research and development in this burgeoning field.

%% file: Preliminary.tex
\subsection{Large Language Models: Foundations and Variants}
\subsubsection{Large Language Models}
\label{sub: llm}
\input{Large_Language_Models}

\subsubsection{Prevalent LLMs}
\label{sub: prevalent llm}
\input{Prevalent_LLMs}

\subsubsection{Multimodal Large Language Models} 
\label{sub: MLLM}
\input{Multimodal_Large_Language_Models}

\subsection{Optimization , Datasets, and Evaluation of Large Language Models}
\subsubsection{Instruction Tuning}
\label{sub: instruction tuning}
\input{Instruction_Tuning}

\subsubsection{Alignment Tuning and RLHF}
\label{sub: alignment tuning and rlhf}
\input{Alignment_Tuning_and_RLHF}

\subsubsection{Datasets for LLM}
\label{sub: datasets}
\input{Datasets_for_LLM}

\subsubsection{LLM evaluation}
\label{sub: evaluation}
\input{LLM_evaluation}

%% file: Large_Language_Models.tex
\paragraph{Background}
\input{Background}

\paragraph{Emergent abilities}
\input{Emergent_abilities}

\paragraph{Scaling Laws of LLMs} 
\input{Scaling_Laws_of_LLMs}

%% file: Background.tex
LLMs refer to neural language models with a large number of parameters, typically over billions of parameters. These models are built on the transformer architecture 
\cite{vaswani2017attention} and are pre-trained on vast text
corpora~\cite{devlin2018bert}. Prior to the emergence of LLMs, the advent of transformers revolutionized the development approach for neural language models, shifting from end-to-end training to a pre-train then fine-tune paradigm. Under the pre-train fine-tune paradigm, pre-trained models can be repeatedly utilized, significantly enhancing the scalability of neural language models. Consequently, the scale of parameters is continuously growing larger. 
For instance, OpenAI's GPT-1 possessed 120 million parameters, while GPT-2 boasted
1.5 billion parameters. This number surged to 175 billion for GPT-3 and soared to 1.76 trillion for the latest GPT-4~\cite{achiam2023gpt}.

%% file: Emergent_abilities.tex
Research suggests that the rapid expansion of the parameter scale may lead to
emergent abilities \cite{wei2022emergent}, which are
formally defined as abilities that are not present in small models but arise in large language models, constituting one of the most prominent characteristics distinguishing LLM from previous PLM. In conclusion, emerging abilities can be categorized into threefolds.

\textbf{In-context learning}. In-context learning 
\cite{wei2022emergent}\cite{sanh2021multitask}, known as ICL defined in GPT-3 
\cite{brown2020language}, illustrates the ability of LLMs to acquire new task
capabilities based on a small set of examples in context. 
Importantly, this process does not require additional training or gradient updates,
indicating that the LLM is capable of completing new tasks with only prompts.
In addition, \cite{wei2022emergent}
reveals that ICL is associated with both the LLM and the downstream task.

\textbf{Instruction following}. 
Natural language descriptions, known as instructions, are essential for
fine-tuning LLMs. Instruction tuning organizes fine-tuning datasets in
the format of natural language descriptions (instructions). Research 
\cite{ouyang2022training} shows that with instruction tuning, LLMs are enabled
to follow task instructions for new tasks without using explicit examples, 
demonstrating better generalization capability across inputs of various tasks. \cite{chung2024scaling} discovered that to achieve evident efficacy,
instruction tuning should be conducted on a relatively large-scale LLM, e.g.,
over 60B parameters. 

\textbf{Step-by-step reasoning}.
Constrained by parameter size, PLMs often struggle to solve tasks requiring
intricate reasoning. In contrast, scaling up in parameter size equips language
models with the Chain-of-Thought (CoT) \cite{wei2022emergent}. CoT enhances language models' performance on tasks involving logic, calculation, and decision making by structuring the input
prompt to human reasoning. Thanks to CoT, LLMs are enabled to tackle tasks that
demand intermediate reasoning steps to derive the final answer, akin to
constructing a step-by-step prompt that invokes a thinking and inference process
within the model. 

Emergent abilities in large language models (LLMs) have significantly boosted various real-world applications, across fields such as natural language \cite{kalla2023study,team2023gemini, liu2024deepseekv3}, healthcare \cite{wang2023power, biswas2023role}, legal \cite{li2024legalagentbench}, financial \cite{xing2024designing} and multiple scientific disciplines \cite{ahn2024large, wang2023scibench}. Despite the promising emergent capabilities, there are three main limitations that restrict the further and deeper applications of LLMs.
Firstly, the inconsistency across models and tasks. LLMs trained on different architectures or datasets may demonstrate emergent behavior to varying degrees. Some models might excel in certain tasks while failing to exhibit the same level of ability in others, resulting in unpredictable performance when applied to diverse real-world scenarios \cite{bommasani2021opportunities}.
Secondly, the hallucinations and factual errors. 
LLMs often generate text that is fluent and coherent. However, they can also produce hallucinations, outputs that seem plausible but contain factual inaccuracies or misleading information 
\cite{bender2021dangers, lin2021truthfulqa}. This tendency is particularly problematic in contexts where precise and reliable information is crucial, such as legal, medical, or scientific applications.
Finally, the deficiency in deep understanding. The performance of LLMs largely stems from recognizing statistical patterns in vast datasets rather than a genuine semantic understanding of the content \cite{bender2021dangers}.This superficial grasp of language limits their effectiveness in tasks requiring in-depth logical reasoning and nuanced comprehension across models and tasks.

In conclusion, emergent abilities grant LLMs remarkable problem-solving capabilities, though they remain imperfect. To bridge the gap between LLMs and real-world applications, integrating traditional algorithms, expert systems, or hybrid models may be necessary to enhance reliability, accuracy, and domain-specific expertise.

%% file: Scaling_Laws_of_LLMs.tex
Thanks to the exceptional scalability of the transformer architecture \cite{vaswani2017attention}, language models also exhibit high scalability. The scaling laws for LLM describe how the model grows and performs as the volume of training data increases.

In general, a scaling law includes four parameters, which also characterize a language model: 
(1) Parameters count $N$. The number of parameters of an LLM is often associated
with the number of transformer layers and the hidden size, except for some MoE LLMs.
(2) Data size $D$. In LLM, this refers to the number of tokens for training. 
(3) Computation cost $C$. This is typically measured in terms of time
and computational resources.  (4) Loss $L$. The performance of training is usually
evaluated by the training loss. There are two representative scaling laws for transformer LLMs.

\noindent\textbf{The Kaplan scaling law} Proposed by Kaplan 
\cite{kaplan2020scaling}, the law examines the statistical relations between the
parameters $C, N, D$ and $L$ over a wide range of values, models and data tokens. The relationships can be expressed through the following equations:

\begin{align}
  L(N) &= \left(\frac{N_c}{N}\right)^{\alpha_N}, \alpha_{N} \sim 0.076, N_c \sim 8.8 \times 10^{13} \\
  L(D) &= \left(\frac{D_c}{D}\right)^{\alpha_D}, \alpha_{D} \sim 0.095, D_c \sim 5.4 \times 10^{13} \\
  L(C) &= \left(\frac{C_c}{C}\right)^{\alpha_C}, \alpha_{C} \sim 0.050, N_c \sim 3.1 \times 10^{8} 
  \enskip,
\end{align}

where the loss $L$ is influenced by 
parameters $N$, $D$, and $C$, shedding light on decision-making processes when
computational resources are limited. 

\noindent\textbf{The Chinchilla scaling law} Proposed by DeepMind
\cite{hoffmann2022training}, the law provides guidelines for compute-optimal
training of LLMs, specifically when computational resources are limited. Through
rigorous experiments spanning a wide range of model sizes from 70M to 16B and
dataset sizes from 5B to 500B tokens, they derived a scaling law with different
coefficients compared to Kaplan's, as shown below:

\begin{equation}
  \label{eq: chin}
  L(N, D) = E + \frac{A}{N^\alpha} + \frac{B}{D^\beta}
  \enskip,
\end{equation}
where $E$ denotes the loss of an ideal generative process on the test
data. Furthermore, claimed by the research, the constants in this formula
are $\alpha =0.34,\beta =0.28, A=406.4, B=410.7,L_{0}=1.69$. Moreover, there is a
general constraint that model the relationship between $C$ and $(N, D)$: $C = 6 N D$, which means that it costs six FLOPs per parameter to train one token. Thus, the optimal selection of model size and data size can be determined and expressed as: 

\begin{align}
  N_{opt} &= 0.6\,C^{0.45}\\
  D_{opt} &= 0.3\,C^{0.55}\\
  L_{opt} &= 1070\,C^{-0.154} + 1.7
  \enskip.
\end{align}

From the equations, scaling laws can guide decisions regarding model size. Given a fixed compute budget (e.g., 100K GPU hours), they enable predictions on whether a smaller model trained for a longer duration or a larger model trained for a shorter time would yield better performance. Additionally, scaling laws provide insight into the benefits of continued training. The diminishing returns they imply suggest that beyond a certain point, increasing compute resources may not lead to a substantial enough performance gain to justify the additional cost.

In addition, based on the statistical modeling illustrated by equation \ref{eq: chin}, one approximate estimation for Chinchilla efficient model size and training dataset size can be denoted as:

\begin{align}
  N_{opt} &= 0.1\,C^{0.5}\\
  D_{opt} &= 1.7\,C^{0.5}
  \enskip.
\end{align}

This suggests that the model size and training data volume should be scaled in accordance with the available computational budget. The expected ratio of training tokens to model parameters is approximately 17:1. However, in real-world applications, this ratio is often slightly higher, as additional training data beyond the 17× scaling rule can still contribute to performance improvements when sufficient computational resources are available. For instance, GPT-2 was trained on 40B tokens with 1.5B parameters, LLaMA was trained on 1.4T tokens with 65B parameters, and DeepSeek-V3 was trained on 14.8T tokens with 0.671T parameters. While all these ratios exceed 17, they remain close to this scaling guideline.

\textbf{PEFT and Sustainability of AI Research}
Training large models from scratch is highly energy-intensive. For example, training LLaMA-3.1 405B can demand 40 million GPU hours on H100, resulting in a substantial carbon footprint. While fully Supervised Fine-Tuning (SFT) can enhance an existing LLM using a relatively smaller set of training samples, it still requires updating the entire parameter network.
In contrast, Parameter-Efficient Fine-Tuning (PEFT) methods—such as adapters or low-rank adaptations—enable fine-tuning a large pre-trained model for specific tasks by updating only a small subset of parameters (typically just 1–2\% of the total). As a result, PEFT significantly reduces computational costs; for instance, a full SFT process that requires 4 million GPU hours can be reduced to 400K GPU hours or less with PEFT.

By lowering GPU usage, PEFT not only decreases energy consumption but also mitigates the environmental impact. Moreover, this reduction in compute requirements is crucial for sustainable AI research, as PEFT provides a cost-effective and efficient approach for the AI community and researchers to conduct experiments and develop new models.

%% file: Prevalent_LLMs.tex
\paragraph{The GPT Family}

Generative Pre-trained Transformers (GPT) constitute a series of decoder-only
Transformer-based language models, pioneered by OpenAI. This family encompasses
GPT-1~\cite{radford2018improving}, GPT-2~\cite{radford2019language}, GPT-3,
InstrucGPT~\cite{ouyang2022training}, ChatGPT, GPT-4, GPT-4o, CODEX~\cite{chen2021evaluating}, and WebGPT~\cite{nakano2021webgpt}. 
GPT-1 and GPT-2 belong to PLMs, while following GPT-3, all subsequent
models in this family are classified as LLMs.

GPT-3~\cite{brown2020language} is widely recognized as the first LLM due to its significantly larger size compared to previous
PLMs, showcasing emergent abilities not observed in smaller PLMs before. A key emergent ability demonstrated by GPT-3 is in-context learning~\cite{kojima2022large}, enabling the model to solve
various downstream tasks without the need for fine-tuning. Distinct with other GPT-family LLMs, GPT-4 and GPT-4o are both multi-modal LLMs. GPT-4~\cite{achiam2023gpt} is one of the most powerful LLM reported to train on a transformer network of 1.8 trillion parameters which exhibits great capabilities in image understanding and reasoning. GPT-4o, while inheriting the powerful intelligence of GPT-4, has further enhanced its capabilities in text, image, and speech processing. Compared to existing models, it particularly excels in visual and audio comprehension.

\paragraph{The LLaMA Family} LLaMA stands as a series of open-source LLMs developed by Meta. To date, the official release includes: LLaMA, LLaMA-2, and LLaMA-3.x, spanning parameter scales from 1 billion to
405 billion. Beyond the weights provided by Meta, the qualities of these LLMs are
further extended through supervised fine-tuning and parameter-efficient
fine-tuning. 

LLaMA-1 \cite{touvron2023llama} was released in February 2023. Although LLaMA is open-sourced and possesses fewer parameters, LLaMA-13B demonstrates significant improvements over GPT-3 (175
billion parameters) across various benchmarks. As a consequence, LLaMA has
emerged as a widely adopted and exemplary base model for large language model
research. LLaMA-2 \cite{touvron2023llama} was developed in partnership with Microsoft and released half a year later.
The model maintains the same architecture as the LLaMA-1 but is
trained with 40\% more data. LLaMA-3 was released by Meta in April 2024, offering two parameter
sizes: 8B and 70B. These models underwent pre-training on approximately 15
trillion tokens of text sourced from publicly available data and are fine-tuned over 10 million human-annotated examples. 
Subsequently, Meta released LLaMA-3.1 \cite{vavekanand2024llama}, a 405B open-sourced LLM, which focuses on improving text generation capabilities and achieves performance comparable to leading models like GPT-4. Then, in September 2024, LLaMA-3.2 was released, introducing both vision models (11B and 90B) and lightweight text-only models (1B and 3B) for mobile device use. LLaMA-3.2 marked Meta's first open-source AI model capable of processing both images and text, broadening the scope of potential applications. The smaller models were designed for efficient performance on mobile devices, promoting wider adoption in edge computing scenarios.

\paragraph{The OpenAI o1 Family}
In September 2024, a new series of large language model, OpenAI-o1\footnotemark[1] \cite{openai2024openaio1card}, excels in complex reasoning tasks, using Chain-of-Thought (CoT) reasoning to outperform GPT-4o in areas like math, coding, and science. The release includes two versions: o1-preview and o1-mini. The o1-preview is an early iteration of the full model, while the o1-mini is a lightweight version optimized for size and speed. When solving problems, o1 uses the CoT\footnotemark[2] strategy like human deep thinking. Reinforcement learning helps o1 refine its thinking and strategies, find and correct errors, break down complex steps, and change approaches when necessary, improving reasoning. The reward model combines text and number scores for evaluation. 

Then previewed in December 2024, OpenAI o3-mini\footnotemark[3], the newest, most cost-efficient model was offically released in January 2025, which provides a specialized alternative for technical domains requiring precision and speedwhich. It delivers exceptional STEM capabilities—with particular strength in science, math, and coding—all while maintaining the low cost and reduced latency of OpenAI o1-mini.

\footnotetext[1]{\url{https://openai.com/index/introducing-openai-o1-preview/}}
\footnotetext[2]{\url{https://openai.com/index/learning-to-reason-with-llms/}}
\footnotetext[3]{\url{https://openai.com/index/openai-o3-mini/}}

\paragraph{The DeepSeek Family}
DeepSeek-LLM is a newly established LLM series that has garnered significant attention from both academia and industry. Developed by the company DeepSeek, the first version, DeepSeek-V1 \cite{bi2024deepseek}, was trained on 2 trillion tokens and released in January 2024, featuring two core models: 7B and 67B, along with their respective chat variants.
In the same month, DeepSeek introduced DeepSeek-MoE (Mixture of Experts) \cite{dai2024deepseekmoe} 16B, which delivers performance comparable to LLaMA 2 7B while requiring only 40\% of the computational cost. This model introduces an innovative Mixture of Experts (MoE) architecture, integrating shared expert isolation with fine-grained expert segmentation. Additionally, it incorporates a novel load-balancing strategy that optimizes both expert and device balance, enhancing computational efficiency. They made significant progress with DeepSeek-V2 \cite{liu2024deepseekv2}, a large MoE-LLM trained on 8.1 trillion tokens, featuring 2 shared experts, 160 routed experts, and 236 billion parameters. This version introduced Multi-head Latent Attention (MLA), which significantly reduces GPU memory consumption while maintaining the same level of precision. It outperforms the widely used Grouped-Query Attention (GQA) strategy adopted by LLaMA 3. Subsequently, they released DeepSeek-V3 \cite{liu2024deepseekv3} in December 2024. Building upon V2, the V3 model introduces its Multi-Token Prediction (MTP) approach and an Auxiliary-Free Load Balancing strategy to further enhance efficiency. Additionally, it integrates DualPipe \cite{li2021chimera}, cross-node all-to-all communication techniques, and a minimal-overhead memory-saving strategy, achieving a groundbreaking industrial milestone—training a 671B-parameter MoE-LLM with FP8 precision. The performance of the DeepSeek-V3 model is remarkable, achieving state-of-the-art (SOTA) results among all open-source LLMs and demonstrating performance comparable to GPT-4o and Claude 3.5 Sonnet. Moreover, it offers significant advantages in training and inference costs, requiring less than 10\% of the training cost of LLaMA 3-405B and only 9\% of the inference cost of Claude 3.5 Sonnet, revolutionizing the development of industrial LLMs. Then, DeepSeek released R1 \cite{guo2025deepseek}, a reinforcement learning-focused model leveraging the Group Relative Policy Optimization (GRPO) \cite{shao2024deepseekmath} algorithm. R1 delivers performance comparable to OpenAI-o1 in mathematical and logical reasoning tasks, while requiring only 2\% of the computational cost, marking a major breakthrough in efficiency and scalability.

\paragraph{The Claude Family}
Claude \cite{claude} represents a series of conversational AI models developed by Anthropic, designed with a focus on safety, helpfulness, and natural language understanding. This family includes Claude 1, Claude 2, Claude 2.1, Claude 3 Opus, Claude 3 Sonnet, and Claude 3 Haiku.

Claude 1 marked the initial release of Anthropic's conversational AI, introducing the concept of Constitutional AI to the field. Claude 2 and its subsequent update, Claude 2.1, brought significant improvements in language understanding, context retention, and response coherence. These versions demonstrated enhanced capabilities in handling complex queries and maintaining longer, more contextually rich conversations.

Claude 3 models (Opus, Sonnet, and Haiku) represent the latest advancements in the Claude family, each tailored for distinct applications. Opus, the most advanced model, integrates cutting-edge multimodal capabilities, enabling it to process both textual and visual inputs with deep reasoning and high-level comprehension, excelling in complex problem-solving tasks. Sonnet, optimized for efficiency and speed, is ideal for scenarios requiring rapid, precise, and contextually appropriate replies.  Haiku prioritizes simplicity and elegance, delivering concise, poetic, and highly relevant responses, making it particularly well-suited for creative and literary applications. Together, these models set new benchmarks for AI-driven interaction and analytical reasoning.

Each model in the Claude family is continuously refined to improve performance, safety, and alignment with user needs, ensuring that they remain at the forefront of conversational AI technology.

\paragraph{The Gemini Family}
Gemini \cite{anil2023gemini} constitutes a series of multimodal Transformer-based language models, developed by Google DeepMind. This family includes Gemini 1, Gemini 1.5, and Gemini 2, each introducing significant advancements in multi-modal understanding, long-context reasoning, and integration with Google's ecosystem. Unlike GPT family models, which initially focused on text generation, Gemini models were designed from the ground up to be native multimodal models, enabling seamless processing of text, images, audio, and video. Gemini 1 marked Google’s transition from its Bard chatbot to a more advanced multimodal LLM, introducing cross-modal reasoning and excelling in mathematical problem-solving, coding, and knowledge retrieval, though it faced limitations in real-world usability. Gemini 1.5 introduced a 1 million-token context window, significantly improving long-document processing, dialogue coherence, and complex multi-step reasoning. Additionally, it implemented memory capabilities, allowing it to retain user-specific context across interactions. The latest version, Gemini 2 further enhanced reasoning, tool integration, and inference speed, introducing a ``Flash Thinking" mode that enables intermediate reasoning steps for improved transparency. It also deepened integration with Google Search, Docs, and other productivity tools, optimizing it for real-world applications.

\paragraph{Other Representative LLMs} 

Mistral Series
\cite{jiang2023mistral} is an open-sourced LLM developed by Mistral AI. The basic Mistral-7B
demonstrates superior performance across all evaluated benchmarks, surpassing
all open-sourced 13B LLMs and even outperforming LLaMA-34B in reasoning,
mathematics, and code generation tasks. Mistral 7B employs Grouped Query
Attention (GQA) to enable faster inference and Sliding Window Attention (SWA) to handle longer text sequences efficiently.
Subsequently, Mistral AI introduced two additional models: Mixtral 8×7B and
Mixtral 8×22B. These models utilize the Sparse Mixture of Experts (SMoE)
technique~\cite{riquelme2021scaling}, which selectively activates a subset of experts for each input, thereby significantly reducing computational load. 

The PaLM \cite{chowdhery2023palm} (Pathwaysutilized Language Models) is
developed by Google as a collection of decoder-only LLMs. The first PaLM model
was trained on a high-quality text corpus of 780 billion tokens, boasting a remarkable 540 billion parameters. Unlike
prevalent LLMs which primarily utilize GPUs for training, PaLM is pre-trained
with the Pathways system on 6144 TPU v4 chips to facilitate rapid and efficient
training. In the following days, U-PaLM~\cite{tay2022transcending}, FlAN-PaLM~\cite{chung2024scaling} and PaLM-2 were released.

%% file: Multimodal_Large_Language_Models.tex
\paragraph{MLLM: Background}

Multimodal Large Language Model (MLLM), is an extension of LLM which adopts multimodal information as input such as text, sound, video, etc. to enable multiple dimensional reasoning and text generation.

Before the emergence of MLLM, significant
research efforts were dedicated to multi-modality. These efforts can generally
be categorized into representative and generative paradigms. An exemplary work
in the representative paradigm is CLIP~\cite{radford2021learning}, which serves
as a foundational contribution. 

This process yields a visual encoder ~\cite{cherti2023reproducible}\cite{sun2023eva} and a text encoder, effectively establishing a bridge for downstream multimodal tasks. 
In contrast, generative
frameworks~\cite{wang2022ofa}\cite{cho2021unifying} approach multimodal tasks by
transforming them into sequence-to-sequence tasks. MLLM distinguishes itself
from previous multimodal research in two key aspects. (1) Composition: MLLM is
comprised of at least one LLM with billion-scale
parameters. 
(2) Training techniques: MLLM introduces and incorporates novel training
techniques derived from LLM to enhance multimodal performance. 

\begin{figure}[htbp]
    \centering
    \includegraphics[width=0.9\columnwidth]{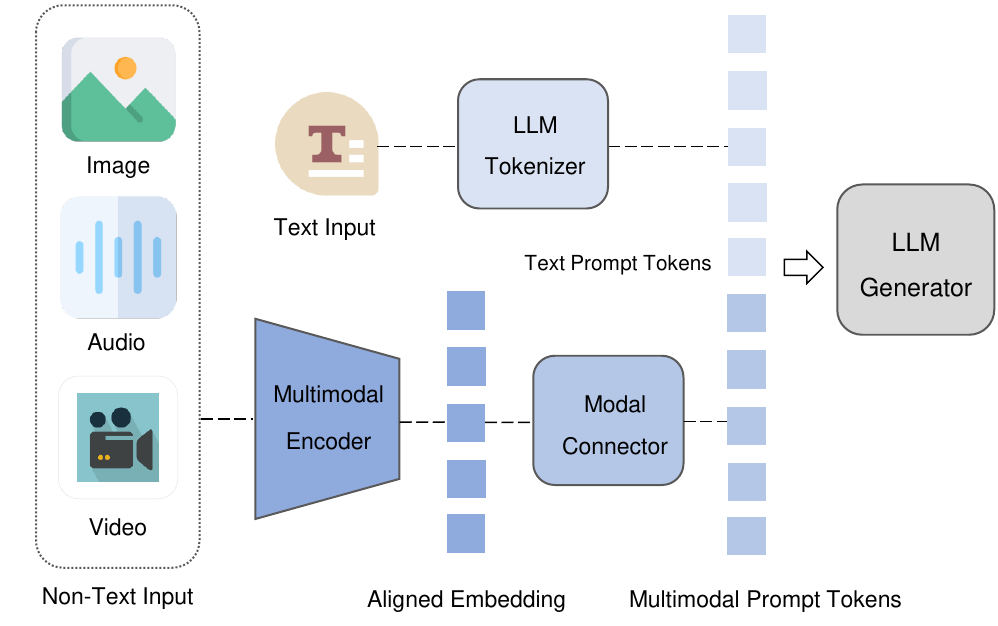} 
    \caption{Architecture of MLLM: This figure shows a common architecture and workflow of an MLLM.} 
    \label{fig: multimodal} 
\end{figure}

\paragraph{MLLM: Architecture} 

Figure \ref{fig: multimodal} illustrates the mainstream architecture of
multimodal large language models, typically composed of three modules: a
multimodal encoder, an LLM, and a modal connector. 

\textbf{Multimodal Encoder}. This module incorporates
non-text inputs, such as images or audio, and encoding the raw information into
a more compact representation. It is noteworthy that the encoder is aligned
with one or several encoders in advance to ensure associated meanings are
preserved. It is more advisable to directly adopt and fine-tune a pre-trained multimodal encoder, such as
CLIP~\cite{radford2021learning}, EVA-CLIP~\cite{sun2023eva}, or ViT-G~\cite{zhai2022scaling}, rather than starting from scratch to train a new encoder for generalized data.

\textbf{LLM}. It is also more efficient to adopt a pre-trained LLM instead of training from the start. Through tremendous pre-training on web corpus, LLMs have been embedded with rich world knowledge, and demonstrate strong generalization and reasoning capabilities. 

\textbf{Modal Connector}. This module serves as a crucial bridge between
different modalities, allowing efficient communication with the LLM. It
accomplishes this by projecting information into a space that the LLM can
readily comprehend. Through training the connector, the encoded multimodal tokens can be transformed to LLM prompt tokens that illustrate the content presented by the image, video, etc.
Consequently, the LLM will generate the expected content based on the request and prompt.

%% file: Instruction_Tuning.tex
Instruction tuning in large language models has undergone significant development, evolving from initial efforts in multi-task fine-tuning without explicit instruction prompts to sophisticated techniques leveraging diverse tasks and templates.
Early work focused on improving downstream task performance through large-scale multi-task fine-tuning \cite{raffel2020exploring, liu2019multi, aghajanyan2021muppet, aribandi2021ext5}, while other efforts \cite{khashabi2020unifiedqa, mccann1806natural, keskar2019unifying} converted a range of NLP tasks into a single generative question answering format using prompt instructions. 
The instruction tuning began in 2020 with the release of several task collections, including Natural Instructions \cite{mishra2021cross}, Flan 2021 \cite{wei2021finetuned}, and PromptSource \cite{bach2022promptsource}. 
These collections aggregated large NLP datasets and provided templatized instructions for zero-shot prompting, enabling models to generalize to unseen instructions. 
MetaICL \cite{min2021metaicl} emphasized few-shot prompting without explicit instructions, using input-output examples to teach tasks in-context. 
Research confirmed the benefits of task and template diversity, with some studies highlighting the advantages of inverting inputs and outputs to create new tasks \cite{min2021metaicl}. 
The subsequent phase saw the expansion and combination of resources, with collections like SuperNatural Instructions \cite{wang2022benchmarking} and OPT-IML \cite{iyer2022opt} integrating more datasets and tasks. 
This phase also introduced multilingual instruction tuning, as seen in xP3 \cite{muennighoff2022crosslingual}, and incorporated Chain-of-Thought training prompts in Flan 2022 \cite{chung2022scaling}.
These expanded collections included most tasks from previous resources, establishing a strong foundation for future open-source work. 
Current and future research is exploring new directions, such as synthetic data generation for creative and open-ended dialogue tasks \cite{wang2022self, honovich2022unnatural, ye2022guess, gupta2022instructdial} and integrating human feedback on model responses \cite{ouyang2022training, glaese2022improving, nakano2021webgpt, bai2022constitutional}.
These approaches are viewed as complementary to foundational instruction tuning methods, driving further advancements in the field.

A recent advance in instruction tuning is the potential to complement or replace few-shot in-context learning with parameter-efficient fine-tuning. 
Compared to instruction tuning, parameter-efficient fine-tuning can achieve performance comparable to full parameter tuning while being computationally more cost-effective. 
Previous studies \cite{liu2022few, wei2021finetuned, vu2021spot, singhal2023large} have demonstrated that parameter-efficient fine-tuning can be effectively integrated with instruction tuning, either before or after the instruction tuning process. 
Additionally, this body of research highlights that parameter-efficient fine-tuning can enhance the performance and applicability of instruction tuning across different domains.

%% file: Alignment_Tuning_and_RLHF.tex
Despite the emergent abilities brought by increasing parameters of language
models, hallucination exhibit to become a challenge for LLMs to produce
satisfying response. To address this issue, alignment tuning is applied to align the models
with specific human preferences. There are three primary targets for alignment tuning, respectively presented as
helpfulness, honesty and harmlessness. From the targets' names, it can be concluded that the alignment criteria are closely associated with human's recognition, making it difficult to formulate them as optimization objectives for LLMs. Therefore, human feedback is widely
adopted as an assistance to reinforce LLMs' performance.

RLHF~\cite{knox2008tamer, christiano2017deep}
emerged as a method to fine-tune language models using human feedback, aiming to align the LLMs with human preferences, and consequently enhancing alignment performance.  

\begin{figure}[htbp]
    \centering
    \includegraphics[width=0.75\columnwidth]{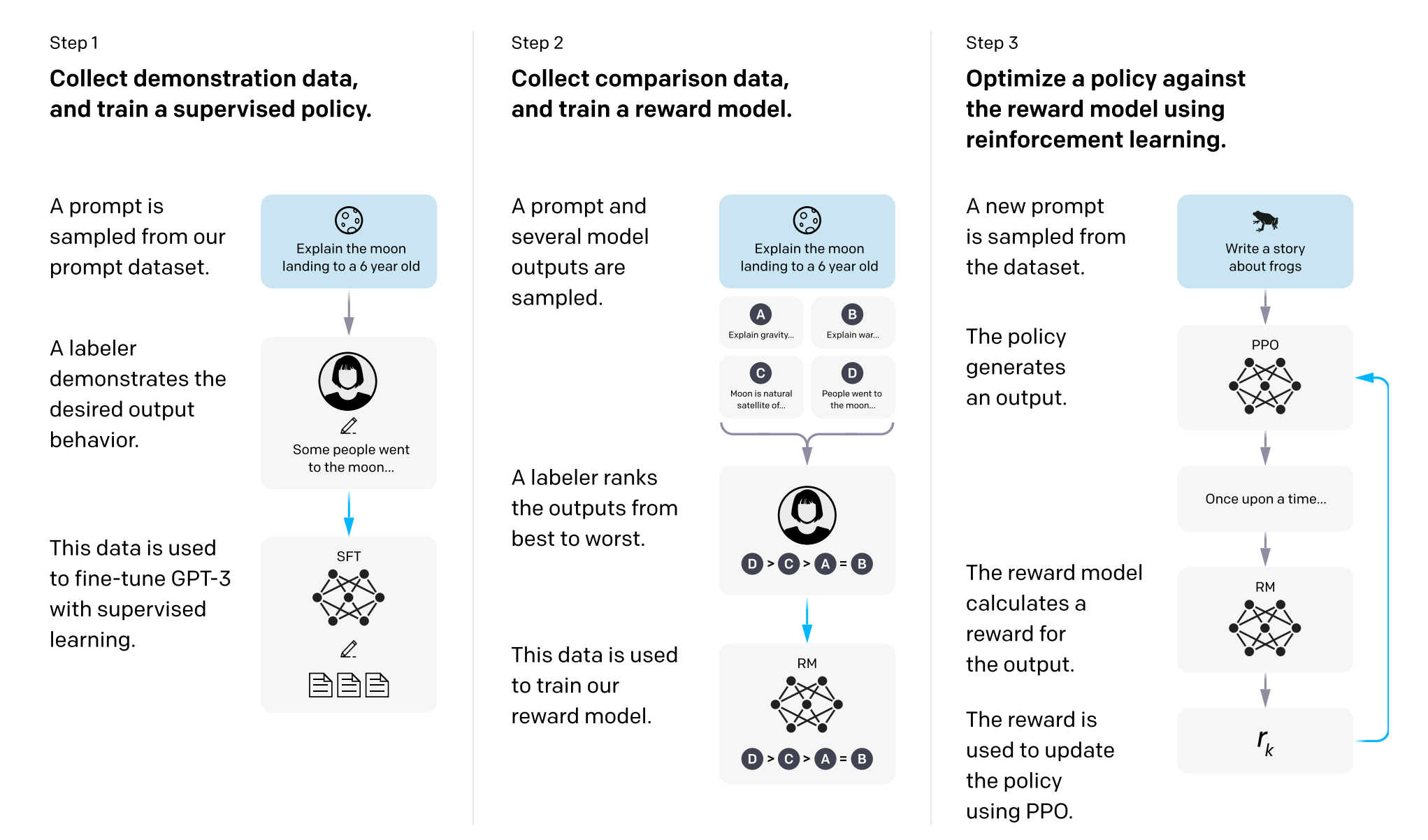} 
    \caption{RLHF Workflow: This figure is from InstructGPT, which interprets the RL process.} 
    \label{fig: RLHF} 
\end{figure}

Generally, an RLHF system\cite{ouyang2022training} comprises three key components: a pre-trained language model, a reward model learned from human feedback, and a reinforcement learning algorithm
to train the language model. Figure \ref{fig: RLHF} shows the three key steps.

\begin{itemize}

  \item \textbf{Supervised Fine-Tuning (SFT):} Initially, a supervised dataset consisting
    of input prompts and desired outputs is applied to fine-tune the language
    model. These prompts and outputs can be written by human labelers for some specific tasks while ensuring the diversity of tasks. This step helps the model learn expected behaviors.

  \item \textbf{Reward Model Training:} A reward model is trained using human feedback
    data. The LLM is employed to generate a certain number of output texts using
    sampled prompts as input. Then human labelers rank these output pairs based
    on their preferences.  
    Given human predictions, the reward model is trained to predict these rankings, effectively learning human preferences. Notably, \cite{lee2023rlaif} proposes an approach, namely Reinforcement Learning from AI Feedback (RLAIF), the annotation of preference on response pairs can be generated by an AI agent, increasing the automatic ability of the reinforcement process.

  \item \textbf{Reinforcement Learning Fine-Tuning:} The final step involves formalizing the alignment process as a reinforcement learning problem. Here, the pre-trained language model acts as a policy generating text, with the reward model providing feedback scores. To prevent the model from deviating too far from its initial state, a penalty term is often included in the reward
  function. The language model is then optimized using algorithms like SARSA~\cite{sutton1995generalization}, DQN~\cite{fan2020theoretical}, PPO~\cite{schulman2017proximal}, DPO~\cite{rafailov2024direct}, and GRPO~\cite{shao2024deepseekmath}, iteratively improving its performance based on human-aligned rewards.
\end{itemize}

%% file: Datasets_for_LLM.tex
A critical component of the development and deployment of LLM is the datasets used at various stages of their lifecycle, which significantly influence their capabilities and performance. In this section, we delve into the datasets that are instrumental in the Pre-training, SFT, and RLHF.
The Pre-training phase is where an LLM absorbs the foundational knowledge from a diverse array of textual data. This stage is pivotal, as it sets the stage for the model's general understanding of language. The datasets used in Pre-training are vast and varied, encompassing everything from the sprawling expanse of the internet to curated collections of literature and encyclopedias.
SFT is the process where the LLM is fine-tuned on specific tasks or domains. This phase refines the model's abilities, enabling it to perform with greater precision and relevance in targeted applications. SFT datasets are often more specialized and may include annotated examples that guide the model towards desired behaviors and outputs.
RLHF is the stage where the LLM is further optimized based on human feedback. This phase enhances the model's alignment with human preferences and values, ensuring that its outputs are more aligned with user expectations. RLHF datasets typically consist of human-labeled examples and feedback, which help the model learn to prioritize high-quality and contextually appropriate responses.

\textbf{Commonly Used Datasets for Pre-training}.
\input{Commonly_Used_Datasets_for_Pre-training}

\textbf{Commonly Used Datasets for SFT and RLHF}.
\input{Commonly_Used_Datasets_for_SFT_and_RLHF}

%% file: Commonly_Used_Datasets_for_Pre-training.tex
In the realm of LLM, the pre-training phase is instrumental in establishing a robust foundation upon which the model's linguistic prowess is built. LLM, with their exponentially larger parameter counts, necessitate an extensive and diverse corpus of training data that spans a multitude of topics and linguistic expressions. This data not only serves as the bedrock for the model's comprehension of language but also influences its ability to generalize and adapt to new contexts and tasks. To meet these requirements, a variety of comprehensive and accessible datasets have been curated and made available for the research community.

In this section, we embark on an overview of the datasets that are pivotal in the pre-training of LLM. We categorize these datasets based on the type of content they provide, which can be broadly divided into seven distinct groups: Webpages, Books, Code, Social Media, Wikipedia, and a diverse array of other sources. Each of these categories contributes unique elements to the model's knowledge base, ensuring a well-rounded understanding of human language and its myriad uses. Here are 2 typical Pre-training Datasets and their importance in evaluating PEFT Methods:

\begin{itemize}
    \item \textbf{Common Crawl:}
The Common Crawl corpus is an extensive, unstructured, multilingual dataset of webpages, encompassing over eight years of web crawler data. This dataset is available in various formats, including web archive, web archive transformation, and web-extracted text. Many pre-training corpora are obtained through data preprocessing based on this corpus, which provides a vast and diverse source of text for language models. Its unstructured nature and multilingual content make it an ideal resource for training models that need to handle a wide variety of text types and languages. Importantly, the Common Crawl corpus plays a crucial role in evaluating PEFT methods. Its vast and varied content provides a comprehensive base for pre-training models that can then be fine-tuned using PEFT techniques. This allows researchers to assess how effectively PEFT methods can enhance model performance across diverse linguistic contexts.
\end{itemize}

\begin{itemize}
    \item \textbf{The Pile:}
The Pile is a large-scale, diverse language modeling dataset consisting of 22 data subsets, designed to capture text in as many forms as possible and cover a wide range of textual content. The corpus includes academic papers, code, legal materials, patents, subtitles, chat content, parallel corpora, and more. This diversity ensures that models trained on The Pile are exposed to a broad spectrum of language use cases, making them more adaptable to various downstream tasks. In the context of evaluating PEFT methods, The Pile offers a robust testbed. Its rich diversity of text types allows researchers to evaluate how well these fine-tuning methods can adapt models to different domains and tasks, thereby enhancing their understanding of the effectiveness of PEFT methods in various applications.
\end{itemize}

\input{Table_of_Pretrain_Datasets}

%% file: Table_of_Pretrain_Datasets.tex
\begin{table*}[t]
\centering
\resizebox{\textwidth}{!}{
\begin{tabular}{p{13em}p{7.5em}p{6.5em}p{5em}p{10em}}
  \toprule
        {\textbf{Collections}} & {\textbf{Categories}} & {\textbf{Publication Time}} & {\textbf{Size}} & {\textbf{URL}} \\
        \midrule
        Common Crawl\footnotemark[4] & Webpages & 2023 & 400TB & \url{https://commoncrawl.org/}   \\
        WuDaoCorpora-Text~\cite{yuan2021wudaocorpora} & Webpages & 2023 & 5TB & \url{https://data.baai.ac.cn/details/WuDaoCorporaText}   \\
        BookCorpusOpen ~\cite{Zhu_2015_ICCV} & Books & 2015 & 9.05GB & \url{https://huggingface.co/datasets/defunct-datasets/bookcorpusopen}   \\
        PG-19 ~\cite{rae2019compressive} & Books & 2020 & 11.74GB & \url{https://huggingface.co/datasets/deepmind/pg19}   \\
        The Stack~\cite{kocetkov2022stack} & Code & 2022 & 3TB & \url{https://huggingface.co/datasets/bigcode/the-stack}   \\
        OpenWebText~\cite{Gokaslan2019OpenWeb} & Social Media & 2019 & 38GB & \url{https://huggingface.co/datasets/Skylion007/openwebtext}   \\
        Pushshift Reddit~\cite{baumgartner2020pushshift} & Social Media & 2020 & 89.1GB & \url{https://zenodo.org/records/3608135}   \\
        Wikipedia\footnotemark[5] & Wikipedia & 2023 & 71.8GB & \url{https://huggingface.co/datasets/wikimedia/wikipedia}   \\
        The Pile~\cite{gao2020pile} & Others & 2020 & 800GB & \url{https://pile.eleuther.ai/}   \\
        S2ORC~\cite{lo2019s2orc} & Others & 2020 & 80.5GB & \url{https://huggingface.co/datasets/sentence-transformers/s2orc}   \\
        MultiUN~\cite{eisele2010multiun} & Others & 2010 & 31.8GB & \url{https://huggingface.co/datasets/Helsinki-NLP/multiun}   \\
  \bottomrule
\end{tabular}
}
\caption{A Curated List of Datasets for Pre-Training. This table provides a comprehensive overview of various datasets used for pre-training purposes in natural language processing tasks. It includes details such as the collection name, the corpus it belongs to, publication year, size in terms of tokens, and the URL for accessing the dataset. The datasets listed cover a range of sources from web pages to books, offering a diverse set of data for training models in different domains.}
\label{tab:my_label1}
\end{table*}

\footnotetext[4]{\url{https://commoncrawl.org/}}
\footnotetext[5]{\url{https://www.wikipedia.org/}}

%% file: Commonly_Used_Datasets_for_SFT_and_RLHF.tex
Two critical stages in LLM are SFT and RLHF. These stages are designed to enhance the model's performance on specific tasks and align its outputs with human preferences. This section provides an overview of these two stages, highlighting their significance and the datasets used to support them.

SFT is a process where LLM are trained on specialized datasets to improve their performance on specific tasks. This stage is crucial for adapting the model to particular domains or applications. SFT involves using annotated datasets that provide examples of desired outputs for given inputs. By training on these datasets, the model learns to generate more accurate and contextually relevant responses.
RLHF is particularly effective in enhancing the model's ability to follow human instructions. These datasets provide a comprehensive set of examples that help the model learn to discern correct answers from plausible alternatives.

\footnotetext[6]{\url{https://www.statmt.org/wmt19/}}
\begin{table*}[t]
\centering
\resizebox{\textwidth}{!}{
\begin{tabular}{p{13em}p{7.5em}p{6.5em}p{5em}p{10em}}
  \toprule
        {\textbf{Collections}} & {\textbf{Categories}} & {\textbf{Publication Time}} & {\textbf{Examples}} & {\textbf{URL}} \\
        \midrule
        E2E NLG~\cite{duvsek2020evaluating} & NLP Task & 2020 & 50,000 & \url{https://sites.google.com/site/hwinteractionlab/E2E/} \\
        WikiSQL~\cite{zhongSeq2SQL2017} & NLP Task & 2017 & 80,654 & \url{https://huggingface.co/datasets/Salesforce/wikisql} \\
        WebNLG~\cite{web_nlg} & NLP Task & 2017 & 27,731 & \url{https://huggingface.co/datasets/web_nlg} \\
        SAMSum~\cite{gliwa-etal-2019-samsum} & Daily Chat & 2019 & 16,369 & \url{https://huggingface.co/datasets/Samsung/samsum} \\
        OASST1~\cite{wang2023openchat} & Daily Chat & 2023 & 161,443 & \url{https://huggingface.co/datasets/OpenAssistant/oasst1} \\
        WMT\footnotemark[6] & Others & 2019 & 124,448,248 & \url{https://huggingface.co/datasets/wmt/wmt19} \\
        XSUM~\cite{Narayan2018DontGM} & Others & 2018 & 200,000 & \url{https://huggingface.co/datasets/EdinburghNLP/xsum} \\
        DART~\cite{radev2020dart} & Text Generation & 2021 & 82,000 & \url{https://github.com/Yale-LILY/dart} \\
        HH-rlhf~\cite{bai2022training} & Dialogue and Preference & 2022 & 169,000 & \url{https://huggingface.co/datasets/Anthropic/hh-rlhf}  \\
        PKU-SafeRLHF~\cite{ji2024beavertails} & Dialogue and Preference & 2023 & 362,000 & \url{https://huggingface.co/datasets/PKU-Alignment/PKU-SafeRLHF}  \\
        HotpotQA~\cite{yang2018hotpotqa} & Question-Answering & 2018 & 113,000 & \url{https://huggingface.co/datasets/hotpotqa/hotpot_qa}  \\
        SHP~\cite{pmlr-v162-ethayarajh22a} & Community Preference & 2022 & 385,000 & \url{https://huggingface.co/datasets/stanfordnlp/SHP}  \\
  \bottomrule
\end{tabular}
}
\caption{A Curated List of datasets for SFT and RLHF. This table provides an overview of the datasets used in SFT and RLHF phases, categorized by their primary purposes and characteristics. This categorization helps in understanding the diversity and scope of data used to train and fine-tune models in different phases of development. The URLs provided allow researchers and practitioners to access these datasets for further analysis and experimentation.}
\label{tab:my_label2}
\end{table*}

%% file: LLM_evaluation.tex
The burgeoning field of LLM research has necessitated the development of robust evaluation frameworks to accurately gauge the capabilities and limitations of these sophisticated AI systems. Evaluation serves multiple critical functions: it benchmarks model performance across a spectrum of tasks, identifies areas for improvement, and ensures that advancements in LLM technology align with ethical and practical standards. In the academic and professional realms of LLM evaluation, it is widely recognized that a multifaceted approach is essential to gauge the capabilities and limitations of these advanced AI systems comprehensively. The Qwen blog’s evaluation of the Qwen2.5 base language model\footnotemark[7], underscore the importance of using multiple benchmarks to assess the model’s performance across various domains thoroughly.

Platforms such as Hugging Face offer a suite of datasets for this purpose\footnotemark[8], including IFEval, BBH, MATH~\cite{hendrycks2021measuring}, GPQA~\cite{rein2024gpqa}, and MUSR ~\cite{sprague2024musrtestinglimitschainofthought}. These datasets encompass a broad spectrum of tasks, ranging from language modeling to problem-solving in mathematics, ensuring a comprehensive evaluation of model competencies. Models like Qwen2.5 is evaluated using a diverse array of datasets that cover general tasks such as MMLU, and HellaSwag, as well as specialized tasks in math and science with datasets like GPQA and MATH, and coding tasks including HumanEval and MBPP. Additionally, multilingual capabilities are assessed through datasets like Multi-Exam and Multi-Translation.

\footnotetext[7]{\url{https://qwenlm.github.io/blog/qwen2.5-llm/}}
\footnotetext[8]{\url{https://huggingface.co/spaces/open-llm-leaderboard/open_llm_leaderboard}}

To achieve a comprehensive evaluation of a language model’s performance, it is often necessary to employ a combination of benchmarks. These benchmarks should be representative of real-world scenarios and cover diverse domains and linguistic complexities. The evaluations include a variety of tests that measure the model’s ability to handle extended dialogues and manage a variety of tasks. By leveraging these diverse datasets and assessments, researchers can effectively benchmark LLM and guide their development towards practical applications, ensuring alignment with ethical and practical standards. This section explores benchmarking in two parts: general tasks and specialized tasks.

\begin{itemize}
    \item \textbf{General Tasks:}
General tasks are designed to assess the broad capabilities of LLM across a wide range of subjects and skills. These benchmarks are essential for evaluating the foundational knowledge and general reasoning abilities of LLM. These benchmarks help determine how well models can understand and generate text in various contexts, ensuring that they possess a solid understanding of language fundamentals. Datasets such as MMLU, ARC, and HellaSwag are commonly used for general evaluations.
\end{itemize}

\begin{itemize}
    \item \textbf{Specialized Tasks:}
Specialized tasks focus on evaluating LLM in specific domains, such as mathematics, coding, and natural language understanding. These benchmarks are designed to assess the model's proficiency in particular areas, providing a deeper understanding of their specialized skills. Specialized tasks are crucial for identifying domain-specific strengths and weaknesses, ensuring that models can effectively apply their knowledge in practical scenarios. 
\end{itemize}

\begin{table*}[t]
\centering
\resizebox{\textwidth}{!}{
\begin{tabular}{p{13em}p{7.5em}p{6.5em}p{5em}p{10em}}
  \toprule
        {\textbf{Collections}} & {\textbf{task}} & {\textbf{Publication Time}} & {\textbf{examples}} & {\textbf{URL}} \\
        \midrule
        MMLU~\cite{hendryckstest2021} & general & 2021 & 15,908 & \url{https://huggingface.co/datasets/cais/mmlu}  \\
        ARC~\cite{allenai:arc} & general & 2018 & 7,787 & \url{https://huggingface.co/datasets/allenai/ai2_arc}  \\
        HellaSwag~\cite{zellers2019hellaswag} & general & 2019 & 59,950 & \url{https://huggingface.co/datasets/Rowan/hellaswag} \\
        GLUE~\cite{wang2018glue} & natural language understanding & 2018 & 1,485,043 & \url{https://huggingface.co/datasets/nyu-mll/glue} \\
        SuperGLUE~\cite{wang2019superglue} & natural language understanding & 2019 & 196,309 & \url{https://huggingface.co/datasets/aps/super_glue} \\
        GSM8K~\cite{cobbe2021gsm8k} & Science and mathematics & 2021 & 17,584 & \url{https://huggingface.co/datasets/openai/gsm8k} \\
        Theoremqa~\cite{chen2023theoremqa} & Science and mathematics & 2023 & 800 & \url{https://huggingface.co/datasets/TIGER-Lab/TheoremQA} \\
        Humaneval~\cite{chen2021evaluating} & code & 2021 & 164 & \url{https://huggingface.co/datasets/openai/openai_humaneval} \\
        MBPP~\cite{austin2021program} & code & 2021 & 1,401 & \url{https://huggingface.co/datasets/google-research-datasets/mbpp} \\
        AGIEval~\cite{zhong2023agieval} & Exam & 2023 & 8,062 & \url{https://github.com/ruixiangcui/AGIEval} \\
        GAOKAO-Bench~\cite{Zhang2023EvaluatingTP} & Exam & 2023 & 2,811 & \url{https://github.com/OpenLMLab/GAOKAO-Bench} \\
        TruthfulQA~\cite{lin2021truthfulqa} & other & 2021 & 1,634 & \url{https://huggingface.co/datasets/truthfulqa/truthful_qa} \\
        BBH~\cite{suzgun2022challenging} & other & 2022 & 6,511 & \url{https://huggingface.co/datasets/lukaemon/bbh} \\
  \bottomrule
\end{tabular}
}
\caption{A typical list of available datasets for LLM Evaluation. This table provides an exhaustive compilation of datasets pertinent to the evaluation of LLM. These datasets span a diverse array of tasks, from general to domain-specific, aiming to holistically assess the performance of LLM across various scenarios. The table delineates the publication timeline, the number of examples, and the access points (URLs) for each dataset, facilitating researchers in procuring and utilizing these resources.}
\label{tab:my_label3}
\end{table*}

%% file: PEFT_Taxonomy.tex
\begin{figure}[htbp]
    \centering
    \includegraphics[width=\columnwidth]{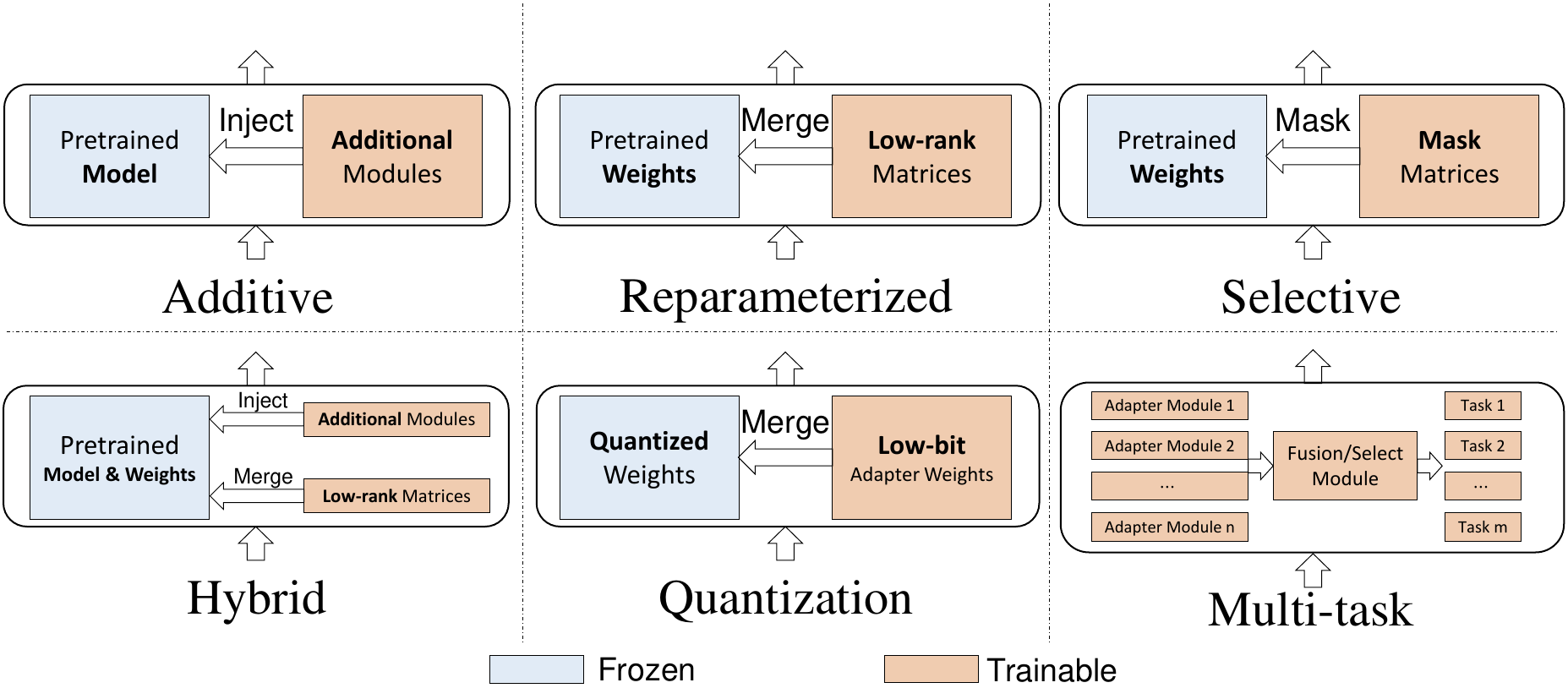} 
    \caption{Illustration of the main idea of different types of PEFT methods} 
    \label{fig:additive_reparameterized_subtractive_compare} 
\end{figure}

PEFT techniques are typically divided into three primary categories:
\textbf{Additive PEFT} (\ref{additive_peft}), which introduces additional trainable components or parameters into the pre-existing model; \textbf{Reparameterized PEFT} (\ref{reparameterized_peft}), a method that restructures the model's parameters during the training phase and then reverts to the original form for inference; and \textbf{Selective PEFT} (\ref{subtractive_peft}), which focuses on optimizing a specific subset of the model's parameters. Besides these, there is the \textbf{Hybrid PEFT} (\ref{hybrid_peft}), which combines the strengths of various PEFT approaches. Additionally, there are specialized adaptations such as \textbf{Quantization PEFT} (\ref{quantization_peft}) designed for the quantization process, and \textbf{Multi-task PEFT} (\ref{multi_task_peft}) aimed at enhancing multi-task learning capabilities. A conceptual illustration of the core principles underlying these PEFT methodologies is presented in Figure~\ref{fig:additive_reparameterized_subtractive_compare}. A comprehensive classification of PEFT methods is depicted in Figure~\ref{fig:taxonomy_of_peft}. The main ideas, number of trainable parameters, applications, and limitations of different types of PEFT methods are summarized in Table~\ref{tab:overview_of_peft}. To facilitate a more intuitive understanding of the performance differences among various PEFT methods, Table~\ref{tab:peft_performance_compare} presents the performance results of representative PEFT methods of different types across various base models and tasks.

\tikzstyle{my-box}=[
 rectangle,
 draw=hidden-draw,
 rounded corners,
 text opacity=1,
 minimum height=1.5em,
 minimum width=5em,
 inner sep=2pt,
 align=center,
 fill opacity=.5,
 ]
 \tikzstyle{leaf}=[my-box, minimum height=1.5em,
 fill=hidden-orange!60, text=black, align=left,font=\scriptsize,
 inner xsep=2pt,
 inner ysep=4pt,
 ]

\begin{figure*}[t]
\centering
\resizebox{\textwidth}{!}{
\begin{forest}
        forked edges,
        for tree={
            grow=east,
            reversed=true,
            anchor=base west,
            parent anchor=east,
            child anchor=west,
            base=left,
            font=\small,
            rectangle,
            draw=hidden-draw,
            rounded corners,
            align=left,
            minimum width=4em,
            edge+={darkgray, line width=1pt},
            s sep=3pt,
            inner xsep=2pt,
            inner ysep=3pt,
            ver/.style={rotate=90, child anchor=north, parent anchor=south, anchor=center},
        },
        where level=1{text width=7.0em,font=\scriptsize}{},
        where level=2{text width=7.0em,font=\scriptsize}{},
        where level=3{text width=7.0em,font=\scriptsize}{},
        [
            PEFT Methods for PLMs, ver
            [
                Additive 
                [
                    Adapter
                    [
                        Sequential Adapter~{\cite{houlsby2019parameter},} 
                        Residual Adapter~{\cite{lin2020exploring},} \\
                        AdapterDrop~{\cite{ruckle2020adapterdrop},} 
                        Tiny-Attn Adapter~{\cite{zhao2022tiny},} \\
                        Parallel Adapter~{\cite{he2021towards},} 
                        CIAT~{\cite{zhu2021counter},} 
                        CoDA~{\cite{lei2024conditional},} \\
                        Hadamard Adapter~{\cite{chen2023hadamard},} 
                        Compacter~{\cite{karimi2021compacter},} \\
                        SparseAdapter~{\cite{he2022sparseadapter}} 
                        , leaf, text width=35.6em
                    ]
                ]
                [
                    Soft Prompt 
                    [
                        Prefix-tuning~{\cite{li2021prefix},} 
                        p-tuning~{\cite{liu2021gpt},} \\
                        p-tuning v2~{\cite{liu2021p},} 
                        prompt-tuning~{\cite{lester2021power},} \\
                        SMoP~{\cite{choi2023smop},} 
                        APT~{\cite{zhang2023towards},} 
                        IDPG~{\cite{wu2022idpg},} \\
                        LPT~{\cite{liu2022late},} 
                        SPT~{\cite{zhu2023spt},}
                        APrompt~{\cite{wang2023aprompt},} \\
                        DePT~{\cite{shi2023dept},} 
                        InfoPrompt~{\cite{wu2024infoprompt},} 
                        Xprompt~{\cite{ma2022xprompt},} \\
                        PTP~{\cite{chen2023ptp}}
                        , leaf, text width=35.6em  
                    ]
                ]
                [
                    Scale and Shift 
                    [
                        $\text{(IA)}^{3}$~{\cite{liu2022few},}
                        MoV~{\cite{zadouri2023pushing},}
                        SSF~{\cite{lian2022scaling},} \\
                        PASTA~{\cite{yang2023parameter}} 
                        , leaf, text width=35.6em
                    ]
                ]
                [
                    Others 
                    [
                        IPA~{\cite{lu2023inference},} 
                        LST~{\cite{sung2022lst},} 
                        Attention-Fusion~{\cite{cao2022attention}} 
                        , leaf, text width=35.6em
                    ]
                ]
            ]
            [
                Reparameterized
                [
                    Low-rank \\ Decomposition 
                    [
                        Intrinsic SAID~{\cite{aghajanyan2020intrinsic},} 
                        LoRA~{\cite{hu2021lora},} \\
                        KronA~{\cite{edalati2022krona}}
                        , leaf, text width=35.6em
                    ]
                ]
                [
                    LoRA \\ Derivatives
                    [
                        Dynamic Rank
                        [
                            DyLoRA~{\cite{valipour2023dylora},} 
                            AdaLoRA~{\cite{zhang2023adalora},} \\
                            IncreLoRA~{\cite{zhang2023increlora},} 
                            SoRA~{\cite{ding2023sparse}} 
                            , leaf, text width=27em
                        ]
                    ]
                    [
                        LoRA \\Improvement
                        [
                            LoRA+~{\cite{hayou2024lora+},} 
                            LoRA-FA~{\cite{zhang2023lora},} \\
                            DoRA~{\cite{liu2024dora},} 
                            Laplace-LoRA~{\cite{yang2023bayesian},} \\
                            Kernel-mix~{\cite{chen2022empowering},} 
                            PeriodicLoRA~{\cite{meng2024periodiclora},} \\
                            HydraLoRA~{\cite{tian2024hydralora},}
                            AFLoRA~{\cite{liu2024aflora},} \\
                            LoRA-SP~{\cite{wu2024lora},} 
                            SuperLoRA~{\cite{chen2024superlora}}
                            , leaf, text width=27em
                        ]
                    ]
                ]
            ]
            [
                Selective
                [
                    Unstructural \\ Masking 
                    [
                        U-Diff pruning~{\cite{guo2020parameter},} 
                        U-Bitfit~{\cite{lawton2023neural},} 
                        PaFi~{\cite{liao2023parameter},} \\
                        FishMask~{\cite{sung2021training},} 
                        Fish-Dip~{\cite{das2023unified},} 
                        LT-SFT~{\cite{ansell2021composable},} \\
                        SAM~{\cite{fu2023effectiveness},} 
                        Child-tuning~{\cite{xu2021raise},} 
                        U-MAM~{\cite{lawton2023neural},} \\
                        Threshold-Mask~{\cite{zhao2020masking},} 
                        LoRAPrune~{\cite{zhang2023pruning}}
                        , leaf, text width=35.6em
                    ]
                ]
                [
                    Structural \\ Masking 
                    [
                        S-Diff pruning~{\cite{guo2020parameter},} 
                        S-Bitfit~{\cite{lawton2023neural},} \\
                        FAR~{\cite{vucetic2022efficient},} 
                        Bitfit~{\cite{zaken2021bitfit},} 
                        Xattn Tuning~{\cite{gheini2021cross},} \\
                        SPT~{\cite{he2023sensitivity},} 
                        S-MAM~{\cite{lawton2023neural}} 
                        , leaf, text width=35.6em
                    ]
                ]
            ]
            [
                Hybrid
                [
                    UniPELT~{\cite{mao2021unipelt},} 
                    S4~{\cite{chen2023parameter},} 
                    MAM Adapter~{\cite{he2021towards},} \\
                    LLM-Adapters~{\cite{hu2023llm},} 
                    NOAH~{\cite{zhang2022neural},} 
                    AUTOPEFT~{\cite{zhou2024autopeft},} \\
                    $\text{S}^{3}\text{Delta-M}$~{\cite{hu2022sparse},} 
                    ProPETL~{\cite{zeng2023one}} 
                    , leaf, text width=44.2em
                ]
            ]
            [
                Quantization
                [
                    BI-Adapter~{\cite{jie2023revisiting},} 
                    PEQA~{\cite{kim2024memory},}
                    QLoRA~{\cite{dettmers2024qlora},} \\
                    LQ-LoRA~{\cite{guo2023lq},} 
                    QA-LoRA~{\cite{xu2023qa},} 
                    QDyLoRA~{\cite{rajabzadeh2024qdylora},} \\
                    LoftQ~{\cite{li2023loftq},} 
                    BitDelta~{\cite{liu2024bitdelta}}
                    , leaf, text width=44.2em
                ]
            ]
            [
                Multi-task 
                [
                    Adapter\\-based 
                    [
                        AdapterFusion~{\cite{pfeiffer2020adapterfusion},}
                        AdaMix~{\cite{wang2022adamix},} \\
                        PHA~{\cite{zhao2023prototype},} 
                        AdapterSoup~{\cite{chronopoulou2023adaptersoup},} \\
                        MerA~{\cite{he2023mera},}
                        Hyperformer~{\cite{mahabadi2021parameter}}
                        , leaf, text width=35.6em
                    ]
                ]
                [
                    Soft Prompt\\-based 
                    [
                        SPoT~{\cite{vu2022spot},} 
                        ATTEMPT~{\cite{asai2022attempt},} 
                        MPT~{\cite{wang2022multitask},} \\
                        IPT~{\cite{qin2021exploring},} 
                        TPT~{\cite{su2021transferability}} 
                        , leaf, text width=35.6em
                    ]
                ]
                [
                    LoRA\\-based 
                    [
                        LoRAHub~{\cite{huang2023lorahub},} 
                        MOELoRA~{\cite{liu2023moelora},} \\
                        L-LoRA~{\cite{tang2023parameter},} 
                        MTLoRA~{\cite{agiza2024mtlora}}
                        , leaf, text width=35.6em
                    ]
                ]
            ]      
        ]
    \end{forest}
}
\caption{Taxonomy of PEFT Methods}
\label{fig:taxonomy_of_peft}
\end{figure*}

\begin{table*}[t]
\centering
\resizebox{\textwidth}{!}{
\begin{tabular}{p{7.5em}p{7.5em}p{7.5em}p{4.5em}p{5em}p{5em}p{5em}}
  \toprule
  \textcolor[rgb]{ .251,  .251,  .251}{\textbf{Category}} & \textcolor[rgb]{ .251,  .251,  .251}{\textbf{Main Idea}} & \textcolor[rgb]{ .251,  .251,  .251}{\textbf{Representative Methods}} & \textcolor[rgb]{ .251,  .251,  .251}{\textbf{\#Trainable Parameters}} & \textcolor[rgb]{ .251,  .251,  .251}{\textbf{Applications}} & \textcolor[rgb]{ .251,  .251,  .251}{\textbf{Advantages}} & \textcolor[rgb]{ .251,  .251,  .251}{\textbf{Limitations}} \\
  \midrule
  \textcolor[rgb]{ .251,  .251,  .251}{\textbf{Additive}} & \textcolor[rgb]{ .251,  .251,  .251}{Add trainable components, freeze original.} & \textcolor[rgb]{ .251,  .251,  .251}{Sequential Adapter~\cite{houlsby2019parameter}, Prefix-tuning~\cite{li2021prefix}, $(IA)^{3}$~\cite{liu2022few}, IPA~\cite{lu2023inference}} & \textcolor[rgb]{ .251,  .251,  .251}{\#Params of additional modules} & \textcolor[rgb]{ .251,  .251,  .251}{Single-task, rapid adaptation.} & \textcolor[rgb]{ .251,  .251,  .251}{Minimal updates, flexible insertion.} & \textcolor[rgb]{ .251,  .251,  .251}{Computational overhead, design-sensitive.} \\
  \midrule
  \textcolor[rgb]{ .251,  .251,  .251}{\textbf{Reparameterized}} & \textcolor[rgb]{ .251,  .251,  .251}{Low-rank decomposition, tune low-rank matrices.} & \textcolor[rgb]{ .251,  .251,  .251}{LoRA~\cite{hu2021lora}, AdaLoRA~\cite{zhang2023adalora}, DoRA~\cite{liu2024dora}} & \textcolor[rgb]{ .251,  .251,  .251}{\#Params of low-rank matrices} & \textcolor[rgb]{ .251,  .251,  .251}{Large-scale, efficient updates.} & \textcolor[rgb]{ .251,  .251,  .251}{Fewer parameters, no inference latency.} & \textcolor[rgb]{ .251,  .251,  .251}{Low-rank constraints, hyperparameter tuning.} \\
  \midrule
  \textcolor[rgb]{ .251,  .251,  .251}{\textbf{Selective}} & \textcolor[rgb]{ .251,  .251,  .251}{Update subsets (e.g., biases, masked params).} & \textcolor[rgb]{ .251,  .251,  .251}{U-Bitfit~\cite{lawton2023neural}, FAR~\cite{vucetic2022efficient}} & \textcolor[rgb]{ .251,  .251,  .251}{\#Params of selected subsets (e.g., biases, masked params)} & \textcolor[rgb]{ .251,  .251,  .251}{Resource-constrained environments.} & \textcolor[rgb]{ .251,  .251,  .251}{Critical updates, low memory.} & \textcolor[rgb]{ .251,  .251,  .251}{Task-sensitive, parameter selection.} \\
  \midrule
  \textcolor[rgb]{ .251,  .251,  .251}{\textbf{Hybrid}} & \textcolor[rgb]{ .251,  .251,  .251}{Combine multiple PEFT methods dynamically.} & \textcolor[rgb]{ .251,  .251,  .251}{UniPELT~\cite{mao2021unipelt}, MAM-Adapter~\cite{he2021towards}} & \textcolor[rgb]{ .251,  .251,  .251}{\#Params of used PEFT modules} & \textcolor[rgb]{ .251,  .251,  .251}{Complex tasks, multimodal.} & \textcolor[rgb]{ .251,  .251,  .251}{Task flexibility, improved performance.} & \textcolor[rgb]{ .251,  .251,  .251}{High complexity, search overhead.} \\
  \midrule
  \textcolor[rgb]{ .251,  .251,  .251}{\textbf{Quantization}} & \textcolor[rgb]{ .251,  .251,  .251}{Quantize model, enable efficient tuning.} & \textcolor[rgb]{ .251,  .251,  .251}{QLoRA~\cite{dettmers2024qlora}, BitDelta~\cite{liu2024bitdelta}} & \textcolor[rgb]{ .251,  .251,  .251}{\#Params of used PEFT modules} & \textcolor[rgb]{ .251,  .251,  .251}{Edge devices, low resources.} & \textcolor[rgb]{ .251,  .251,  .251}{Low storage, low-precision inference.} & \textcolor[rgb]{ .251,  .251,  .251}{Precision loss, quantization balance.} \\
  \midrule
  \textcolor[rgb]{ .251,  .251,  .251}{\textbf{Multi-task}} & \textcolor[rgb]{ .251,  .251,  .251}{Share parameters and dynamic adapters for multi-task.} & \textcolor[rgb]{ .251,  .251,  .251}{AdapterFusion~\cite{pfeiffer2020adapterfusion}, SPoT~\cite{vu2022spot}, MOELoRA~\cite{liu2023moelora}} & \textcolor[rgb]{ .251,  .251,  .251}{\#Params of shared and task-specific modules} & \textcolor[rgb]{ .251,  .251,  .251}{Multi-task, cross-task knowledge.} & \textcolor[rgb]{ .251,  .251,  .251}{Redundant reduction, task transfer.} & \textcolor[rgb]{ .251,  .251,  .251}{Task conflicts, routing complexity.} \\
  \bottomrule
\end{tabular}
}
\caption{An Overview of Different Types of PEFT Methods: Main Idea, Number of Trainable Parameters, Applications, and Limitations.}
\label{tab:overview_of_peft}
\end{table*}

\begin{table*}[t]
  \centering
  \resizebox{\textwidth}{!}{
    \begin{tabular}{c|c|c|ccccccccc}
    \toprule
    \textbf{Model} & \textbf{PEFT Type} & \textbf{PEFT Method} & \textbf{\#TPs} & \textbf{CoLA} & \textbf{SST2} & \textbf{MRPC} & \textbf{STS-B} & \textbf{QQP} & \textbf{MNLI} & \textbf{QNLI} & \textbf{RTE} \\
    \midrule
    \multirow{9}[10]{*}{RoBERTa-base} &       & FT    & 124.6M & 59.07 & 92.89 & 88.24/91.58 & 90.87/90.61 & 90.81/87.72 & 86.27 & 91.07 & 72.2 \\
\cmidrule{2-12}          & \multirow{3}[2]{*}{Additive} & AdapterS & 7.41M & 63.32 & 94.31 & 90.44/93.18 & 91.25/90.94 & 90.81/86.55 & 87.33 & 92.06 & 73.56 \\
          &       & Prefix-tuning & 0.96M & 59.31 & 93.81 & 87.25/91.03 & 88.48/88.32 & 87.75/84.09 & 85.21 & 90.77 & 54.51 \\
          &       & (IA)3 & 0.66M & 59.58 & 93.92 & 87.00/90.52 & 90.30/90.32 & 87.99/84.10 & 83.95 & 90.88 & 71.12 \\
\cmidrule{2-12}          & \multirow{2}[2]{*}{Reparameterized} & LoRA  & 0.89M & 62.09 & 94.04 & 87.50/90.68 & 90.66/90.83 & 88.83/85.21 & 86.54 & 92.02 & 72.92 \\
          &       & AdaLoRA & 1.03M & 59.82 & 93.92 & 87.99/91.33 & 90.83/90.73 & 88.58/84.98 & 86.26 & 91.43 & 70.04 \\
\cmidrule{2-12}          & \multirow{2}[2]{*}{Selective} & BitFit & 0.69M & 61.32 & 94.72 & 89.22/92.41 & 90.34/90.27 & 88.12/84.11 & 84.64 & 91.09 & 77.98 \\
          &       & Child-Tuning & /     & 60.33 & 93.58 & 89.22/92.20 & 91.14/90.93 & 90.98/88.04 & 87.4  & 92.2  & 77.62 \\
\cmidrule{2-12}          & Hybrid & MAM Adapter & 46.78M & 61.42 & 94.87 & 89.31/92.21 & 90.74/90.42 & 88.31/83.20 & 86.63 & 90.19 & 72.62 \\
    \midrule
    \multirow{9}[10]{*}{RoBERTa-large} &       & FT    & 355.3M & 65.78 & 95.54 & 89.22/92.28 & 91.74/91.76 & 89.30/86.68 & 89.42 & 93.61 & 81.23 \\
\cmidrule{2-12}          & \multirow{3}[2]{*}{Additive} & AdapterS & 19.77M & 67.03 & 96.37 & 89.94/92.54 & 92.58/92.42 & 92.19/88.50 & 91    & 94.31 & 85.25 \\
          &       & Prefix-tuning & 2.03M & 59.01 & 95.76 & 88.24/91.37 & 90.92/91.07 & 88.88/85.45 & 89.3  & 93.32 & 74.01 \\
          &       & (IA)3 & 1.22M & 61.15 & 94.61 & 86.52/90.33 & 92.22/92.03 & 89.45/86.25 & 88.63 & 94.25 & 81.23 \\
\cmidrule{2-12}          & \multirow{2}[2]{*}{Reparameterized} & LoRA  & 1.84M & 64.47 & 96.67 & 87.50/91.19 & 91.66/91.44 & 90.15/86.91 & 90.76 & 95    & 79.78 \\
          &       & AdaLoRA & 2.23M & 65.85 & 94.95 & 89.46/92.34 & 92.05/91.80 & 89.60/86.30 & 90.36 & 94.62 & 77.98 \\
\cmidrule{2-12}          & \multirow{2}[2]{*}{Selective} & BitFit & 1.32M & 68.01 & 96.1  & 90.93/93.38 & 91.93/91.77 & 89.48/86.43 & 89.98 & 94.47 & 87.73 \\
          &       & Child-Tuning & /     & 63.08 & 95.07 & 90.69/93.43 & 92.36/92.18 & 91.52/88.75 & 35.45 & 93.15 & 86.25 \\
\cmidrule{2-12}          & Hybrid & MAM Adapter & 122.2M & 67.39 & 95.81 & 90.12/92.77 & 92.44/92.18 & 90.87/86.65 & 90.62 & 94.31 & 86.62 \\
    \midrule
    \multirow{3}[4]{*}{DeBERTaV3-base} &       & FT    & /     & 69.2  & 95.3  & 89.5/93.3 & 91.6/91.1 & 92.4/89.8 & 90.5  & 94    & 82 \\
\cmidrule{2-12}          & \multirow{2}[2]{*}{Quantization} & QLoRA & /     & N.A.  & 86.5  & 73.8/82.8 & 83.0/82.8 & 86.8/82.3 & 75.4  & 82.4  & 55.9 \\
          &       & LoftQ & /     & 37.4  & 90.2  & 83.8/88.6 & 87.1/86.9 & 90.3/86.9 & 84.7  & 86.6  & 61.4 \\
    \bottomrule
    \end{tabular}
    }
  \caption{Performance evaluation across various PEFT methods for fine-tuning common base models (RoBERTa-base, RoBERTa-large, and DeBERTaV3-base) on the GLUE benchmark. All performance metrics are cited from prior published works~\cite{xu2023parameter,li2023loftq}. Metrics may vary by task: Matthews correlation for COLA, accuracy/F1 score for MRPC and QQP, Pearson/Spearman correlation for STS-B, average matched accuracy for MNLI, and accuracy for the remaining tasks. Higher metric values indicate superior performance. $\#TP$ denotes the number of trainable parameters for each method.}
  \label{tab:peft_performance_compare}
\end{table*}

\subsection{Additive PEFT}
\label{additive_peft}
\input{Additive_PEFT}

\subsection{Reparameterized PEFT}
\label{reparameterized_peft}
\input{Reparameterized_PEFT}

\subsection{Selective PEFT}
\label{subtractive_peft}
\input{Subtractive_PEFT}

\subsection{Hybrid PEFT}
\label{hybrid_peft}
\input{Hybrid_PEFT}

\subsection{Quantization PEFT}
\label{quantization_peft}
\input{Quantization_PEFT}

\subsection{Multi-task PEFT}
\label{multi_task_peft}
\input{Multi-task_PEFT}

%% file: Additive_PEFT.tex
Full-parameter fine-tuning is computationally expensive and could adversely affect the model's capacity to generalize. To address this, additive PEFT methods add a small set of trainable parameters to a pre-trained model, carefully integrated into its architecture. When fine-tuning for particular downstream tasks, it is only these extra components or parameters are adjusted, keeping the original pre-trained model parameters unchanged. This approach significantly reduces the need for storage, memory, and computation. Based on where and how these additional trainable parameters are incorporated into the model's architecture, there are primarily three types of additive PEFT techniques: \textit{Adapter}, \textit{Soft Prompt}, and \textit{Scale and Shift}. We will delve into some of the principal studies on these techniques.

\subsubsection{Adapter}
\label{adapter}
\input{Adapter}

\subsubsection{Soft Prompt}
\label{soft_prompt}
\input{Soft_Prompt}

\subsubsection{Scale and Shift}
\label{scale_shift}
\input{Scale_and_Shift}

\subsubsection{Others}
\label{others}
\input{Others}

%% file: Adapter.tex
\begin{figure*}[htbp]
    \centering
    \includegraphics[width=\textwidth]{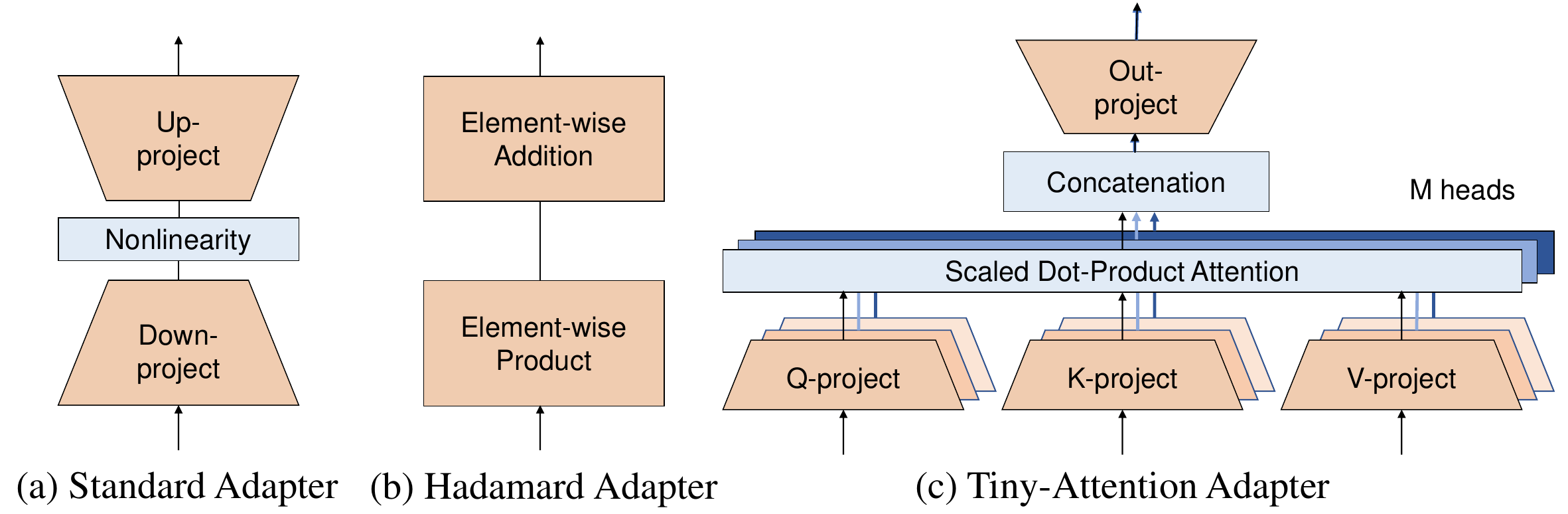} 
    \caption{Illustration of three representative types of adapter.} 
    \label{fig:different_types_adapter} 
\end{figure*}

Adapter methods enable parameter-efficient fine-tuning by inserting small adapter layers into pre-trained models, which learn task-specific transformations while keeping the base model frozen. These adapters, typically consisting of a down-projection, a non-linear activation function and an up-projection layer (the standard adapter shown in Figure~\ref{fig:different_types_adapter} (a)), adapt the representations to downstream tasks with minimal overhead. For example, in \textbf{Sequential Adapter}~\cite{houlsby2019parameter}, two serial adapters are inserted after the attention layer and the feed-forward
layer in transformer blocks. \textbf{Residual Adapter}~\cite{lin2020exploring} dynamically adapts a pre-trained language model, such as GPT-2, to various downstream tasks using low-rank residual adapters and task embeddings, with the adapter module formulated as:
\begin{equation}
    \text{Adapter}(H_i) = (\text{ReLU}(\text{LN}(H_i)W^E_i))W^D_i + H_i
    \enskip,
\end{equation}
where \( H_i \) is the hidden representation of the $i^{\text{th}}$ layer, \( W^E_i \) and \( W^D_i \) are the adapter parameters, and \( \text{LN} \) denotes layer normalization. \textbf{AdapterDrop}~\cite{ruckle2020adapterdrop} dynamically removes adapters from the lower layers of a transformer during training and inference, which significantly enhances inference speed in multi-task settings with minimal impact on task performance. \textbf{Tiny-Attn Adapter}~\cite{zhao2022tiny} applies a multi-head attention mechanism with tiny per-head dimension the intermediate embeddings of each token to obtain the modified embeddings, and employs parameter-averaging technique to reduce inference cost during deployment. \textbf{Parallel Adapter}~\cite{he2021towards} integrates the adapter network to both the attention and feed-forward layers of the transformer in a parallel manner, facilitating a more efficient incorporation of the module. \textbf{CIAT} (Counter-Interference Adapter for Multilingual Machine Translation)~\cite{zhu2021counter} employs an embedding adapter to refine multilingual word embeddings and parallel layer adapters to de-noise the multilingual interference in intermediate layers, improving the translation performance with a small parameter overhead. \textbf{CoDA} (Condition Adapter)~\cite{lei2024conditional} enhances inference efficiency by selectively activating computations on a subset of input tokens, determined by a soft top-\( k \) operation, thus balancing model expressivity and computational efficiency. \textbf{Hadamard Adapter}~\cite{chen2023hadamard} (shown in Figure~\ref{fig:different_types_adapter} (b)) employs a weight vector and a bias vector, applying the Hadamard product (element-wise multiplication) and element-wise addition to the self-attention outputs, resulting in new self-attention outputs. \textbf{Compacter}~\cite{karimi2021compacter} incorporates concepts from adapters, low-rank methods, and hypercomplex multiplication layers. It introduces task-specific weight matrices by combining shared ``slow" weights with ``fast" rank-one matrices computed through Kronecker products, tailored to each COMPACTER layer's requirements. \textbf{SparseAdapter}~\cite{he2022sparseadapter} prunes a significant portion of parameters at initialization, using a sparsity-inducing method to maintain performance while reducing computational overhead, and further improving capacity through a ``Large-Sparse" configuration that scales up the bottleneck dimension with an increased sparsity ratio.

%% file: Soft_Prompt.tex
Soft prompt methods involve appending a sequence of trainable continuous vectors, known as soft prompts, to the input of pre-trained language models. These soft prompts act as additional context that guides the model towards the desired output for a specific task. During training, the soft prompts are optimized to facilitate the model's adaptation to the new task, while the rest of the model remains largely unchanged, making the approach parameter-efficient. Based on the intuition that a properly optimized context, in the form of continuous word embeddings, can guide the language model towards performing an NLG task without altering its parameters, \textbf{Prefix-tuning}~\cite{li2021prefix} and \textbf{prompt-tuning}~\cite{lester2021power} involve prepending a prefix \( P_{\theta} \) of trainable vectors \(\theta\) to the input. The activations for these prefix indices are treated as free parameters. To stabilize the optimization process, \( P_{\theta} \) is parametrized by reparameterizing it through a smaller matrix \( P'_{\theta} \), which is then composed with a feedforward neural network (MLP), i.e., \(P_{\theta} = \operatorname{MLP} (P'_{\theta})\). \textbf{p-tuning}~\cite{liu2021gpt} leverages trainable continuous prompt embeddings, which are concatenated with discrete prompts to form an input sequence for a pretrained language model. This sequence is then mapped to a hidden representation through an embedding function parameterized by a prompt encoder, such as an LSTM or MLP, and is optimized via backpropagation to minimize a task-specific loss function. \textbf{p-tuning v2}~\cite{liu2021p} is an optimized prompt tuning method that universally matches the performance of fine-tuning across various model scales and NLU tasks by applying trainable continuous embeddings to every layer of the pre-trained model as prefix tokens, thus increasing the capacity of continuous prompts and reducing the gap to fine-tuning, especially for smaller models and more challenging tasks. \textbf{SMoP} (Sparse Mixture-of-Prompts)~\cite{choi2023smop} utilizes a gating mechanism to route each input instance to one of multiple short soft prompts, which are specialized in handling different subsets of the data, thereby achieving efficient training and inference while maintaining performance gains typically induced by longer soft prompts. The routing probability for the $j$-th prompt is calculated as \(p_j(X) = [\operatorname{softmax}(L_\mu(\bar{X}))]_j\), where \(L_\mu\) is a small linear router model, \(\bar{X}\) is the average of input embeddings, and \(\mu\) are the parameters of the router model. \textbf{APT} (Adaptive Prefix Tuning)~\cite{zhang2023towards} dynamically customizes the prefix at each layer of a Transformer model through a gate mechanism. It utilizes both fine-grained gated weight assignment and coarse-grained scaled weight specification. The pseudo prefix tokens \(\hat{P}_i\) in the \(i^{\text{th}}\) layer are updated as follows:
\begin{equation}
    \hat{P}_i = \lambda_i \odot \alpha_i \cdot [P_{ik}, P_{iv}]
    \enskip,
\end{equation}
where \([P_{ik}, P_{iv}]\) represents the keys-values pair of the original pseudo prefix tokens, \(\lambda_i\) is a learnable scaled weight, \(\odot\) denotes element-wise multiplication, and \(\alpha_i\) represents the gated weights, which are calculated as:
\begin{equation}
    \alpha_i = \operatorname{sigmoid}(h_{i-1}W_i)
    \enskip,
\end{equation}
where \( h_{i-1} \) represents the hidden states from the previous layer, and \( W_i \) are the parameters to be learned. \textbf{IDPG} (Instance Dependent Prompt Generation)~\cite{wu2022idpg} works on the principle of generating prompts for each input instance using a lightweight model \( G \) that takes the instance representation \( x \) and task \( T \) as inputs to produce a task-specific prompt \( W_p(T, x) \), which is then inserted into the input sequence \( x \) for fine-tuning the pre-trained language model \( M \) with a unified template, as denoted by the equations:
\begin{equation}
\begin{aligned}
W_p(T, x) &= G(M(x), T), \quad x \in D_{train}, \\
h[CLS] &= M(\operatorname{concat}[x, W_p(T, x)]) 
\enskip.
\end{aligned}
\end{equation} 
\textbf{LPT} (Late Prompt Tuning)~\cite{liu2022late} is a method that inserts a ``late prompt" into a pre-trained model (PTM) at an intermediate layer. This late prompt is created by a neural prompt generator (NPG) which uses the hidden states from the model layer just before the prompt insertion. This process generates a prompt that is tailored to each specific instance, enhancing the model's performance and efficiency. The generation of this instance-aware prompt involves a series of steps that include transformations and combinations of various elements derived from the model's hidden states. Once created, the prompt is reshaped to be integrated into the model's processing workflow. \textbf{SPT} (Selective Prompt Tuning)~\cite{zhu2023spt} initializes a prompt hyper-network where each intermediate layer of the pre-trained model (PTM) has a prompt generation layer controlled by a learnable probabilistic gate \( \alpha_i \), which is optimized to determine the importance of each layer for the task at hand, using the formulation \( a_i = \sigma(\alpha_i) \), where \( \sigma \) is the sigmoid function, and \( p_i \), the prompt at layer \( i \), is calculated as \( p_i = (1 - \tau \cdot a_i) \cdot p_{\text{prev},i} + \tau \cdot a_i \cdot p_{\text{new},i} \), with \( \tau \) being a hyper-parameter that decides whether to discard the previous layer's prompt when a new one is generated. \textbf{APrompt}~\cite{wang2023aprompt} introduces trainable query, key, and value prompts, denoted as \( P_q, P_k, \) and \( P_v \), into the self-attention mechanism of a Transformer encoder layer, which are integrated into the respective matrices to guide the attention computation during fine-tuning, while keeping the majority of the model parameters frozen. The new attention computations are formulated as:
\begin{equation}
    \begin{aligned}
        L(\cdot) &= \operatorname{MLP} ( \operatorname{LN} ( \operatorname{MSA} (\cdot) ) ), \\
        \operatorname{MSA}(\cdot) &= \operatorname{softmax}\left(\frac{Q^T_{\text{new}} K_{\text{new}}}{\sqrt{d}}\right) V_{\text{new}}
        \enskip,
    \end{aligned}
\end{equation}
where \( \operatorname{MLP} \) and \( \operatorname{LN} \) represent the frozen multi-layer perceptron and layer norm, \( \operatorname{MSA} \) is the multi-head self-attention module, \( Q_{\text{new}} \) is the new query matrix, \( K_{\text{new}} \) and \( V_{\text{new}} \) are the new key and value matrices augmented with attention prompts, and \( d \) is the dimension of the embeddings. \textbf{DePT} (Decomposed Prompt Tuning)~\cite{shi2023dept} decomposes a trainable soft prompt matrix \( P \in \mathbb{R}^{l \times d} \) into a shorter trainable prompt matrix \( P_s \in \mathbb{R}^{m \times d} \) and a pair of low-rank matrices \( A \in \mathbb{R}^{s \times r} \) and \( B \in \mathbb{R}^{r \times d} \), where the rank \( r \ll \min(s, d) \). These components are optimized with different learning rates \( \alpha_1 \) and \( \alpha_2 \) respectively. The updated word embedding matrix for the \(i^{\text{th}}\) sample is given by \( W'_i = W_i + BA \), where \( W_i \) is the original word embedding matrix. The loss function to be optimized is \( L_{\text{DePT}} = -\sum_{i=1}^{N} \log P(y_i | [P_s, W'_i]; \Theta) \), where \(\Theta\) represents the frozen pretrained model weights. \textbf{Xprompt}~\cite{ma2022xprompt} operates on the principle of hierarchical structured pruning to identify and retain only the most effective soft prompt tokens, denoted as \( p_i \), and their components, denoted as \( q_{i,e} \), by calculating their importance scores \( I_{p_i} \) and \( I_{q_{i,e}} \) using the following expressions:
\begin{equation}
    \begin{aligned}
        I_{p_i} &= \mathbb{E}_{x \sim D_x} \left| \frac{\partial L(x)}{\partial \gamma_i} \right|, \\
        I_{q_{i,e}} &= \mathbb{E}_{x \sim D_x} \left| \frac{\partial L(x)}{\partial \zeta_i} \right|
        \enskip,
    \end{aligned}
\end{equation}
where \( L \) is the loss function, \( D_x \) is the training data distribution, \( \gamma_i \) and \( \zeta_i \) are mask variables for token-level and piece-level pruning respectively, and the importance scores determine the contribution of each prompt token and piece to the model's performance. \textbf{InfoPrompt}~\cite{wu2024infoprompt} maximizes the mutual information between the prompt \( P \) and the parameters of the classification head \( \theta \), denoted as \( I(P; \theta|X) \), and between the prompt \( P \) and the encoded representation from the pretrained language model \( Z = \Phi(P, X) \), denoted as \( I(P; Z|X) \), by optimizing two novel loss functions, referred to as the head loss and the representation loss, respectively. \textbf{PTP} (Prompt Tuning with Perturbation-based Regularizer)~\cite{chen2023ptp} introduces perturbation-based regularizers to stabilize prompt tuning by smoothing the loss landscape. This can be formulated as:
\begin{equation}
    \min_{\theta} \mathbb{E}_{(s,y) \sim D} \left[ L \left( M \left( \theta, s + \delta, y \right) \right) \right]
    \enskip,
\end{equation}
where \( \delta \) is the perturbation sampled from either a Gaussian distribution (\( \delta \sim \mathcal{N} \) for PTP-RN) or generated by an adversarial attack algorithm (\( \delta = \arg\max_{\| \delta \| \leq \epsilon} L \left( \theta, s + \delta, y \right) \) for PTP-ADV). \(s\) is the input sequence, \(y\) is its label, \(M\) is the large language model, \(\theta\) represents the trainable prompt parameters, and \(L\) is the loss function.

%% file: Scale_and_Shift.tex
\begin{figure*}[htbp]
    \centering
    \includegraphics[width=\textwidth]{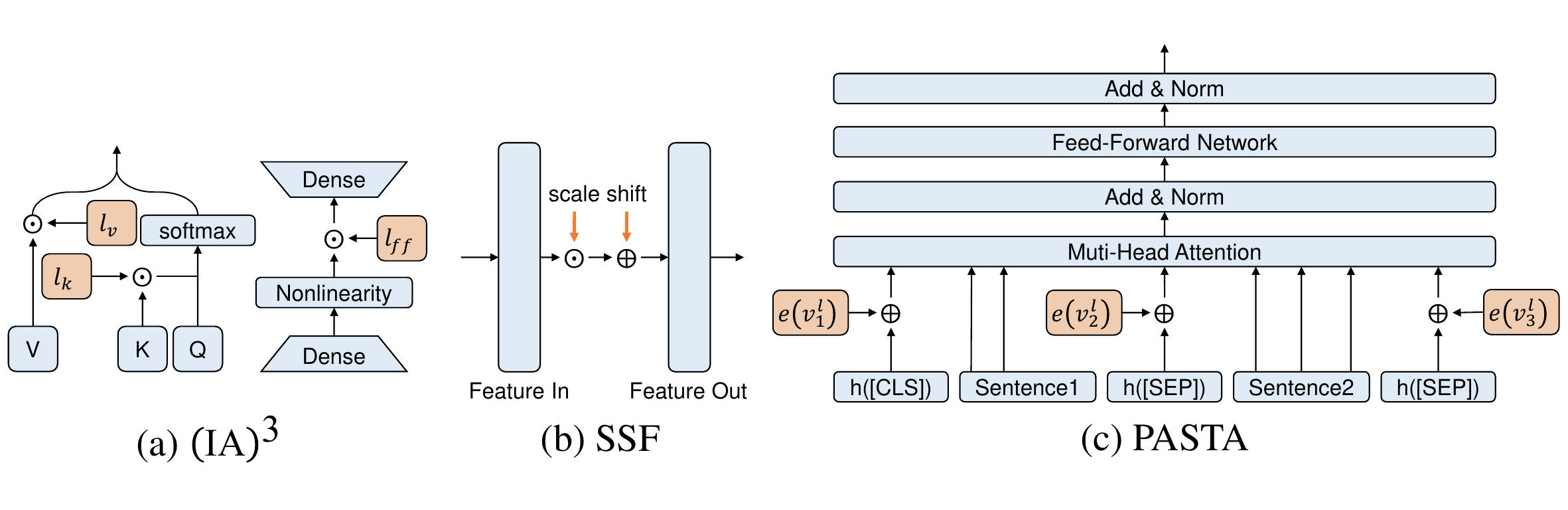} 
    \caption{Illustration of three representative scale and shift algorithms.} 
    \label{fig:scale_and_shift_adapter} 
\end{figure*}
$\textbf{(IA)}^{3}$ (Infused Adapter by Inhibiting and Amplifying Inner Activations)~\cite{liu2022few} shown in Figure~\ref{fig:scale_and_shift_adapter} (a) is a PEFT method for scaling inner activations of a model by learned vectors. For a decoder with \( L \) layers, $\text{(IA)}^{3}$ adds scaling vectors \( l_k, l_v, \) and \( l_{ff} \) (initialized as ones) to scale key, value, and feed-forward activations, respectively. This allows for task-specific adaptations while updating a tiny fraction (\(\leq 0.01\%\)) of the model's parameters, facilitating mixed-task batches. The method can be applied permanently to weight matrices if the model is dedicated to a single task, avoiding extra computations. \textbf{MoV} (Mixture of Vectors)~\cite{zadouri2023pushing} introduces a parameter-efficient Mixture of Experts (MoE) architecture that updates only lightweight experts, less than 1\% of an 11B parameter model. It generalizes well to unseen tasks. Computation is routed with soft merging: \(E_{\text{mix}} = \sum_{i=1}^{n} s_i \cdot E_i\); \(y = E_{\text{mix}}(x)\), where \(E_i\) represents each expert, \(s_i\) is the gating weight for each expert, and \(x\) is the input. This approach ensures robust performance under strict parameter constraints. \textbf{SSF}~\cite{lian2022scaling} shown in Figure~\ref{fig:scale_and_shift_adapter} (b) modifies deep features extracted by a pre-trained model through linear transformations to match the distribution of the target dataset. Given an input \( x \in \mathbb{R}^{(N^2+1) \times d} \), the output \( y \) is computed as:
\begin{equation}
    y = [\gamma \odot x + \beta]^T
    \enskip,
\end{equation}
where \( \gamma \) and \( \beta \) are learnable scale and shift parameters, respectively, and \( \odot \) denotes element-wise multiplication. This approach requires tuning far fewer parameters than full fine-tuning. \textbf{PASTA} (PArameter-efficient tuning with Special Token Adaptation)~\cite{yang2023parameter}, as illustrated in Figure~\ref{fig:scale_and_shift_adapter} (c), modifies special token representations in pretrained models. For the \( l^{\text{th}} \) Transformer layer, given input \( H^{(l)} = \{h_i^{(l)}\}_{i=1}^N \), where \( h_i^{(l)} \in \mathbb{R}^d \), PASTA updates the input as \( H_{\text{mod}}^{(l)} = \{h_i^{(l)} + m_i^{(l)}\}_{i=1}^N \), where \( m_i^{(l)} \) is defined as:
\begin{equation}
    m_i^{(l)} = 
\begin{cases} 
0 & \text{if } i \text{ is not a special token} \\
e(v_p^{(l)}) & \text{if } i \text{ is the } p\text{-th special token}
\end{cases}
\enskip,
\end{equation}
with \( e(v_p^{(l)}) \in \mathbb{R}^d \) being the trainable vector for the \( p \)-th special token at layer \( l \).

%% file: Others.tex
\textbf{IPA} (Inference-time Policy Adapters)~\cite{lu2023inference} tailors LLMs to specific objectives without fine-tuning. IPA combines the output distribution of a base LLM with a smaller, trainable adapter policy. The adapter is optimized via reinforcement learning (RL) to align the LLM's output with user-defined goals. At inference, the base model's distribution and the trained adapter's distribution are merged for decoding as follows:

\begin{equation}
\begin{split}
   p_{\text{combined}}(\text{output} \mid \text{input}) &= \alpha p_{\text{base}}(\text{output} \mid \text{input})
    \quad + \\ &(1-\alpha) p_{\text{adapter}}(\text{output} \mid \text{input})
    \enskip,
\end{split}
\end{equation}

where \( p_{\text{base}} \) is the base model's probability distribution, \( p_{\text{adapter}} \) is the adapter's distribution, and \( \alpha \) controls their mixture. \textbf{LST} (Ladder Side-Tuning)~\cite{sung2022lst} introduces a side network that predicts outputs using shortcuts (ladders) from a pre-trained backbone, avoiding backpropagation through the entire backbone. Formally, given a backbone \( f_N(f_{N-1}(\ldots f_2(f_1(x)) \ldots)) \), the side network \( g \) takes intermediate activations \( z_i \) as inputs, where \( z_i = f_i(x) \). The final output \( \hat{y} \) is computed by \( g(z_i; \theta_g) \), significantly reducing memory cost. Here, \( x \) is the input, \( f_i \) represents the \( i \)-th layer function, and \( \theta_g \) are the parameters of the side network. \textbf{Attention-Fusion}~\cite{cao2022attention} aggregates intermediate layer representations from a pre-trained model to compute task-specific token representations. This module trains only \(0.0009\%\) of total parameters and achieves competitive performance to full fine-tuning. Formally, given a pre-trained model with \(L\) layers, the output \(\mathbf{h}^{(l)}_i\) of each layer \(l\) for token \(i\) is used to compute a weighted sum \(\mathbf{r}_i = \sum_{l=1}^{L} \alpha^{(l)}_i \mathbf{h}^{(l)}_i\), where \(\alpha^{(l)}_i\) represents the attention weight for layer \(l\) on token \(i\).

%% file: Reparameterized_PEFT.tex
Reparameterization is a technique for improving the training efficiency and performance of a model by transforming its parameters. In the context of PEFT, the transformation involves low-rank parameterization, which entails constructing a low-rank learnable parameter matrix to adapt to specific downstream tasks. During training, only the low-rank parameter matrix is fine-tuned, and at inference time, the learned matrix is combined with the pre-trained parameters to ensure that inference speed is not affected.
\subsubsection{Low-rank Decomposition}
\label{low_rank_decomposition}
\input{Low-rank_Decomposition}

\subsubsection{LoRA Derivatives}
\label{lora_derivatives}
\input{Dynamic_Rank}

\input{LoRA_Improvement}

%% file: Low-rank_Decomposition.tex
\textbf{LoRA} (Low-rank Adaptation)~\cite{hu2021lora} introduces low-rank trainable matrices \( A \in \mathbb{R}^{d \times r} \) and \( B \in \mathbb{R}^{r \times k} \) to update the pre-trained weight matrix \( W_0 \in \mathbb{R}^{d \times k} \) via \( \Delta W = B A \), where \( W = W_0 + \Delta W \) is used for inference without additional latency. \textbf{KronA}~\cite{edalati2022krona} is a Kronecker product-based adapter module for efficient fine-tuning of Transformer-based pre-trained language models (PLMs). The tuned weight matrix \( W_{\text{tuned}} \) is computed as the original PLM weight matrix \( W \) plus a scaled Kronecker product of two learnable matrices \( A_k \) and \( B_k \): 
\begin{equation}
    W_{\text{tuned}} = W + s[A_k \otimes B_k]
    \enskip,
\end{equation}
where \( s \) is a scaling factor, and \( \otimes \) denotes the Kronecker product operator.

%% file: Dynamic_Rank.tex
\paragraph{Dynamic Rank}
\begin{figure*}[htbp]
    \centering
    \includegraphics[width=\textwidth]{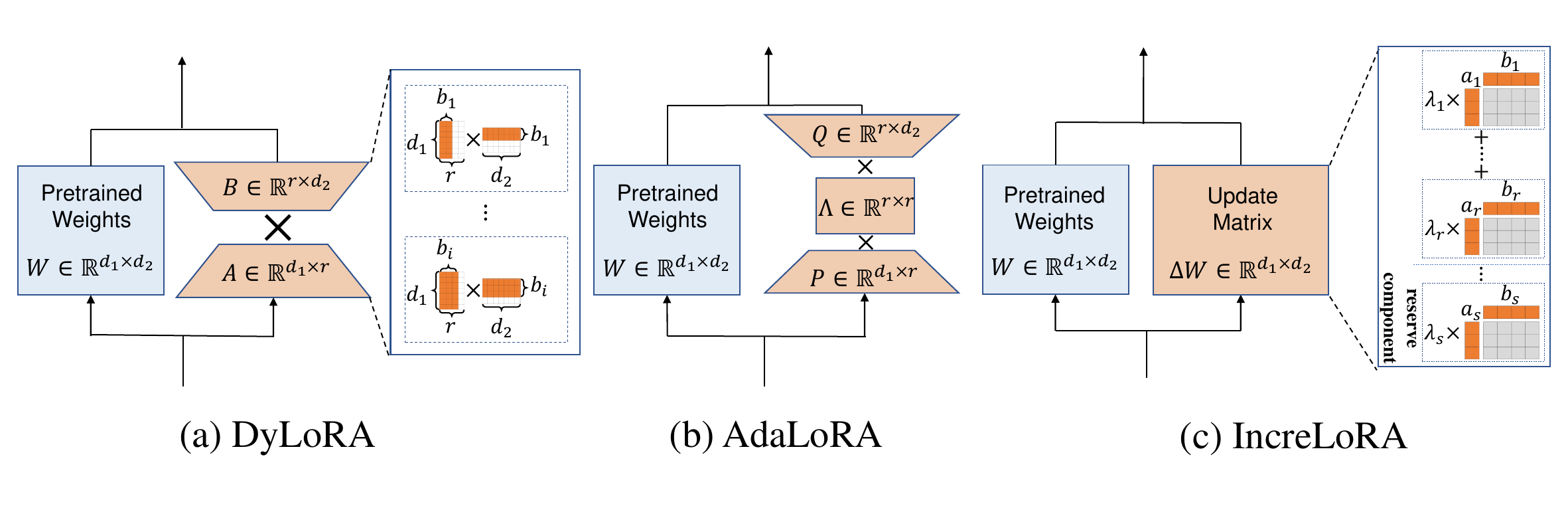} 
    \caption{Illustration of three representative dynamic rank methods in LoRA.} 
    \label{fig:dynamic_rank_lora} 
\end{figure*}
\textbf{DyLoRA}~\cite{valipour2023dylora} shown in Figure~\ref{fig:dynamic_rank_lora} (a) introduces a dynamic low-rank adaptation technique by training Low-Rank Adapter (LoRA) blocks for a range of ranks during training, where the representation learned by the adapter module is sorted at different ranks, enabling the model to be flexible and perform well across a wider range of ranks without additional training time or the need for rank selection. \textbf{AdaLoRA}~\cite{zhang2023adalora} illustrated in Figure~\ref{fig:dynamic_rank_lora} (b) dynamically allocates the budget among weight matrices based on their importance scores, where incremental updates are parameterized in the form of a singular value decomposition as \( W = W_{0} + P \Lambda Q \), with \( P \in \mathbb{R}^{d_1 \times r} \), \( Q \in \mathbb{R}^{r \times d_2} \), and \( \Lambda \in \mathbb{R}^{r \times r} \) being the left singular vectors, right singular vectors and singular values, respectively. \textbf{IncreLoRA}~\cite{zhang2023increlora} presented in Figure~\ref{fig:dynamic_rank_lora} (c) incrementally allocates trainable parameters during the training process based on the importance scores of each module, which is formulated as follows: 
\begin{equation}
    W = W_{0} + \sum_{i=1}^{r} \lambda_i w_i = W_{0} + \sum_{i=1}^{r} \lambda_i b_i a_i
    \enskip,
\end{equation}
where \( W_{0} \) is the pretrained weight matrix, \( r \ll \min(in, out) \), \( w_i \) is a rank-1 matrix, \( a_i \in \mathbb{R}^{in} \), \( b_i \in \mathbb{R}^{out} \), and \( \lambda_i \) is a scaling factor updated through backpropagation, with \( \lambda_i \) initialized to zero to ensure the initial update matrix is zero. \textbf{SoRA} (Sparse low-rank Adaption)~\cite{ding2023sparse} introduces a gate unit, optimized with a proximal gradient method to control the sparsity of the LoRA's low-rank matrices. The gate unit enables dynamic adjustment of the rank of LoRA during training, enhancing representation power while maintaining parameter efficiency. During inference, blocks corresponding to zero entries in the gate unit are eliminated, reducing the SoRA module to a concise, rank-optimal LoRA.

%% file: LoRA_Improvement.tex
\paragraph{LoRA Improvement}
\textbf{LoRA+}~\cite{hayou2024lora+} introduces a novel technique by applying different learning rates to the down- and up-projection matrices $A$ and $B$: $\eta_{B} = \lambda \eta_{A}$, where $\lambda$ is a fixed value greater than 1, focusing on tuning $\eta_{A}$ for enhanced model adaptability. Designed to mitigate the significant memory requirements for activations that are intrinsic to LoRA, \textbf{LoRA-FA} (Low-Rank Adaptation with Frozen-A)~\cite{zhang2023lora} freezes the pre-trained weight \( W \) and the projection-down weight \( A \), and only update the projection-up weight \( B \) during the fine-tuning process, which results in a model weight change \( \Delta W \) that resides in a low-rank space defined by the column space of \( A \). The method is designed to reduce the activation memory footprint without incurring additional computational overhead. \textbf{DoRA} (Weight-Decomposed Low-Rank Adaption)~\cite{liu2024dora} aims to bridge the gap in performance between LoRA and full fine-tuning (FT) by leveraging a novel weight decomposition approach. It decomposes the pre-trained weight matrix $W_0 \in \mathbb{R}^{d \times k}$ into \textit{magnitude} and \textit{direction}. During fine-tuning, only the \textit{direction} component is updated using a low-rank approximation $\Delta W = BA$, where $B \in \mathbb{R}^{d \times r}$ and $A \in \mathbb{R}^{r \times k}$, and $r \ll \min(d, k)$. Here, $r$ denotes the rank of the low-rank approximation, $d$ and $k$ represent the dimensions of the weight matrix. This allows for efficient parameter updates while preserving the original weight's magnitude, enhancing learning capacity and stability. \textbf{Laplace-LoRA}~\cite{yang2023bayesian} introduces a Bayesian approach to LoRA for fine-tuning LLMs. It addresses the issue of overconfidence in fine-tuned LLMs by estimating predictive uncertainty. Laplace-LoRA approximates the posterior distribution over LoRA parameters using a Laplace approximation, leading to better-calibrated models. Mathematically, given a maximum a posteriori (MAP) estimate $\theta_{\text{MAP}}$, the predictive distribution for a new input $x^*$ is approximated as:
\begin{equation}
    f_{\theta}(x^*) \sim \mathcal{N}\left(f_{\theta_{\text{MAP}}}(x^*), \Lambda\right)
    \enskip,
\end{equation}
where
\(
\Lambda = (\nabla_{\theta}f_{\theta}(x^*)|_{\theta=\theta_{\text{MAP}}}) \Sigma (\nabla_{\theta}f_{\theta}(x^*)|_{\theta=\theta_{\text{MAP}}})^\top.
\)
Here, \(\nabla_{\theta}f_{\theta}(x^*)\) represents the gradient of the prediction with respect to the parameters, and \(\Sigma\) is the covariance matrix of the Laplace approximation. The prior precision \(\lambda\) is optimized using the Laplace marginal likelihood on the training dataset:
\begin{equation}
    P(y|X) \approx \exp(L(y, X; \theta_{\text{MAP}})) (2\pi)^{D/2} |\Sigma|^{1/2}
    \enskip,
\end{equation}
Samples from the predictive distribution are obtained by:
\begin{equation}
    \tilde{f}_{\theta}(x^*) = f_{\theta_{\text{MAP}}}(x^*) + L\xi,
\end{equation}
where \(L\) is the Cholesky factor of \(\Lambda\) and \(\xi\) is a vector of independent standard normal random variables. This method improves calibration without requiring a separate validation set, making it suitable for small datasets. \textbf{PeriodicLoRA} (PLoRA)~\cite{meng2024periodiclora} enhances LoRA's learning capacity by periodically accumulating low-rank updates to form a higher-rank matrix. During each stage, only LoRA weights \(W_{\text{LoRA}}\) are updated. At the end of each stage, \(W_{\text{LoRA}}\) is unloaded into the backbone parameters \(W_{\text{backbone}}\), i.e., \(W_{\text{backbone}} \leftarrow W_{\text{backbone}} + \Delta W_{\text{LoRA}}\), and then \(W_{\text{LoRA}}\) is reinitialized. This increases the effective update rank without additional memory cost. \textbf{HydraLoRA}~\cite{tian2024hydralora} enhances LoRA by adopting an asymmetric structure for efficient fine-tuning. It segments the LoRA into multiple ``intrinsic components," each with a distinct matrix \( B_k \), sharing a common matrix \( A \). The update formula is given by:
\begin{equation}
    \alpha W = W_0 + r \sum_{k=1}^{N} A B_k
    \enskip,
\end{equation}
where \( W_0 \) is the original weight matrix, \( r \) is a scaling factor, \( A \) and \( B_k \) are low-rank matrices, and \( N \) is the number of components. A trainable MoE router dynamically allocates samples to these components for fine-tuning. \textbf{AFLoRA}~\cite{liu2024aflora} incrementally freezing trainable low-rank matrices based on a novel freezing score, computed using smoothed gradient \( \bar{I}(t)_{A_l} \), uncertainty tensor \( \bar{U}(t)_{A_l} \), and their Hadamard product to determine the stability of weights throughout training, as described by the equations:
\begin{equation}
    \begin{aligned}
        I(t)_{A_l} &= |\nabla L(\theta)|, \\
        \bar{I}(t)_{A_l} &= \beta_1 \bar{I}(t-1)_{A_l} + (1 - \beta_1) I(t)_{A_l}, \\
        U(t)_{A_l} &= | I(t)_{A_l} - \bar{I}(t)_{A_l} |, \\
        \bar{U}(t)_{A_l} &= \beta_2 \bar{U}(t-1)_{A_l} + (1 - \beta_2) U(t)_{A_l}, \\
        s(t)_{A_l} &= \text{mean}(\bar{I}(t)_{A_l} \odot \bar{U}(t)_{A_l})
        \enskip,
    \end{aligned}
\end{equation}
where \( A_l \) represents the low-rank tensor, \( L(\theta) \) is the loss function, \( \beta_1 \) and \( \beta_2 \) are smoothing factors, and \( t \) denotes the current training step. \textbf{LoRA-SP}~\cite{wu2024lora} selectively freezes half of the parameters in the matrices \( A \) and \( B \) during fine-tuning, with the adapted weight matrix \( \Delta W \) calculated as \( \Delta W = (A \odot S)(B \odot S)^\top \), where \( S \) is a binary selection matrix that determines which parameters to update or freeze, and \( \odot \) denotes element-wise multiplication. \textbf{SuperLoRA}~\cite{chen2024superlora} generalizes LoRA approach by jointly adapting all weight updates \( \Delta W \) across layers through a high-order tensor decomposition, where \( \Delta W_{\text{group}_g} \) is computed as 
\begin{equation}
    F(\Delta W_{\text{lora}_g}) = F \left( \bigotimes_{k=1}^{K} \left(C_{gk} \prod_{m=1}^{M}~_{\times m} A_{gkm}\right) \right) 
    \enskip,
\end{equation}
with \( F \) being a projection function, \( M \) the order of tensor modes, \( K \) the number of Kronecker splits, \( C_{gk} \) the core tensor, \( A_{gkm} \) the plane factors, \(\prod_{m=1}^{M}~_{\times m}\) the tensor products from model-$1$ to model-$M$, and \(\bigotimes\) the Kronecker product.

%% file: Subtractive_PEFT.tex
Contrary to Additive PEFT, Selective PEFT selects a very small subset of the pre-trained model's parameters for fine-tuning to adapt to specific downstream tasks through a parameter masking matrix. Depending on the way the parameters are masked, Selective PEFT can be divided into unstructured masking and structured masking.
\subsubsection{Unstructural Masking}
\label{Unstructural_Masking}
\input{Unstructural_Masking}

\subsubsection{Structural Masking}
\label{Structural_Masking}
\input{Structural_Masking}

%% file: Unstructural_Masking.tex
\begin{figure}[htbp]
    \centering
    \includegraphics[width=\columnwidth]{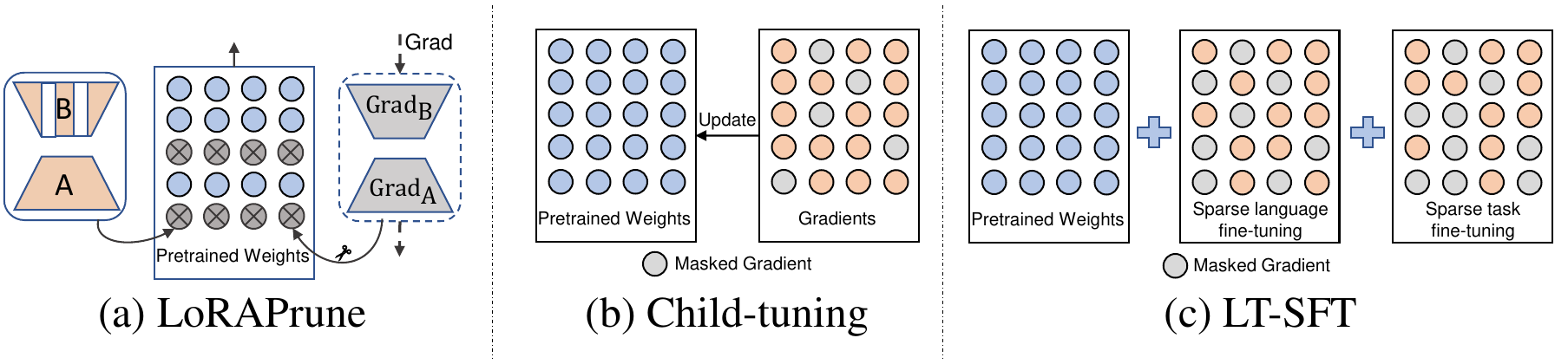} 
    \caption{Illustration of three representative unstructural masking methods} 
    \label{fig:unstructural_masking}
\end{figure}
\textbf{U-Diff pruning}~\cite{guo2020parameter} introduces a task-specific ``diff" vector $\delta_\tau$ that is added to pretrained model parameters $\theta$. The task-specific parameters are defined as $\theta_\tau = \theta + \delta_\tau$. During training, $\delta_\tau$ is adaptively pruned using a differentiable $L_0$-norm approximation to encourage sparsity. $\theta$ remains fixed. This method enables efficient transfer learning, modifying only a small fraction of the parameters per task. \textbf{U-Bitfit}~\cite{lawton2023neural} determines which components of the bias update vector \( \Delta b \) should be zero or non-zero, based on a first-order approximation of the change in training loss from pruning a bias parameter \( \theta \), calculated as \( -\theta \cdot \frac{\partial L}{\partial \theta} \). \textbf{PaFi}~\cite{liao2023parameter} generates a universal sparse mask for parameter selection without training. PaFi identifies the least significant pre-trained parameters by their magnitude and fine-tuning only those, represented as selecting parameters \( \theta_i \) where \( |\theta_i| \leq \text{sort}(|\theta|)_k \) for the mask \( m \). \textbf{FishMask}~\cite{sung2021training} precomputes a fixed sparse mask for neural network parameters, selecting the top \(k\) parameters based on their Fisher information to be updated during training. This ``FISH (Fisher-Induced Sparse uncHanging) mask" enables efficient training by updating only a subset of parameters, which reduces memory and communication costs compared to full model updates. \(k\) represents the number of parameters to be selected for updates, and Fisher information measures parameter importance for the given task. \textbf{Fish-Dip}~\cite{das2023unified} dynamically updates the importance of model parameters for fine-tuning based on feedback from the most regressing samples, using the empirical Fisher information to create a sparsity mask that focuses training on a subset of parameters, as denoted by the equation:
\begin{equation}
    \hat{F}_\theta \approx \frac{1}{n} \sum_{\{(x_i,y_i)| L_{tr}(x_i, y_i) \in \text{top}_n \}} \left(\frac{\partial \log p_\theta(y_i|x_i)}{\partial \theta}\right)^2 
    \enskip,
\end{equation}
where \( \hat{F}_\theta \) represents the empirical Fisher information, \( n \) is the number of most regressing training examples, \( p_\theta(y_i|x_i) \) is the output probability for the given input \( x_i \) and parameters \( \theta \), and the sum is taken over the top \( n \) regressing examples as determined by their loss \( L_{tr} \) during training. \textbf{LT-SFT} (see Figure~\ref{fig:unstructural_masking} (c))~\cite{ansell2021composable} introduces a composable sparse fine-tuning method for cross-lingual transfer learning. It learns sparse, real-valued masks based on a variant of the Lottery Ticket Hypothesis (LTH). Task-specific masks are derived from supervised data in the source language, while language-specific masks are obtained through masked language modeling in the target language. These masks are composed with the pre-trained model to enable zero-shot cross-lingual transfer. The sparsity of the masks reduces parameter overlap and interference, improving modularity and preventing overfitting. \textbf{SAM} (Second-order Approximation Method)~\cite{fu2023effectiveness} approximates the original optimization problem using a second-order Taylor expansion to make it analytically solvable, and directly determines the parameters to optimize by solving the approximation function, which is formulated as:
\begin{equation}
    \min_{\Delta\theta} \left[ L(\theta_0) + \nabla L(\theta_0)^T M \Delta\theta + \frac{1}{2} (M\Delta\theta)^T H M \Delta\theta \right]
    \enskip,
\end{equation}
subject to \( \|M\|_0 = \lfloor mp \rfloor; M_{ij} = 0, \forall i \neq j; M_{ii} \in \{0, 1\} \), where \( \theta_0 \) are the pre-trained parameters, \( \Delta\theta \) is the difference vector, \( M \) is the parameter mask matrix, \( L \) is the loss function, \( \nabla L(\theta_0) \) is the gradient of the loss function at \( \theta_0 \), and \( H \) is an approximated diagonal Hessian matrix. \textbf{Child-tuning} (see Figure~\ref{fig:unstructural_masking} (b))~\cite{xu2021raise} updates only a subset of parameters, referred to as the child network, during fine-tuning while masking out the gradients of the remaining parameters in the backward pass, which can be formulated as:
\begin{equation}
    w_{t+1} = w_t - \eta \odot \frac{\partial L(w_t)}{\partial w_t} \odot M_t
    \enskip,
\end{equation}
where \( w_t \) represents the model parameters at the \( t^{\text{th}} \) iteration, \( \eta \) is the learning rate, \( L(w_t) \) is the loss function, and \( M_t \) is a 0-1 mask indicating the child network. \textbf{U-MAM}~\cite{lawton2023neural} is an unstructured neural architecture search approach for parameter-efficient tuning of large pre-trained language models. It involves pruning a dense low-rank update from an initial parameter-efficient tuning architecture to find an efficient subset of parameters to fine-tune. \textbf{Threshold-Mask}~\cite{zhao2020masking} learns selective binary masks for pre-trained language model weights without fine-tuning, where each linear layer \( W_l \) is associated with a real-valued matrix \( M_l \) initialized randomly, and a binary mask \( M_l^{bin} \) is obtained by applying a thresholding function, used to select important weights: \( (m_l^{bin})_{i,j} = 1 (m_{l,i,j} \geq \tau) \) with \( m_{l, i, j} \in M_{l}\) and the global thresholding hyperparamter \(\tau\), and the masked weights are computed as \( \hat{W}_l = W_l \odot M_l^{bin} \), with \( M_l \) updated during training via the straight-through estimator: \( M_l \leftarrow M_l - \eta \frac{\partial L(\hat{W}_l)}{\partial M_l^{bin}} \). \textbf{LoRAPrune} (see Figure~\ref{fig:unstructural_masking} (a))~\cite{zhang2023pruning} approximates the importance of each parameter in the pre-trained model weights \( W_0 \) by utilizing the gradients of the low-rank matrices \( A \) and \( B \), which are then used to perform structured pruning in an iterative and progressive manner, efficiently reducing the model's size while maintaining performance.

%% file: Structural_Masking.tex
\textbf{S-Diff pruning}~\cite{guo2020parameter} introduces a structured pruning strategy by dividing the weight parameters into local groups and strategically removing them collectively. \textbf{S-Bitfit}~\cite{lawton2023neural} selects whether to update each bias parameter \( b \) with a learned update \( \Delta b \), where the decision is based on a pruning criterion that sums the first-order approximation of the loss change over the entire bias update \( \Delta b \), expressed as \( -\sum_{\theta \in \Delta b} \theta \cdot \frac{\partial L}{\partial \theta} \). \textbf{FAR} (Freeze And Reconfigure)~\cite{vucetic2022efficient} leverages overparameterization in BERT-like models to efficiently fine-tune them on resource-constrained devices. FAR selectively updates parameters based on their importance, determined through priming, while freezing others. This reduces memory usage and fine-tuning time, with minimal impact on performance. Notation-wise, if \( P \) represents the total parameters, \( P_{\text{frozen}} \subset P \) denotes frozen parameters, and \( P_{\text{active}} = P \setminus P_{\text{frozen}} \) are active parameters updated during fine-tuning. \( P_{\text{frozen}} \) is selected using priming to ensure optimal performance. \textbf{BitFit}~\cite{zaken2021bitfit} modifies only the bias terms of a pre-trained BERT model, demonstrating competitive performance with full fine-tuning on small to medium datasets and practical utility for deploying multi-task models in memory-constrained environments. \textbf{Xattn Tuning}~\cite{gheini2021cross} updates only cross-attention parameters in Transformer models for machine translation, showing it can achieve near-equivalent performance to fine-tuning the entire model, while also leading to crosslingually aligned embeddings that can mitigate catastrophic forgetting and enable zero-shot translation capabilities. \textbf{SPT}~\cite{he2023sensitivity} identifies task-specific sensitive parameters by measuring their impact on loss reduction, denoted as \( s_n \), and then adaptively allocates trainable parameters to these positions under a given budget \( \tau \), utilizing both unstructured tuning for individual parameters and structured tuning for weight matrices with a high number of sensitive parameters, as indicated by \( \sigma_{\text{opt}} \). \textbf{S-MAM}~\cite{lawton2023neural} is a structured neural architecture search approach for parameter-efficient tuning of large pre-trained language models. It selects and fine-tunes a fixed rank of parameters within the model's attention mechanisms and feed-forward networks.

%% file: Hybrid_PEFT.tex
\begin{figure}[htbp]
    \centering
    \includegraphics[width=\columnwidth]{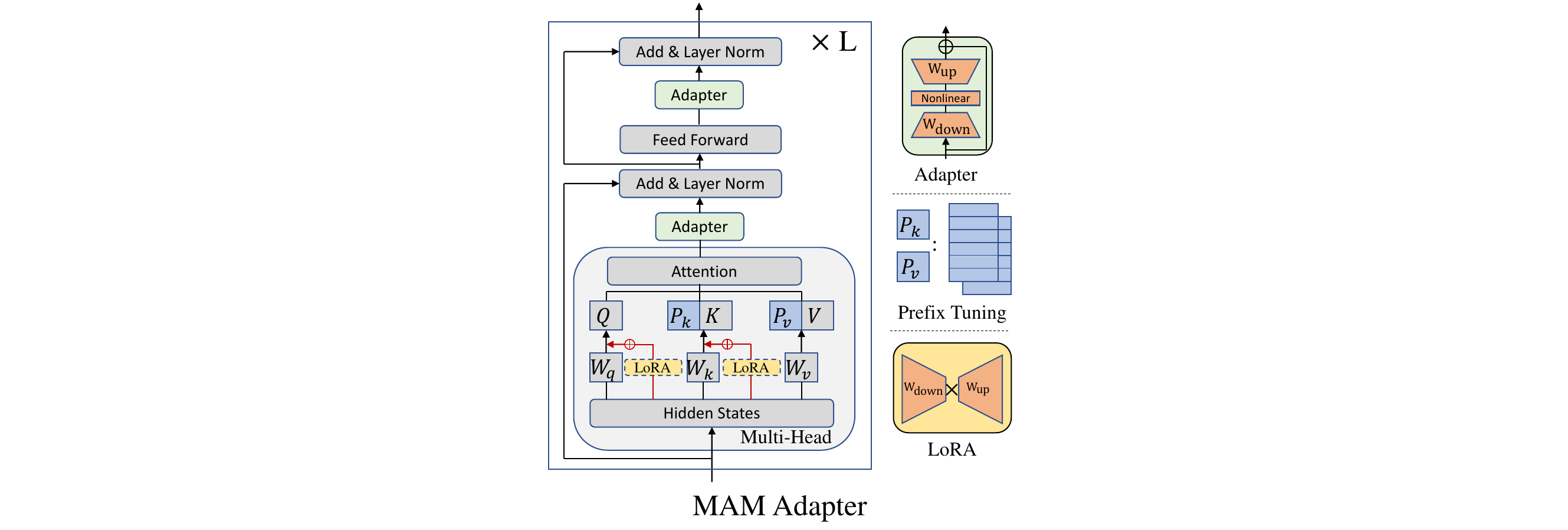} 
    \caption{Illustration the principles of MAM-Adapter, a representative hybrid PEFT method} 
    \label{fig:hybrid_mam_adapter}
\end{figure}

Due to the significant performance differences of different types of PEFT methods on various tasks, many studies aim to enhance model performance by combining the advantages of different types of PEFT methods. These research efforts are summarized as Hybrid PEFT methods. A representative hybrid PEFT method, known as MAM-Adapter, is illustrated in Figure~\ref{fig:hybrid_mam_adapter}.

\textbf{UniPELT}~\cite{mao2021unipelt} operates on the principle of dynamically activating the most suitable parameter-efficient language model tuning (PELT) submodules for a given task through a gating mechanism, which is mathematically represented as \( h'_{A} = G_{A}h_{A} + h_{F} \), where \( h'_{A} \) is the final output, \( h_{A} \) is the output of the adapter submodule, \( h_{F} \) is the direct input to the adapter, and \( G_{A} \) is the gating function that modulates the contribution of the adapter submodule based on the specific data and task setup. \textbf{S4}~\cite{chen2023parameter} discovers design patterns by grouping layers in a spindle pattern, uniformly allocating trainable parameters, tuning all groups, and assigning tailored strategies to different groups, consistently outperforming existing fine-tuning strategies across various NLP tasks and models. \textbf{MAM Adapter}~\cite{he2021towards} is a unified framework for parameter-efficient transfer learning methods by reframing them as modifications to specific hidden states in pretrained models, which can be mathematically represented as \( h \leftarrow (1 - \lambda(x))h + \lambda(x)\Delta h \), where \( h \) is the original hidden representation, \( \lambda(x) \) is a gating scalar, and \( \Delta h \) is the modification vector computed by a function \( f \) applied to the input \( x \). \textbf{LLM-Adapters}~\cite{hu2023llm} discusses the use of different adapters such as Series Adapters, Parallel Adapters, and LoRA (Low-Rank Adaptation), which are incorporated into the model's architecture at optimal locations. \textbf{NOAH}~\cite{zhang2022neural} employs neural architecture search to automatically design optimal "prompt modules" for large vision models, tailored to each downstream dataset, enhancing transfer learning, few-shot learning, and domain generalization. \textbf{AUTOPEFT}~\cite{zhou2024autopeft} automates the configuration selection for PEFT of large pre-trained language models. It employs a multi-objective Bayesian optimization approach to discover a set of Pareto-optimal configurations that balance task performance with parameter efficiency, significantly outperforming existing PEFT methods with minimal training costs. \textbf{$\text{S}^{3}\text{Delta-M}$}~\cite{hu2022sparse} automatically searches for an optimal trainable structure within pre-trained models by using a unified framework of various Delta Tuning methods. It employs bi-level optimization and a shifted global sigmoid function to control sparsity, achieving high performance with minimal trainable parameters. \textbf{ProPETL}~\cite{zeng2023one} enables the sharing of a single prototype network across different layers and tasks, with binary masks learned to prune sub-networks, significantly reducing parameter storage while improving efficiency and performance over other methods.

%% file: Quantization_PEFT.tex
Quantization is another widely used and studied technique aimed at improving computational efficiency and reducing memory usage. We summarize the PEFT methods that use and research quantization technology, as Quantization PEFT.
\begin{figure*}[htbp]
    \centering
    \includegraphics[width=\textwidth]{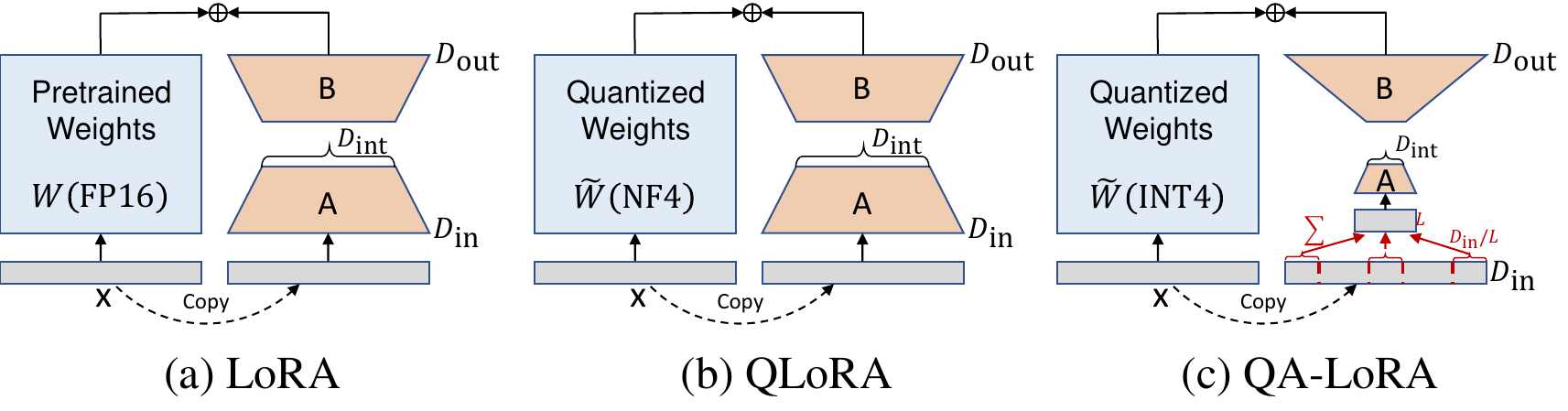} 
    \caption{Illustration of the difference among LoRA, QLoRA and QA-LoRA} 
    \label{fig:quantization_lora}
\end{figure*}
\textbf{BI-Adapter}~\cite{jie2023revisiting} introduces a novel method for low-precision adapter training in vision models. It utilizes the observation that adapter parameters converge to flat minima, suggesting robustness to precision reduction. The method employs a quantization-aware training strategy, minimizing the quantization error by clustering weight parameters into Gaussian distributions. Specifically, weights \( w \) are standardized \( w' = \frac{w - \mu}{\sigma} \), quantized, and then de-standardized to backpropagate gradients effectively. This approach significantly reduces model size with minimal impact on performance, addressing storage and transmission inefficiencies in multi-task learning. \textbf{PEQA}~\cite{kim2024memory} involves a two-step process: first, decomposing the parameter matrix of each fully-connected layer into a low-bit integer matrix and quantization scales, and second, fine-tuning only the quantization scale while keeping the integer matrix frozen, which can be mathematically represented as:
\begin{equation}
    \tilde{W} = (s_0 + \Delta s) \cdot \left(\operatorname{clamp}\left(\left \lfloor \frac{W_0}{s_0} \right \rceil + z_0, 0, 2^b - 1\right) - z_0 \right)
    \enskip,
\end{equation}
where the notation \(A \cdot B\) denotes the element-wise product of matrices \(A\) and \(B\). The symbol \(\lfloor \cdot \rceil\) represents the rounding function, which rounds its argument to the nearest integer. The function \(\operatorname{clamp}(\cdot, a, b)\) signifies the clamping operation that constrains its input within the range \([a, b]\). Here, \(W_0\) denotes the original weight matrix, \(s_0\) represents the initial scale factor, and \(z_0\) is the zero-point value. The variable \(\Delta s \in \mathbb{R}^{n \times 1}\) signifies the gradient update of \(s_0\), obtained through adaptation to a downstream task, and \(b\) indicates the bit-width. \textbf{QLORA}~\cite{dettmers2024qlora}, a quantized version of LoRA, utilizes 4-bit NormalFloat (NF4) precision for quantizing pretrained models, enhanced by double quantization and a paged optimizer to prevent the gradient checkpointing memory spikes. The NF4 is an information theoretically optimal quantization data type for normally distributed data, delivering enhanced empirical performance over 4-bit Integer and Float representations. While QLoRA converts the FP16 pretrained weights $W$ to the NF4 precision to enable LLM finetuning on a reduced number of GPUs, the auxiliary weights of the LoRA matrix re-quantize the final weights back to FP16 post-finetuning. Therefore, \textbf{QA-LoRA} (Quantization-Aware Low-Rank Adaptation)~\cite{xu2023qa} addresses the imbalance between quantization and adaptation by employing group-wise operations, which increase the flexibility of low-bit quantization while reducing that of the adaptation process. The algorithm is straightforward to implement and provides two key benefits: during fine-tuning, LLM weights are quantized (e.g., to $INT4$) to conserve time and memory; post fine-tuning, the LLM and auxiliary weights are seamlessly integrated into a quantized model without accuracy loss. 
The comparative analysis and conceptual distinctions among LoRA, QLoRA, and QA-LoRA methodologies are visually illustrated in Figure~\ref{fig:quantization_lora}. \textbf{LoftQ}~\cite{li2023loftq} introduces a simultaneous process of quantizing an LLM and initializing LoRA with low-rank matrices to mitigate performance gaps. The algorithm approximates the original weights $W \in \mathbb{R}^{d_1 \times d_2}$ with a quantized version $Q \in \mathbb{R}^{d_1 \times d_2}_N$ and low-rank matrices $A \in \mathbb{R}^{d_1 \times r}$ and $B \in \mathbb{R}^{d_2 \times r}$, minimizing the Frobenius norm $\|W - Q - AB^\top\|_F$. LoftQ alternates between quantization and SVD, efficiently approximating the original weights for improved downstream task performance, especially in 2-bit and 2/4-bit mixed precision scenarios. \textbf{LQ-LoRA}~\cite{guo2023lq} iteratively decomposes a pretrained matrix \( W \) into a quantized component \( Q \) and a low-rank component \( L_1L_2 \) by solving the optimization problem:
\begin{equation}
    \arg \min_{Q, L_1, L_2} \| W - (Q + L_1L_2) \|_F 
    \enskip,
\end{equation}
where \( Q \) is fixed during finetuning and only \( L_1 \) and \( L_2 \) are updated. \textbf{QDyLoRA}~\cite{rajabzadeh2024qdylora} is a quantized dynamic low-rank adaptation technique for efficient tuning of large language models. It builds upon the DyLoRA~\cite{valipour2023dylora} method, which enables training across a spectrum of ranks dynamically, and combines it with quantization techniques from QLoRA~\cite{dettmers2024qlora}. The core principle is to allow the model to finetune on a set of predefined ranks and then select the optimal rank for inference, achieving efficiency without compromising performance. Mathematically, the forward pass is given by \( h = W^{\text{DDequant}}_{\text{NF4}} x + \alpha \sum_{b=1}^{r} (W_{\text{up}})_{:,b} (W_{\text{dw}})_{b,:} x \), where \( W^{\text{DDequant}}_{\text{NF4}} \) is the dequantized pretrained weight, \( x \) is the input, \( \alpha \) is the LoRA scalar, \( r \) is the sampled rank, and \( W_{\text{up}} \) and \( W_{\text{dw}} \) are the up- and down-projection matrices, respectively. This approach reduces memory usage during training and inference, making it suitable for large-scale LLMs. \textbf{BitDelta}~\cite{liu2024bitdelta} is an efficient post-training quantization method for compressing large language models after fine-tuning. The core idea is to represent the fine-tuning induced weight delta, \( \Delta = W_{\text{fine}} - W_{\text{base}} \), where \( W_{\text{fine}} \) is the weight matrix of the fine-tuned model and \( W_{\text{base}} \) is the base pre-trained model's weight, using only 1 bit. This is achieved by quantizing \( \Delta \) to its sign bits and a trainable scaling factor \( \alpha \), resulting in \( \hat{\Delta} = \alpha \odot \text{Sign}(\Delta) \). The scaling factor is initialized to minimize the L2 norm of the error and further refined through distillation to align the quantized model's output with the original fine-tuned model. This approach dramatically reduces memory requirements and can enhance inference speed, with minimal impact on performance.

%% file: Multi-task_PEFT.tex
\begin{figure}[htbp]
    \centering
    \includegraphics[width=\columnwidth]{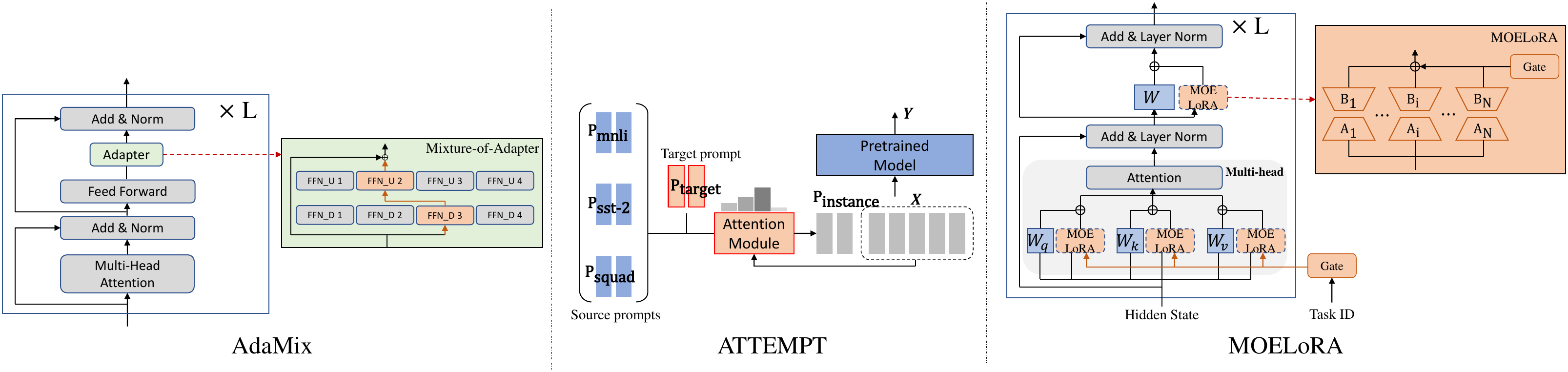} 
    \caption{Illustration of three representative multi-task PEFT methods: AdaMix (Adapter-based), ATTEMPT (Soft Prompt-based), and MOELoRA (LoRA-based)} 
    \label{fig:multi_task}
\end{figure}

The previously introduced PEFT methods were mainly designed for single downstream task. This section focuses on PEFT for multi-task learning. Figure~\ref{fig:multi_task} illustrates three multi-task PEFT approaches: AdaMix (Adapter-based), ATTEMPT (Soft Prompt-based), and MOELoRA (LoRA-based).

\subsubsection{Adapter-based}
\input{Adapter-based}
\subsubsection{Soft Prompt-based}
\input{Soft_Prompt-based}

\subsubsection{LoRA-based}
\input{LoRA-based}

%% file: Adapter-based.tex
\textbf{AdapterFusion}~\cite{pfeiffer2020adapterfusion} employs a two-stage approach to transfer learning, where it first extracts knowledge into task-specific adapters and then composes this knowledge in a separate step to exploit multi-task representations without destructive interference. \textbf{AdaMix}~\cite{wang2022adamix} integrates multiple adaptation modules within each Transformer layer of a pre-trained language model, enabling efficient tuning with a mixture of these modules while maintaining most of the model's weights unaltered. \textbf{PHA}~\cite{zhao2023prototype} leverages an instance-dense retriever and a prototypical hypernetwork to efficiently generate task-specific adapter layers by retrieving prototype embeddings and feeding them into the hypernetwork, enabling sample-efficient multi-task learning and new task generalization. \textbf{AdapterSoup}~\cite{chronopoulou2023adaptersoup} improves the generalization of pretrained language models to new domains by averaging the weights of adapters trained on different domains, without the need for additional training or increasing inference cost. \textbf{MerA}~\cite{he2023mera} efficiently incorporates pretrained adapters into a single model through model fusion, aligning the parameters via optimal transport based on weights and activations to enhance performance in few-shot learning scenarios. \textbf{Hyperformer}~\cite{mahabadi2021parameter} integrates hypernetwork-based adapter layers into a transformer model, enabling the model to share knowledge across tasks while adapting to each individual task through task-specific adapters generated by shared hypernetworks.

%% file: Soft_Prompt-based.tex
\textbf{SPoT} (Soft Prompt Transfer)~\cite{vu2022spot} leverages soft prompts to adapt pre-trained language models efficiently. It first trains a soft prompt \( p \) on one or more source tasks, where \( p \in \mathbb{R}^{d} \) represents a sequence of continuous vectors with dimensionality \( d \). This learned prompt is then used to initialize the prompt for a target task, facilitating transfer learning. SPoT significantly improves upon the performance of prompt tuning and matches or outperforms full model fine-tuning while using significantly fewer task-specific parameters. \textbf{ATTEMPT} (ATTEntional Mixtures of Prompt Tuning)~\cite{asai2022attempt} leverages pre-trained soft prompts \(P_1, \ldots, P_t\) for different high-resource tasks and a new target prompt \(P_{\text{target}}\). An attention module \(G\) computes attention scores between input \(X\) and each prompt token to produce an instance-wise prompt \(P_{\text{instance}} = \sum_{j=1}^{t+1} a_j P_j\), where \(a_j\) represents the attention weight for prompt \(P_j\). Only \(P_{\text{target}}\) and \(G\) are updated during training, keeping the original language model frozen. This approach is parameter-efficient and flexible for multi-task learning. \textbf{MPT} (Multitask Prompt Tuning)~\cite{wang2022multitask} is a method for efficient transfer learning of large language models across multiple downstream tasks. The core idea is to distill knowledge from multiple task-specific source prompts into a single transferable prompt, \( P^* \), which is then adapted to each target task with minimal additional parameters. The prompt for each source task is decomposed into a shared matrix \( P^* \) and a low-rank task-specific matrix \( W_k = u_k \otimes v_k^T \), where \( u_k \) and \( v_k \) are task-specific vectors. This decomposition is learned through a knowledge distillation process that minimizes the KL-divergence between teacher and student prompts, \( L_{\text{Logits}} \), and an additional mean squared loss on the hidden states, \( L_{\text{Hidden}} \). The total training loss is \( L_{\text{Total}} = L_{\text{PLM}} + \lambda(L_{\text{Logits}} + L_{\text{Hidden}}) \), where \( L_{\text{PLM}} \) is the task loss and \( \lambda \) balances the distillation impact. The innovation lies in leveraging cross-task knowledge within a parameter-efficient framework, which outperforms full finetuning with far fewer task-specific parameters. \textbf{IPT} (Intrinsic Prompt Tuning)~\cite{qin2021exploring} is a method to reparameterize the adaptation of pre-trained language models to various tasks within a low-dimensional intrinsic task subspace. The key idea is to decompose the soft prompts \( P \) for multiple NLP tasks into a shared, lower-dimensional space using an auto-encoder with projection \( \text{Proj}(\cdot) \) and back-projection \( \text{Projb}(\cdot) \) functions. The auto-encoder is trained to minimize the reconstruction loss \( L_{AE} = || P^* - P ||_2^2 \), where \( P^* = \text{Projb}(\text{Proj}(P)) \). The intrinsic dimension \( d_I \) determines the size of this subspace. After finding the subspace, IPT tunes only \( d_I \) parameters to adapt PLMs to new tasks or data, suggesting that the adaptations can be generalized across tasks by optimizing a small set of free parameters in a unified subspace. \textbf{TPT} (transferable prompt tuning)~\cite{su2021transferability} investigates transferring soft prompts across tasks and models to improve prompt tuning (PT) efficiency. Soft prompts \( P = \{p_1, p_2, \ldots, p_l\} \), where \( p_i \in \mathbb{R}^d \) and \( d \) is the input dimension, are prepended to input sequences \( X = \{x_1, x_2, \ldots, x_n\} \). The objective is to maximize the likelihood \( L = p(y|P, x_1,\ldots,x_n) \) of generating desired outputs \( y \), with \( P \) being the only trainable component. Transferability is explored through initializing with similar tasks' prompts and using a cross-model projector. The overlapping rate of activated neurons is found to be a strong indicator of transferability.

%% file: LoRA-based.tex
\textbf{LoRAHub}~\cite{huang2023lorahub} is a dynamic composition of multiple LoRA modules, represented as \( \hat{m} = (w_1A_1 + w_2A_2 + \ldots + w_NA_N)(w_1B_1 + w_2B_2 + \ldots + w_NB_N) \), followed by a gradient-free optimization to determine the coefficients \( w_i \) that best adapt the combined module for performance on new, unseen tasks. \textbf{MOELoRA}~\cite{liu2023moelora} integrates a Mixture-of-Experts (MOE) model with trainable experts \( \{E_i\}_{i=1}^{N} \), each consisting of a pair of low-rank matrices \( B_i \in \mathbb{R}^{d_{in} \times r} \) and \( A_i \in \mathbb{R}^{r \times d_{out}} \), along with a task-motivated gate function that outputs expert weights \( \omega_{ji} \) for task \( T_j \), to efficiently fine-tune large language models for multi-task medical applications while maintaining a compact set of trainable parameters. \textbf{L-LoRA} (Linearized LoRA)~\cite{tang2023parameter} is a novel partial linearization method for parameter-efficient fine-tuning models, which enhances weight disentanglement and improves multi-task fusion capability with a low computational cost overhead by linearizing only the adapter modules and applying model fusion algorithms over the linearized adapters. \textbf{MTLoRA}~\cite{agiza2024mtlora} revolves around the use of Task-Agnostic and Task-Specific Low-Rank Adaptation modules to efficiently adapt a shared transformer backbone for multiple downstream tasks in a Multi-Task Learning architecture, balancing between learning shared features and those specific to individual tasks.

%% file: Applications_of_PEFT.tex
\tikzstyle{my-box}=[
 rectangle,
 draw=hidden-draw,
 rounded corners,
 text opacity=1,
 minimum height=1.5em,
 minimum width=5em,
 inner sep=2pt,
 align=center,
 fill opacity=.5,
 ]
 \tikzstyle{leaf}=[my-box, minimum height=1.5em,
 fill=hidden-orange!60, text=black, align=left,font=\scriptsize,
 inner xsep=2pt,
 inner ysep=4pt,
 ]

\begin{figure*}[t]
\centering
\resizebox{\textwidth}{!}{
\begin{forest}
        forked edges,
        for tree={
            grow=east,
            reversed=true,
            anchor=base west,
            parent anchor=east,
            child anchor=west,
            base=left,
            font=\small,
            rectangle,
            draw=hidden-draw,
            rounded corners,
            align=left,
            minimum width=4em,
            edge+={darkgray, line width=1pt},
            s sep=3pt,
            inner xsep=2pt,
            inner ysep=3pt,
            ver/.style={rotate=90, child anchor=north, parent anchor=south, anchor=center},
        },
        where level=1{text width=7.0em,font=\scriptsize}{},
        where level=2{text width=6em,font=\scriptsize}{},
        where level=3{text width=5.5em,font=\scriptsize}{},
        [
            PEFT Methods for Other PLMs, ver
            [
                Vision Models  
                [
                    Image \\ Classification 
                    [
                        VP~{\cite{bahng2022exploring},}
                        VPT~{\cite{jia2022visual},}
                        NOAH~{\cite{zhang2022neural},} \\
                        Convpass~{\cite{jie2022convolutional},}  
                        AdaptFormer~{\cite{chen2022adaptformer},} \\ 
                        DAM-VP~{\cite{huang2023diversity},} 
                        ILM-VP~{\cite{chen2023understanding},} 
                        EVP~{\cite{wu2024evp},} \\
                        VQT~{\cite{tu2023visual},} 
                        FacT~{\cite{jie2023fact},}
                        DTL~{\cite{fu2024dtl},} \\ 
                        LION~{\cite{wang2024lion}} 
                        , leaf, text width=34.1em
                    ]
                ]
                [
                    Dense \\ Prediction 
                    [
                        Polyhistor~{\cite{liu2022polyhistor},}
                        ViT-Adapter~{\cite{chen2022vision},}
                        SAN~{\cite{xu2023side},} \\ 
                        LoRand~{\cite{yin20231}}
                        , leaf, text width=34.1em
                    ]
                ]
            ]
            [
                Diffusion Models  
                [
                    Generation by \\ Few-shot \\ Fine-tuning 
                    [
                        DreamBooth~{\cite{ruiz2023dreambooth},}
                        Textual Inversion~{\cite{gal2022image},} \\ 
                        DreamArtist~{\cite{dong2022dreamartist},}
                        Extended Textual Inversion~{\cite{voynov2023p+},} \\
                        DiffFit~{\cite{xie2023difffit},} 
                        Cones~{\cite{liu2023cones},} 
                        SVDiff~{\cite{han2023svdiff},} \\ 
                        LyCORIS~{\cite{yeh2023navigating},}  
                        DiffuseKronA~{\cite{marjit2024diffusekrona},} 
                        OFT~{\cite{qiu2024controlling}}
                        , leaf, text width=34.1em
                    ]
                ]
                [
                    Controllable \\ Generation 
                    [
                        Sketch-guided Diffusion~{\cite{voynov2023sketch},}
                        ControlNet~{\cite{zhang2023adding},} \\ 
                        T2I-Adapter~{\cite{mou2023t2i},}
                        Uni-ControlNet~{\cite{zhao2024uni},} \\
                        IP-Adapter~{\cite{ye2023ip}}
                        , leaf, text width=34.1em
                    ]
                ]
            ]
            [
                MLLM 
                [
                    BLIP-2~{\cite{li2023blip},}
                    LLaVA~{\cite{liu2023improved, liu2024visual},}
                    Flamingo~{\cite{alayrac2022flamingo},} \\ 
                    LLaMA Adapter~{\cite{zhang2023llama},}
                    CogVLM~{\cite{wang2023cogvlm},} \\ 
                    Q-Former~{\cite{li2023blip, dai2024instructblip},} 
                    GPT-4~{\cite{achiam2023gpt}}
                    , leaf, text width=41.7em
                ]
            ]
        ]
    \end{forest}
}
\caption{Taxonomy of PEFT Methods for Vision Models, Diffusion Models and MLLM}
\label{fig:taxonomy_of_peft_other}
\end{figure*}

This section presents a comprehensive overview of PEFT methodologies specifically developed for several prominent applications, categorized as follows: \textbf{PEFT in Vision Models} (\ref{sec_peft_vision}), which primarily focuses on adapting pretrained vision models to specialized computer vision tasks (e.g., image classification, image segmentation, object detection, and depth estimation); \textbf{PEFT in Diffusion Models} (\ref{sec_peft_diffusion}), which addresses the adaptation of diffusion models for vision generation tasks; and \textbf{PEFT in MLLM} (\ref{sec_peft_mllm}), which emphasizes training model connectors on domain-specific datasets to bridge multimodal data discrepancies while maintaining input consistency for LLMs. For a structured overview of these applications and their corresponding recommended PEFT techniques, refer to Figure~\ref{fig:taxonomy_of_peft_other}.

\subsection{PEFT in Vision Models}
\label{sec_peft_vision}
Over the past decade, deep learning has achieved significant advancements in the field of computer vision, particularly with the introduction of the ImageNet dataset and the widespread adoption of the pre-training-fine-tuning paradigm based on pretrained vision models (PVMs). Numerous studies have shown that better ImageNet pre-training performance typically leads to improved performance on downstream tasks. As visual pre-trained models continue to evolve, especially with the introduction of Vision Transformer (ViT) architectures, the scale of model parameters has increased significantly, highlighting the inefficiencies of traditional full fine-tuning methods in terms of parameter efficiency. To address these issues and improve parameter efficiency during the fine-tuning process of PVMs, various PEFT methods have emerged. These methods have demonstrated their advantages across multiple domains, including image classification, dense prediction, video analysis, and 3D point cloud analysis. This section will focus on the application of PEFT methods in image classification and dense prediction tasks.
\input{Vision_Models}

\subsection{PEFT in Diffusion Models}
\label{sec_peft_diffusion}
As diffusion models evolve, these models have now surpassed GANs as the mainstream method in the image generation domain. Given their success in image generation, their potential applications in video generation, 3D content generation, and speech synthesis are also becoming increasingly apparent. Additionally, many application domains involve fine-tuning diffusion models, including embedding personalized concepts in image generation, customizing generated images based on reference images, and training multi-view image generation capabilities based on pre-trained text-to-image diffusion models in the 3D content generation domain. Compared to the NLP field, research on PEFT for diffusion models is relatively scarce. Current research mainly focuses on two areas: generation by few-shot finetuning and controllable generation in image generation:
\input{Diffusion_Models}

\subsection{PEFT in MLLM}
\label{sec_peft_mllm}
\input{MLLM}

%% file: Vision_Models.tex
\subsubsection{Image Classification}
\input{Image_Classification}

\subsubsection{Dense Prediction}
\input{Dense_Prediction}

%% file: Image_Classification.tex
\begin{figure}[htbp]
    \centering
    \includegraphics[width=\columnwidth]{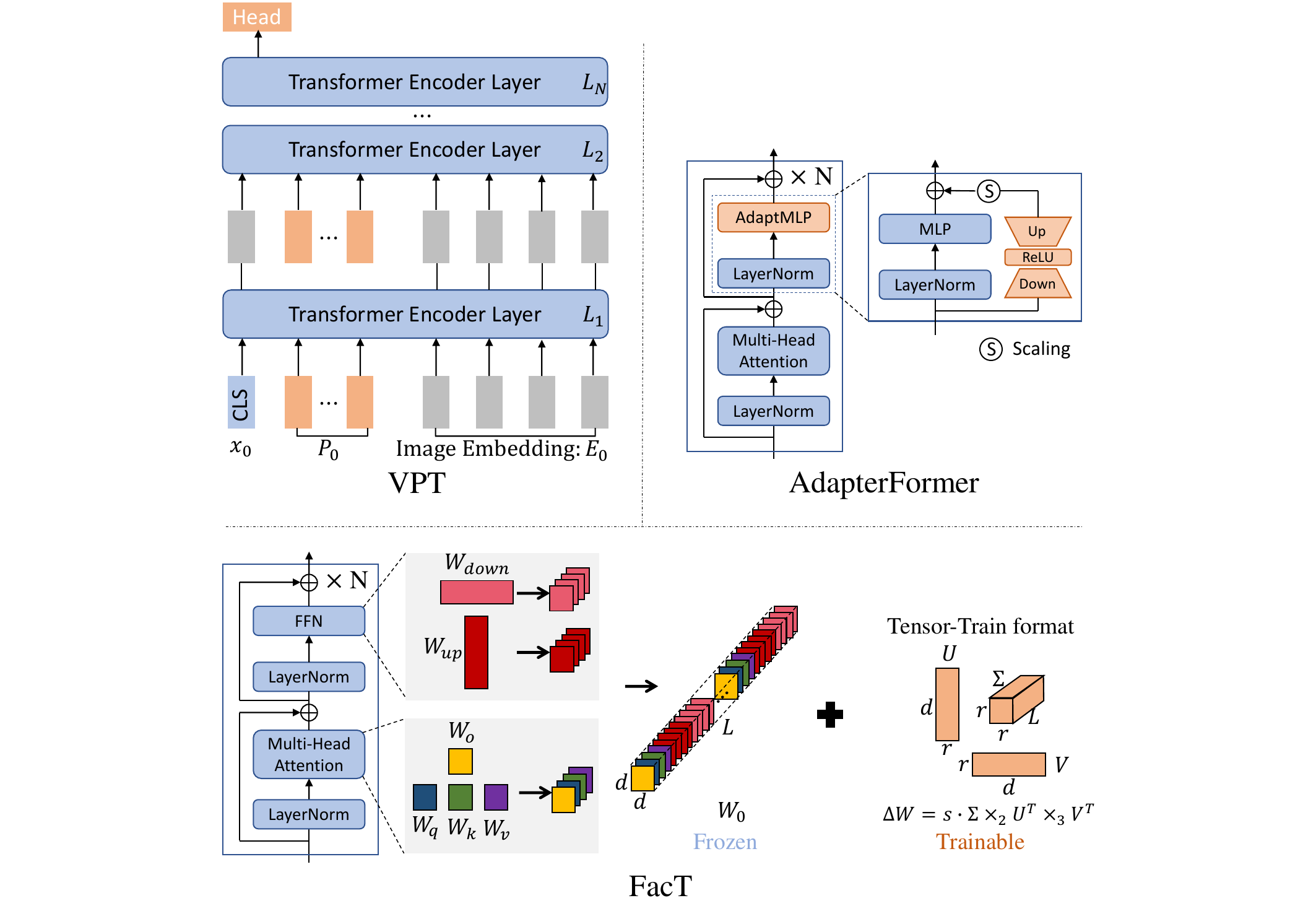} 
    \caption{Illustration of the principles of three PEFT methods for image classification: VPT (Soft Prompt-based), AdapterFormer (Adapter-based), and FacT (LoRA-based). $\times_{i}$ in FacT is mode-$i$ product.} 
    \label{fig:image_classification}
\end{figure}

In this subsection, we introduce PEFT methods for image classification tasks in vision models. Figure~\ref{fig:image_classification} illustrates the principles of three representative PEFT methods discussed in this subsection.

\textbf{VP}~\cite{bahng2022exploring} investigates visual prompting as a means to adapt large-scale pre-trained models for new tasks without updating model parameters. A single image perturbation (\(\delta\)) is learned such that when added to input images (\(x\)), the prompted image (\(x' = x + \delta\)) steers the model's prediction towards a target task. This method is akin to adversarial reprogramming, but it aims for constructive task adaptation. Its effectiveness is demonstrated through experiments, which show competitive performance compared to linear probes. Notably, the approach is input-agnostic and dataset-wide. \textbf{VPT} (Visual Prompt Tuning)~\cite{jia2022visual} adapts pre-trained vision Transformers for downstream tasks by introducing task-specific, learnable parameters (\(P = \{p_k \in \mathbb{R}^d | k \in \mathbb{N}, 1 \leq k \leq m\}\)) into the input sequence, while keeping the backbone of the model frozen. Here, \(d\) represents the dimensionality of the input features, while \(m\) signifies the total number of prompts. These prompts \(P\) are prepended to the input sequence of each Transformer layer and learned alongside a linear classification head during fine-tuning. \textbf{NOAH} (Neural prOmpt seArcH)~\cite{zhang2022neural} automatically searches for the optimal design of prompt modules for large vision models through Neural Architecture Search (NAS). NOAH encompasses three prompt modules: Adapter, LoRA, and VPT, each inserted into Transformer blocks. The search space includes parameters like embedding dimensions ${D}=\{5, 10, 50, 100\}$ and depths ${L}=\{3, 6, 9, 12\}$, determining the range of applications. An AutoFormer-based one-shot NAS algorithm is employed to select the best configuration for each downstream dataset. \textbf{Convpass}~\cite{jie2022convolutional}, convolutional bypasses for ViTs, to serve as adaptation modules during finetuning. Convpass, introduced as a parallel convolutional bottleneck block to the Multi-Head Self-Attention (MHSA) or MLP blocks, "bypasses" the original ViT block. For a ViT layer, the input sequence \( X \in \mathbb{R}^{N \times d} \) is processed through Convpass, reconstructing the spatial structure of the token sequence. During finetuning, only Convpass modules and the classification head are updated. Convpass leverages the inductive bias of convolutional layers, enhancing its suitability for visual tasks, particularly in low-data scenarios. \textbf{AdaptFormer}~\cite{chen2022adaptformer} is a lightweight module designed for efficient fine-tuning of pre-trained ViTs on diverse visual recognition tasks. It introduces additional trainable parameters, consisting of two fully connected layers \(\textit{FC}_1, \textit{FC}_2\), a non-linear activation function (\(\sigma\)), and a scaling factor (\(\alpha\)). These components are placed in parallel with the feed-forward network (FFN) of the original ViT. The learnable parameters of AdaptFormer are updated during the fine-tuning phase, while the pre-trained ViT parameters remain frozen. This design enables AdaptFormer to enhance the transferability of ViTs with minimal parameter updates, thereby improving scalability and performance on various visual tasks. \textbf{DAM-VP} (Diversity-Aware Meta Visual Prompting)~\cite{huang2023diversity} partitions a dataset into homogeneous subsets based on diversity, optimizing a unique prompt for each subset. Prompts are initialized with a meta-prompt learned across multiple datasets, improving convergence speed and performance. During inference, the appropriate prompt is selected based on the feature distance between input and subset prototypes. Formally, for a dataset ${D}$ divided into $K$ subsets ${D}_1, {D}_2, ..., {D}_K$, the optimal prompts $p^*_1, ..., p^*_K$ are found by minimizing the cross-entropy loss:
\begin{equation}
    p^*_1, ..., p^*_K = \arg \min_{p_1, ..., p_K} \sum_{k=1}^{K} \sum_{x \in {D}_k} {L}_{\text{CE}}(M(x + p_k), y)
    \enskip, 
\end{equation}
where $p_k$ is the prompt for subset ${D}_k$, $M$ is the pre-trained model, $x$ is an input image, $y$ is the ground truth label, and ${L}_{\text{CE}}$ is the cross-entropy loss function. \textbf{ILM-VP}~\cite{chen2023understanding} is an iterative label mapping-based visual prompting method. It optimizes the mapping between source and target labels to improve the accuracy of reprogramming pre-trained models for new tasks. The key equation is:
\begin{equation}
    \min_{\delta} \sum_{yt \in Ttr} \min_{ys \in Ss} {L}(f_\theta(x + \delta), ys; yt)
    \enskip, 
\end{equation}
where $\delta$ is the visual prompt, ${L}$ is the cross-entropy loss, $f_\theta$ is the pre-trained model, $x$ is the input image, $Ttr$ is the target training set, $Ss$ is the set of source labels, and $ys$ and $yt$ are the source and target labels, respectively. ILM-VP enhances interpretability by providing meaningful mappings. \textbf{EVP} (Enhanced Visual Prompting)~\cite{wu2024evp} is a method for adapting pre-trained models to downstream tasks without substantial parameter updates. Instead of directly combining the prompt \( P \) and the image \( I \), they shrink \( I \) and pad \( P \) around it, ensuring independence. They also reintroduce input diversity and gradient normalization techniques, originally used in adversarial example generation, to improve the optimization and generalizability of the prompt. This approach outperforms linear probing and matches fully fine-tuning in some cases, with significantly fewer parameters. \textbf{VQT} (Visual Query Tuning)~\cite{tu2023visual} leverages learnable ``query" tokens in each Transformer layer to summarize intermediate features effectively. VQT introduces a set \({Q} = \{q_1, q_2, \ldots, q_n\}\) where \(q_i \in \mathbb{R}^{d}\) represents the \(i\)-th query token with \(d\) being the feature dimension. These queries interact with the intermediate features \(X \in \mathbb{R}^{N \times d}\) through the attention mechanism, where \(N\) is the number of tokens. The output \(Z = \{z_1, z_2, \ldots, z_n\}\) summarizes the layer's information, with \(z_i\) denoting the summary for \(q_i\). This enables efficient transfer learning with memory and parameter savings. \textbf{FacT}~\cite{jie2023fact} is a method for efficient fine-tuning of pre-trained ViTs by updating only a fraction of parameters. The key idea is to tensorize the weights of ViT into a 3D tensor and decompose the weight increments into lightweight factors. During fine-tuning, only these factors are updated and stored. Mathematically, if \(\Delta W\) represents the increment of a weight matrix \(W\), then \(\Delta W\) is approximated as \(\Delta W \approx A \times B\), where \(A\) and \(B\) are the decomposed factors. \(A\) and \(B\) are learned during fine-tuning, reducing storage requirements. \textbf{DTL} (Disentangled Transfer Learning)~\cite{fu2024dtl} addresses the inefficiency of Parameter-Efficient Transfer Learning (PETL) methods in GPU memory usage. DTL employs a Compact Side Network (CSN) to disentangle trainable parameters from the backbone. CSN uses low-rank linear mappings to extract and reintegrate task-specific information. Formally, given a backbone with $N$ blocks, the output $z_{i+1}$ of the $i$-th block is updated as $z'_{i+1} = z_{i+1} + \theta(h_{i+1})$ for $i \geq M$, where $\theta$ is a non-linear activation function, and $h_{i+1}$ captures the task-specific information extracted by CSN. This disentanglement significantly reduces GPU memory footprint and trainable parameters while maintaining or improving accuracy. \textbf{LION} (impLicit vIsion prOmpt tuNing)~\cite{wang2024lion} inserts two equilibrium implicit layers (\(P_1\), \(P_2\)) at the start and end of a frozen pre-trained backbone (\(\theta\)). \(P_1\) and \(P_2\) are defined as:
\begin{equation}
    P_1 = f_{eq}^{(1)}(x; \phi_1), \quad P_2 = f_{eq}^{(2)}(z; \phi_2)
    \enskip,
\end{equation}
where \(x\) is the input, \(z\) is the output of the backbone, and \(\phi_1\), \(\phi_2\) are parameters of the implicit layers. \(f_{eq}\) denotes the equilibrium function. To reduce computational burden, parameters are pruned based on the lottery ticket hypothesis. LION adapts the backbone to downstream tasks efficiently with minimal parameter updates.

%% file: Dense_Prediction.tex
\begin{figure}[htbp]
    \centering
    \includegraphics[width=\columnwidth]{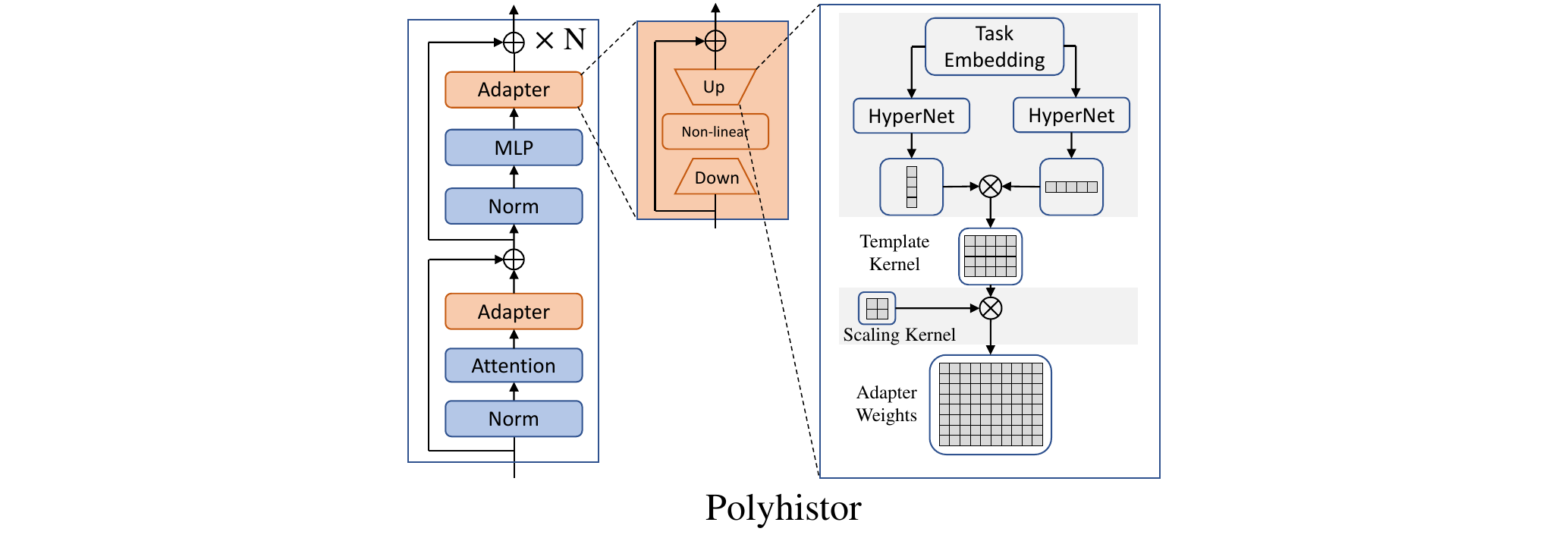} 
    \caption{Illustration of a representative PEFT method for dense prediction: Polyhistor} 
    \label{fig:dense_prediction}
\end{figure}

Dense prediction, encompassing tasks such as image segmentation, object detection, depth estimation, etc., is another crucial task in the field of 2D vision. Unlike image classification tasks, which typically generate a single prediction label for an entire image, dense prediction tasks require making predictions for every pixel in the image, usually resulting in an output image with the same resolution as the input image. Fine-tuning pre-trained models from image classification is a common approach for dense prediction tasks. With the application of PEFT methods in vision tasks, various PEFT methods tailored for dense prediction tasks have been proposed. Figure~\ref{fig:dense_prediction} illustrates a representative PEFT method for dense prediction.

\textbf{Polyhistor}~\cite{liu2022polyhistor} employs a strategy of hypernetworks that are broken down into components, along with scaling kernels applied at each layer, to facilitate the sharing of information across various tasks efficiently and with a minimal number of parameters. In this approach, the weight matrix of each adapter, denoted as \( W \), is decomposed into two distinct elements: a template kernel \( T \) and a scaling kernel \( S \). The weight matrix is then reconstructed through the Kronecker product of these two kernels, represented as \( W = T \otimes S \). This method effectively reduces the number of parameters required while still preserving the level of accuracy in the system. \textbf{ViT-Adapter}~\cite{chen2022vision} leverages the inherent representation power of a plain ViT backbone and augments it with an adapter that incorporates image-specific inductive biases during fine-tuning. This enables the model to capture high-frequency details crucial for tasks like object detection and segmentation. \textbf{SAN} (Side Adapter Network)~\cite{xu2023side} decouples mask proposal generation and class recognition for open-vocabulary semantic segmentation. A lightweight side network is attached to a frozen CLIP model, predicting mask proposals and attention bias to guide CLIP's recognition of the mask's class. This design leverages CLIP's robustness while minimizing additional parameters and computational cost. The attention bias is applied in CLIP's attention mechanism \( \operatorname{Attention}(Q, K, V, \text{bias}) \), where \( Q \), \( K \), and \( V \) represent query, key, and value vectors, enhancing CLIP's awareness of the proposed regions. \textbf{LoRand}~\cite{yin20231} adds lightweight, low-rank adapter modules to a pre-trained vision model, such as the Swin Transformer, without updating the original model's parameters. These adapters consist of multi-branch low-rank projections and non-linearities, enabling them to capture complex representations with minimal parameters. Specifically, for a backbone with parameters \(\theta\), LoRand trains a small subset \(\phi\) (\(1\% - 3\%\)) of \(\theta\), where \(\phi \subset \theta\), achieving competitive performance with full fine-tuning while significantly reducing the number of trainable parameters.

%% file: Diffusion_Models.tex
\subsubsection{Generation by Few-shot Finetuning}
\input{Generation_by_Few-shot_Finetuning}

\subsubsection{Controllable Generation}
\input{Controllable_Generation}

%% file: Generation_by_Few-shot_Finetuning.tex
Generation by few-shot finetuning involves providing a few images (or even just one) of an object or style, and fine-tuning the model on these images. This process allows the model to generate new images that reflect the unique characteristics of the provided examples. 

\textbf{DreamBooth}~\cite{ruiz2023dreambooth} is a method for personalizing text-to-image diffusion models using just a few images of a subject. The technique fine-tunes a pre-trained model with a novel autogenous class-specific prior preservation loss, to bind a unique identifier to the subject and preserve class diversity. This enables generating photorealistic images of the subject in various scenes while maintaining key features. The fine-tuning process involves adjusting the model parameters based on input images and text prompts, leveraging the model's semantic prior and the new loss function to enhance subject fidelity and versatility in image synthesis. \textbf{Textual Inversion}~\cite{gal2022image} is a method that personalizes text-to-image generation by embedding unique concepts as new "pseudo-words" in the latent space of a pre-trained model. This allows intuitive composition into sentences guiding image creation, capturing both semantics and details without retraining the model. The innovation lies in optimizing a single word embedding to represent a concept through reconstruction, balancing distortion and editability. The method's strength is its simplicity and compatibility with existing models, while its limitation is the potential for less precise shape retention. \textbf{DreamArtist}~\cite{dong2022dreamartist} leverages positive-negative prompt-tuning to enable one-shot text-to-image generation. Given a reference image \( I \), it learns a positive embedding \( S^*_p \) that captures the image's characteristics and a negative embedding \( S^*_n \) that rectifies deficiencies. \( S^*_p \) drives diverse generation, while \( S^*_n \) ensures corrections, improving controllability. The embeddings are combined through a fusion function \( f_m(z_p, z_n) \) where \( z_p \) and \( z_n \) represent the latent representations of positive and negative prompts, respectively. This approach facilitates the synthesis of high-quality, diverse, and controllable images from a single reference. In paper~\cite{voynov2023p+}, an \textbf{Extended Textual Conditioning (P+)} space is introduced for text-to-image generation, allowing for more granular control over image synthesis through per-layer textual prompts. The innovation, \textbf{Extended Textual Inversion}, inverts images into P+ space using a set of token embeddings, enhancing expressiveness and precision without compromising editability. This method is advantageous due to its faster convergence and the ability to achieve finer control over image attributes by leveraging the distinct sensitivities of U-net layers to shape or appearance. The downside includes imperfect concept reconstruction and the relatively slow inversion process. \textbf{DiffFit}~\cite{xie2023difffit} fine-tunes only the bias terms and introduces scaling factors \( \gamma \) in specific layers, initialized to 1.0, to adapt to new domains quickly. The method achieves significant training efficiency and reduced storage costs, with \( \gamma \) enhancing feature scaling for better adaptation. The efficacy is theoretically justified by analyzing the shift in distributions caused by the scaling factors. \textbf{SVDiff}~\cite{han2023svdiff} is a method for fine-tuning text-to-image diffusion models by adjusting the singular values (\(\sigma_i\)) of weight matrices (\(W\)), represented as \(W = \sum_i \sigma_i u_i v_i^\top\), where \(u_i\) and \(v_i\) are the left and right singular vectors, respectively. This approach leads to a compact parameter space, reducing overfitting and model size (\(\approx2,200\times\) fewer parameters than DreamBooth). They also introduce Cut-Mix-Unmix for improved multi-subject generation and a single-image editing framework. \textbf{LyCORIS}~\cite{yeh2023navigating} is an open-source library for fine-tuning Stable Diffusion models. It implements methods like LoRA, LoHa, LoKr, GLoRA, and $(IA)^3$. The library aims to simplify the integration and evaluation of these methods. A comprehensive evaluation framework is proposed, using metrics for concept fidelity, text-image alignment, diversity, and style preservation. Experiments highlight the nuanced impacts of hyperparameters and the suitability of different methods for specific tasks. \textbf{DiffuseKronA}~\cite{marjit2024diffusekrona} utilizes a Kronecker product-based adaptation mechanism to efficiently fine-tune large diffusion models for personalized text-to-image generation. The method reduces the parameter count by applying truncated singular value decomposition on critical model layers, enabling subject-specific image synthesis with enhanced stability, interpretability, and text alignment. The approach offers a \(\geq 50\%\) parameter reduction compared to state-of-the-art methods, with comparable or superior image quality. \textbf{OFT} (Orthogonal Finetuning)~\cite{qiu2024controlling} is a method to adapt text-to-image diffusion models for downstream tasks without losing generative performance. OFT preserves the hyperspherical energy which characterizes neuron relationships by applying a layer-shared orthogonal transformation \( R \) to the pretrained weights \( W_0 \). This maintains the pairwise angles among neurons, crucial for semantic information. The transformation is constrained as \( R^T R = R R^T = I \), ensuring minimal deviation from the original model. A variant, Constrained Orthogonal Finetuning (COFT), further limits angular deviation with \( \|R - I\| \leq \epsilon \). The method aims to balance flexibility and stability in finetuning.

%% file: Controllable_Generation.tex
\begin{figure}[htbp]
    \centering
    \includegraphics[width=\columnwidth]{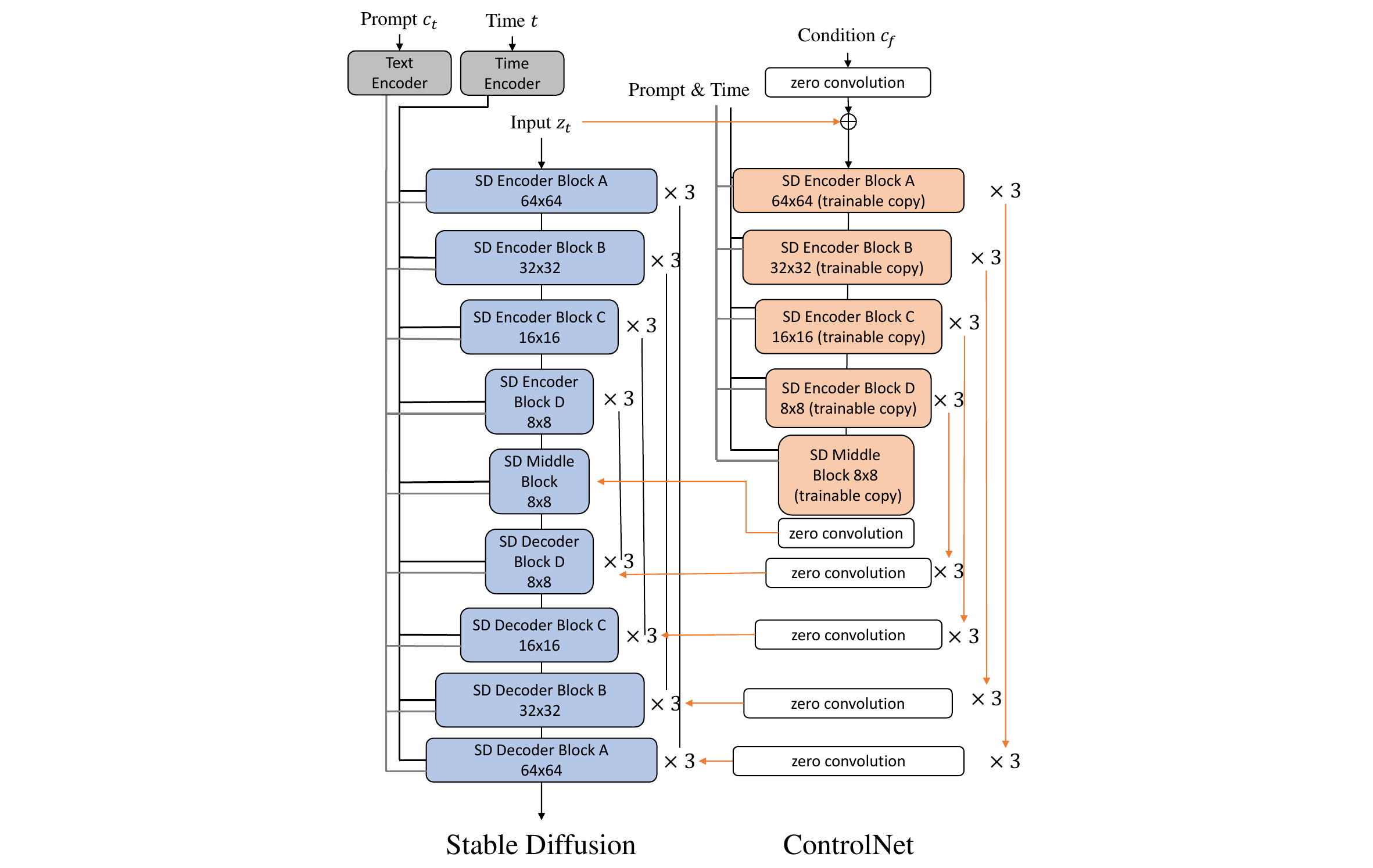}
    \caption{Illustration of the principle of ControlNet, a representative implementation of a controllable generation method} 
    \label{fig:controllable_generation}
\end{figure}

Controllable generation primarily involves adding control sources beyond the prompt to guide the image generation. These control sources can include sketches, keypoints, or other forms of guidance to shape the generated output more precisely. A representative implementation of controllable generation method is shown in Figure~\ref{fig:controllable_generation}

\textbf{Sketch-guided Diffusion}~\cite{voynov2023sketch} is a method to guide pre-trained text-to-image diffusion models using spatial maps like sketches. It involves training a lightweight per-pixel multi-layer perceptron (MLP), named the latent guidance predictor (LGP), to map noisy image features to spatial maps. The LGP is trained on a small dataset, predicting spatial layouts from latent features \( F(\mathbf{z}_t|\mathbf{c}, t) \) extracted from a denoising diffusion probabilistic model (DDPM) network, where \( \mathbf{z}_t \) is a noisy image at timestep \( t \), and \( \mathbf{c} \) presents the conditioning text prompt. \textbf{ControlNet}~\cite{zhang2023adding} enhances pretrained text-to-image diffusion models by adding spatially localized conditions. For a neural block \(F(x; \Theta)\) transforming input \(x\) to output \(y\), ControlNet freezes \(\Theta\) and introduces a trainable copy. Conditions \(c\) are injected through zero-initialized convolution layers (zero convolutions) ensuring no initial noise. \(y_c = F(x, c; \Theta')\) represents the output with conditions, where \(\Theta'\) denotes the updated parameters. This approach facilitates robust finetuning and sudden convergence. \textbf{T2I-Adapter}~\cite{mou2023t2i} enhances controllability of pre-trained text-to-image (T2I) models by learning lightweight adapter models that align the model's internal knowledge with external control signals. This is achieved without modifying the original T2I model, allowing for granular control over generated images' structure and color. Mathematically, let \( \mathcal{M} \) denote the pre-trained T2I model, \( \mathcal{A} \) the adapter, and \( \mathbf{x}_c \) the control signal (e.g., sketches, masks). The adapted model generates images \( \mathbf{x} \) from text prompts \( t \) and control signals \( \mathbf{x}_c \) as follows: 
\begin{equation}
    \mathbf{x} = \mathcal{M}_{\text{adapted}}(t, \mathbf{x}_c) = \mathcal{M}(t) + \omega \cdot \mathcal{A}(\mathbf{x}_c)
    \enskip,
\end{equation}
where \( \omega \) is a weighting factor balancing the influence of the control signal. The adapter \( \mathcal{A} \) is trained to translate \( \mathbf{x}_c \) into a form that can steer \( \mathcal{M} \) towards desired outputs, enabling precise control. \textbf{Uni-ControlNet}~\cite{zhao2024uni} integrates diverse control signals into pre-trained text-to-image (T2I) diffusion models through two lightweight adapters, facilitating efficient and composable control. It employs a multi-scale condition injection strategy, using Feature Denormalization (FDN) to modulate noise features with local conditions:

\begin{multline}
    \text{F}_{\text{DN}r}(\text{Z}_r, c_l) = \text{norm}(\text{Z}_r) \cdot (1 + \text{conv}_\gamma(\text{zero}(h_r(c_l))))
    + \text{conv}_\beta(\text{zero}(h_r(c_l)))
    \enskip,
\end{multline}

where $\text{Z}_r$ are noise features at resolution $r$, $c_l$ are concatenated local conditions, $h_r$ extracts features at resolution $r$, and $\text{conv}_\gamma$ converts features into modulation coefficients. Global controls are aligned with text embeddings via a condition encoder.
\(
h_g(c_g) \rightarrow K \text{ global tokens}
\)
Here, $c_g$ is the global condition, and $K$ is the number of global tokens. \textbf{IP-Adapter}~\cite{ye2023ip} enables pretrained text-to-image models to utilize image prompts effectively. It introduces a decoupled cross-attention mechanism, adding extra layers dedicated to image features while keeping the original text-focused layers intact. During training, these new layers learn to process image embeddings extracted by a CLIP encoder. At inference, the image and text features are processed separately then combined, improving controllability and fidelity of generated images. The core equation is:
\begin{equation}
    \hat{\epsilon}_\theta(x_t, c, t) = w\epsilon_\theta(x_t, c, t) + (1 - w)\epsilon_\theta(x_t, t)
    \enskip,
\end{equation}
where $\hat{\epsilon}_\theta(x_t, c, t)$ is the predicted noise, $w$ is the guidance scale adjusting the influence of condition $c$, $\epsilon_\theta(x_t, c, t)$ is the conditional noise prediction, and $\epsilon_\theta(x_t, t)$ is the unconditional prediction.

%% file: MLLM.tex
\begin{figure}[htbp]
    \centering
    \includegraphics[width=\columnwidth]{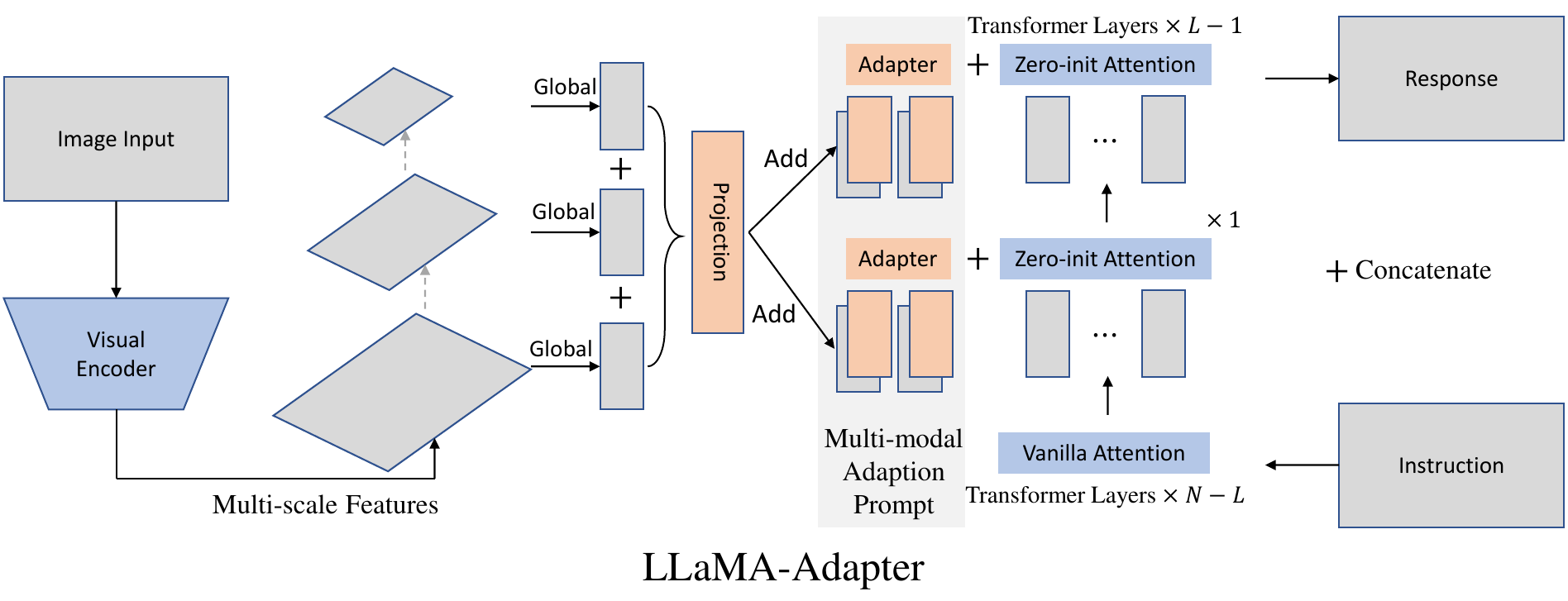} 
    \caption{Illustration of the principle of LLaMA-Adapter, which is a representative PEFT method in MLLM} 
    \label{fig:mllm_llama_adapter}
\end{figure}

The parameter-efficient fine-tuning of MLLM primarily focuses on the model connector.
It is because maintain consistency for both multimodal and textual data is challenging. As a consequence, a modal connector is serially connected right before the LLM, converting multimodal embeddings into understandable text prompt tokens for the LLM. Training the model connector on PEFT dataset bridges the gap between different modal data while ensuring consistency in the
input to the LLM. As a representative PEFT approach within the MLLM framework, the schematic diagram of LLaMA-Adapter~\cite{zhang2023llama} is illustrated in Figure~\ref{fig:mllm_llama_adapter}.

Generally, the parameter scale of the model connector will not be very large, much smaller than the prevalent LLMs. Therefore, full-parameter training instead of PEFT is more prevalent for model connector. Studies of the model connector primarily focus on the structural design, which will be dedicated to improving the training performance.   
A classic design of the modal connector involves employing a set of learnable query tokens to extract information in a query-based manner, a technique first introduced in
\textbf{BLIP-2}~\cite{li2023blip} and subsequently adopted by various
projects~\cite{dai2024instructblip}. These query-based approaches, reminiscent
of Q-Former-style methods, condense visual tokens into a smaller set of
representation vectors. In the meantime, some methods utilize an MLP-based
interface to bridge the modality gap. For instance, the \textbf{LLaVA} series
~\cite{liu2023improved, liu2024visual} employs one or two linear MLPs to
project visual tokens and align feature dimensions with word embeddings. 
In feature-level fusion, additional modules facilitate deep interaction and fusion
between text features and visual features. For example, \textbf{Flamingo}~\cite{alayrac2022flamingo} introduces extra cross-attention layers between the frozen Transformer layers of LLMs, enhancing language features with external visual cues. In addition, adapters and prompt embedding are also
applied to add learnable parameters to fill the gap, such as \textbf{LLaMA Adapter}~\cite{zhang2023llama} and \textbf{CogVLM}~\cite{wang2023cogvlm}.

\begin{figure}[htbp]
    \centering
    \includegraphics[width=\columnwidth]{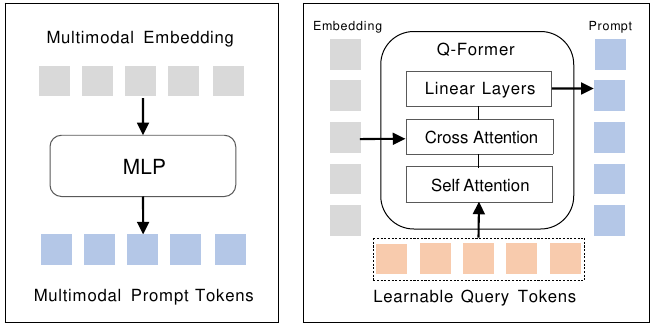} 
    \caption{Modal Connector Design: This figure shows two different mainstream design of the modal connector in MLLM. The first one, a simple MLP for converting the modal. The second one, layers with cross attention and query tokens for training.} 
    \label{fig: qformer} 
\end{figure}

Figure \ref{fig: qformer} illustrates the concrete structures of the two designs. The first one, pioneered by the LLaVA series, is characterized by
its simplicity. As highlighted by \cite{liu2024visual}, an MLP composed of basic
linear layers is adept at transforming multimodal embeddings into LLM prompt
tokens.

In contrast, the second paradigm, known as the \textbf{Q-Former}~\cite{li2023blip, dai2024instructblip}, introduces a transformer
neural network for modal information conversion. Unlike traditional approaches
of directly applying self-attention on input embeddings, Q-Former employs a set
of trainable query tokens. This approach bears resemblance to LLM PEFT methods
such as prefix-tuning and p-tuning, which incorporate external trainable
embedding tokens. However, the key distinction lies in how these methods handle
the tokens: prefix-tuning and p-tuning append them to the input text tokens to
form a comprehensive LLM input, while Q-Former accepts the query tokens as the
primary input.

From both the structural design and training intricacies, it becomes evident
that Q-Former is considerably more complicated compared to the MLP-based LLaVA.
However, this complexity comes with its advantages. A comprehensive transformer
network like Q-Former enables the execution of numerous pre-trained tasks,
facilitating explicit alignment between non-textual and textual modalities.
This, in turn, reduces the quality requirements on the multimodal data.
Nevertheless, LLaVA, as detailed by \cite{liu2024visual}, which incorporates
\textbf{GPT-4}~\cite{achiam2023gpt} as the LLM, reports a slight performance improvement over BLIP-2.
This is largely attributed to the inherent superiority of GPT-4 over BLIP-2's
Flan-T5 across various aspects. Specifically, GPT-4 possesses innate multimodal
reasoning capabilities, a feature lacking in Flan-T5. This observation underscores the fact that a comprehensive modal connector
design may not be necessary when the LLM itself possesses significant power and
capabilities. 

\begin{table*}[t]
  \centering
  \resizebox{\textwidth}{!}{
    \begin{tabular}{c|c|c|ccccccccccccc}
    \toprule
    \multirow{2}[2]{*}{\textbf{Task}} & \multirow{2}[2]{*}{\textbf{Model}} & \multirow{2}[2]{*}{\textbf{PEFT Method}} & \multicolumn{1}{c}{\multirow{2}[2]{*}{\textbf{\shortstack{\#TPs \\ (M)}}}} & \multicolumn{12}{c}{\textbf{Result}} \\
          &       &       &       & \multicolumn{1}{p{3.665em}}{\textbf{\shortstack{CIFAR \\ 100}}} & \multicolumn{1}{p{3.665em}}{\textbf{\shortstack{CIFAR \\ 10}}} & \textbf{Flowers} & \textbf{Food} & \textbf{EuroSAT} & \textbf{SUN} & \textbf{DMLab} & \textbf{SVHN} & \textbf{Pets} & \textbf{DTD} & \textbf{RESISC} & \textbf{CLEVR} \\
    \midrule
    
    \multicolumn{1}{c|}{\multirow{4}[2]{*}{\shortstack{Image \\ Classification}}} & \multirow{4}[2]{*}{CLIP} & FT    & 151.28 & 82.1  & 95.8  & 97.4  & 87.8  & 99    & 79    & 63.5  & 95.7  & 88.5  & 72.3  & 98.1  & 94.4 \\
          &       & VP    & 0.07  & 75.3  & 94.2  & 62    & 83.2  & 95.6  & 68.4  & 41.9  & 88.4  & 86.5  & 57.1  & 84.1  & 81.4 \\
          &       & VPT   & 0.064 & 76.6  & 95    & 76.2  & 84.7  & 94.6  & 69.3  & 48.4  & 86.1  & 92.1  & 61.6  & 84.3  & 58.6 \\
          &       & EVP   & 0.062 & 81.2  & 96.6  & 82.3  & 84.1  & 97.6  & 71    & 62.3  & 90.5  & 90    & 68.4  & 89.7  & 75.9 \\
    \midrule
          &       &       &       & \multicolumn{3}{c}{\textbf{Seg. }} & \multicolumn{3}{c}{\textbf{H.Part}} & \multicolumn{3}{c}{\textbf{Sal.}} & \multicolumn{3}{c}{\textbf{Normals.}} \\
    \midrule
    
    \multicolumn{1}{c|}{\multirow{16}[2]{*}{\shortstack{Dense \\ Prediction}}} & \multicolumn{1}{c|}{\multirow{16}[2]{*}{\shortstack{Swin \\ Transformer \\ -Tiny}}} & Single-task FT & 112.62 & \multicolumn{3}{c}{67.21} & \multicolumn{3}{c}{61.93} & \multicolumn{3}{c}{62.35} & \multicolumn{3}{c}{17.97} \\
          &       & Multi-task FT & 30.06 & \multicolumn{3}{c}{68.71} & \multicolumn{3}{c}{62.13} & \multicolumn{3}{c}{64.18} & \multicolumn{3}{c}{17.35} \\
          &       & Bitfit & 2.85  & \multicolumn{3}{c}{68.57} & \multicolumn{3}{c}{55.99} & \multicolumn{3}{c}{60.64} & \multicolumn{3}{c}{19.42} \\
          &       & Relative bias & 2.64  & \multicolumn{3}{c}{63.51} & \multicolumn{3}{c}{52.35} & \multicolumn{3}{c}{57.74} & \multicolumn{3}{c}{21.07} \\
          &       & VPT-shallow & 2.57  & \multicolumn{3}{c}{62.96} & \multicolumn{3}{c}{52.27} & \multicolumn{3}{c}{58.31} & \multicolumn{3}{c}{20.9} \\
          &       & VPT-deep & 3.43  & \multicolumn{3}{c}{64.35} & \multicolumn{3}{c}{55.24} & \multicolumn{3}{c}{58.15} & \multicolumn{3}{c}{21.07} \\
          &       & PHM layer & 3.14  & \multicolumn{3}{c}{68.55} & \multicolumn{3}{c}{56.28} & \multicolumn{3}{c}{60.35} & \multicolumn{3}{c}{19.23} \\
          &       & Compacter & 2.78  & \multicolumn{3}{c}{68.38} & \multicolumn{3}{c}{56.69} & \multicolumn{3}{c}{59.47} & \multicolumn{3}{c}{19.54} \\
          &       & Compacter++ & 2.66  & \multicolumn{3}{c}{67.26} & \multicolumn{3}{c}{55.69} & \multicolumn{3}{c}{59.47} & \multicolumn{3}{c}{19.54} \\
          &       & LoRA  & 2.87  & \multicolumn{3}{c}{67.26} & \multicolumn{3}{c}{55.69} & \multicolumn{3}{c}{59.47} & \multicolumn{3}{c}{19.54} \\
          &       & Adapter & 11.24 & \multicolumn{3}{c}{69.21} & \multicolumn{3}{c}{57.38} & \multicolumn{3}{c}{61.28} & \multicolumn{3}{c}{18.83} \\
          &       & \multicolumn{1}{p{7.465em}|}{\shortstack{Low-rank \\ adapter}} & 2.89  & \multicolumn{3}{c}{68.31} & \multicolumn{3}{c}{56.53} & \multicolumn{3}{c}{60.29} & \multicolumn{3}{c}{19.36} \\
          &       & \multicolumn{1}{p{7.465em}|}{\shortstack{Shared \\ Adapter}} & 4.74  & \multicolumn{3}{c}{70.21} & \multicolumn{3}{c}{59.15} & \multicolumn{3}{c}{62.29} & \multicolumn{3}{c}{19.26} \\
          &       & Hyperformer & 75.32 & \multicolumn{3}{c}{71.43} & \multicolumn{3}{c}{60.73} & \multicolumn{3}{c}{65.54} & \multicolumn{3}{c}{17.77} \\
          &       & Polyhistor & 8.96  & \multicolumn{3}{c}{70.87} & \multicolumn{3}{c}{59.54} & \multicolumn{3}{c}{65.47} & \multicolumn{3}{c}{17.47} \\
          &       & Polyhistor-Lite & 2.96  & \multicolumn{3}{c}{70.24} & \multicolumn{3}{c}{59.12} & \multicolumn{3}{c}{64.75} & \multicolumn{3}{c}{17.4} \\
    \midrule
          &       &       &       & \textbf{Food} & \textbf{SUN} & \multicolumn{1}{p{4.2em}}{\textbf{\shortstack{DF- \\ 20M}}} & \textbf{Caltech} & \multicolumn{1}{p{4.535em}}{\textbf{\shortstack{CUB- \\ Bird}}} & \textbf{ArtBench} & \multicolumn{1}{p{3.935em}}{\textbf{\shortstack{Oxford \\ Flowers}}} & \multicolumn{1}{p{4.265em}}{\textbf{\shortstack{Standard \\ Cars}}} & \multicolumn{1}{p{4.2em}}{\textbf{\shortstack{Average \\ FID}}} &       &       &  \\
    \midrule
    \multicolumn{1}{c|}{\multirow{8}[2]{*}{\shortstack{Generation \\ by Few-shot \\ Finetuning}}} & \multirow{8}[2]{*}{DiT-XL-2} & FT    & 673.8 & 10.46 & 7.96  & 17.26 & 35.25 & 5.68  & 25.31 & 21.05 & 9.79  & 16.59 &       &       &  \\
          &       & Adapt-Parallel & 4.28  & 13.67 & 11.47 & 22.38 & 35.76 & 7.73  & 38.43 & 21.24 & 10.73 & 20.17 &       &       &  \\
          &       & Adapt-Sequential & 4.28  & 11.93 & 10.68 & 19.01 & 34.17 & 7     & 35.04 & 21.36 & 10.45 & 18.7  &       &       &  \\
          &       & BitFit & 0.61  & 9.17  & 9.11  & 17.78 & 34.21 & 8.81  & 24.53 & 20.31 & 10.64 & 16.82 &       &       &  \\
          &       & VPT-Deep & 2.81  & 18.47 & 14.54 & 32.89 & 42.78 & 17.29 & 40.74 & 25.59 & 22.12 & 26.8  &       &       &  \\
          &       & LoRA-R8 & 1.15  & 33.75 & 32.33 & 120.25 & 86.05 & 56.03 & 80.99 & 164.13 & 76.24 & 81.31 &       &       &  \\
          &       & LoRA-R16 & 2.18  & 34.34 & 32.15 & 121.51 & 86.51 & 58.25 & 80.72 & 161.68 & 75.35 & 81.31 &       &       &  \\
          &       & DiffFit & 0.83  & 6.96  & 8.55  & 17.35 & 33.84 & 5.48  & 20.87 & 20.18 & 9.9   & 15.39 &       &       &  \\
    \midrule
          &       &       &       & \multicolumn{6}{c}{\textbf{CLIP-T}}           & \multicolumn{6}{c}{\textbf{CLIP-I}} \\
    \midrule
    \multicolumn{1}{c|}{\multirow{4}[2]{*}{\shortstack{Controllable \\ Generation}}} & \multicolumn{1}{c|}{\multirow{4}[2]{*}{\shortstack{CLIP \\ ViT-L/14}}} & \multicolumn{1}{p{7.465em}|}{\shortstack{Uni-ControlNet \\  (Global Control)}} & 47    & \multicolumn{6}{c}{0.51}                      & \multicolumn{6}{c}{0.74} \\
          &       & \multicolumn{1}{p{7.465em}|}{\shortstack{T2I-Adapter \\ (Style)}} & 39    & \multicolumn{6}{c}{0.49}                      & \multicolumn{6}{c}{0.65} \\
          &       & \multicolumn{1}{p{7.465em}|}{\shortstack{ControlNet \\ Shuffle}} & 361   & \multicolumn{6}{c}{0.42}                      & \multicolumn{6}{c}{0.62} \\
          &       & IP-Adapter & 22    & \multicolumn{6}{c}{0.59}                      & \multicolumn{6}{c}{0.83} \\
    \bottomrule
    \end{tabular}
    }
    \caption{Performance of PEFT methods in specific applications. All performance metrics are cited from prior published work \cite{wu2024evp, liu2022polyhistor, xie2023difffit, mou2023t2i}. Metrics vary by task: 1. Image Classification: 12 datasets with CLIP. 2. Dense Prediction: 4 datasets with Swim Transformer-Tiny. 3. Generation by Few-shot Finetuning: 9 datasets with DiT-XL-2. 4. Controllable Generation: 2 datasets with CLIP ViT-L/14.}
  \label{tab:performance_specific_applications}
\end{table*}

To further quantify the performance of different PEFT methods in specific applications, we present Table \ref{tab:performance_specific_applications}, which compares various methods based on key metrics such as accuracy and the number of trainable parameters across multiple benchmark tasks. Since existing literature does not provide detailed computational cost analysis, we use the number of trainable parameters as an approximate measure of computational efficiency, serving as a practical proxy for resource consumption across different PEFT methods. As shown in Table \ref{tab:performance_specific_applications}, compared to full fine-tuning, PEFT methods in specific applications significantly reduce the number of trainable parameters while maintaining competitive performance. These results highlight the advantage of PEFT methods in various applications, where they enable efficient adaptation of large models with lower computational and storage costs while preserving task-specific performance.

%% file: Future_Directions.tex
In this section, focusing on potential issues with existing PEFT techniques and aspects that have not received sufficient attention, we propose a series of possible research directions. These directions encompass \textbf{task}, \textbf{data}, \textbf{model}, \textbf{learning mechanisms}, and \textbf{fundamental flaws}.

\begin{enumerate}
    \item \textit{PEFT methods for multi-objective tasks:} 
    Current PEFT methods mainly focus on optimizing for single objectives (e.g., task accuracy), but real-world applications often require balancing multiple objectives (e.g., privacy, fairness, latency). For example, in healthcare, models must preserve patient privacy while maintaining diagnostic accuracy. Existing methods like LoRA or Adapters lack explicit mechanisms to handle such trade-offs. In recent work~\cite{yang2024multi}, the authors addressed the program repair task by incorporating a dual-objective optimization framework, wherein the two objectives were combined through linear weighting with manually predefined coefficients to formulate the model's loss function. Although this study presents a straightforward and effective approach to PEFT for multi-objective tasks, determining the optimal weighting coefficients remains non-trivial. This limitation highlights the need for developing more flexible and task-adaptive methodologies to enhance the robustness and generalizability of such approaches.
    \item \textit{PEFT methods in multimodal learning:} 
    Multimodal models (e.g., vision-language models) face unique challenges in aligning heterogeneous data streams (text, images, audio). Current PEFT methods (e.g., adapters) are primarily designed for unimodal LLMs, leading to suboptimal performance in tasks like visual question answering. Recent work on CLIP adaptations~\cite{zavras2024mind} highlights the need for modality-specific parameter-efficient tuning to bridge domain gaps. Multimodal learning has emerged as one of the most prominent research topics in contemporary machine learning. However, significant challenges persist in effectively integrating cross-modal information through parameter-efficient fine-tuning (PEFT) approaches, particularly in achieving optimal inter-modal alignment and representation learning while maintaining computational efficiency.

    \item \textit{Automated design of adapter modules:} 
    Adapter architectures (e.g., bottleneck layers) rely on manually tuned hyperparameters (e.g., dimension, placement), which limits scalability. Neural Architecture Search (NAS) techniques~\cite{xu2024automatic} could automate adapter design, optimizing for both parameter efficiency and task performance. However, the extensive design space of adapter modules significantly compromises the efficiency of NAS approaches. This limitation necessitates further investigation into more efficient and flexible automated design methodologies that can navigate the complex parameter space effectively while maintaining architectural optimality.
    
    \item \textit{Heuristic search strategies for hybrid PEFT methods:} 
    Hybrid methods (e.g., combining LoRA and adapters) often rely on trial-and-error combinations, lacking principled strategies. For example, in paper~\cite{chen2023parameter}, the authors, under a predefined design space, conduct numerous experiments to determine an ideal hybrid strategy. However, the optimal hybrid strategy may not be included within this artificially predefined design space. Therefore, introducing heuristic search strategies to find the best hybrid strategy is a promising direction for future research.
    
    \item \textit{Continual learning for PEFT methods:} 
    Deployed models must adapt to evolving data distributions (e.g., user preferences in chatbots). Traditional PEFT lacks mechanisms to prevent catastrophic forgetting. Current work~\cite{wei2024online} proposed a method for task-free online continual learning that dynamically adapts pretrained Vision Transformer models by adding new low-rank adaptation parameters when the loss surface plateaus, indicating data distribution shifts, and uses online weight regularization to mitigate catastrophic forgetting. The experimental results presented in this paper demonstrate significant performance improvements through the application of LoRA, establishing a valuable reference framework for investigating continual learning paradigms in other types of PEFT methodologies.
    
    \item \textit{Improving the calibration of fine-tuned LLMs:}To date, numerous PEFT approaches developed for the purpose of adeptly tailoring LLMs to downstream tasks have achieved notable advancements in computational and storage efficiency. Nonetheless, when subjected to fine-tuning on modest datasets, LLMs are often prone to overconfidence in their predictions~\cite{jiang2021can,tian2023just,achiam2023gpt}. This phenomenon is especially pernicious for decision-making processes within safety-critical applications or domains where data is scarce, such as medical diagnostics, financial services, and experimental design~\cite{singhal2023large,lee2024survey,huang2024crispr}. Hence, there exists an exigent demand for the formulation of strategies aimed at refining the calibration of fine-tuned LLMs, ensuring that their predictive outputs are not only dependable but also robust.

    \item \textit{Differential privacy for PEFT methods:} Different downstream tasks often involve varying levels of sensitve and personal data, which further emphasizes the need for privacy in large language model fine-tuning, particularly with PEFT methods. The integration of large language model fine-tuning and differential privacy holds significant promise for future research. However, existing differential privacy techniques, such as DP-SGD~\cite{abadi2016deep} and DP-AdamW~\cite{li2021large}, often result in limited performance and substantial computaitional cost. Therefore, future reasearch should focus on developing methods that preserve privacy while simultaneously optimizing performance and minimizing computational costs. Additionally, exploring scalable, privacy preserving methods tailored to PEFT methods is essential. These advancements will enable secure and efficient fine-tuning of large language models, ensuring robust privacy protections.

\end{enumerate}

%% file: Conclusions.tex
LLMs have garnered widespread attention due to their exceptional performance across a broad spectrum of natural language tasks, beginning with the release of ChatGPT in November 2022. These models have acquired the capability for general-purpose language understanding and generation by training billions of parameters on vast amounts of textual data, as predicted by scaling laws. Traditional full-parameter fine-tuning methods pose significant challenges when customizing these models for specific downstream tasks, particularly on hardware platforms with limited computational capabilities, due to their enormous parameter scale and computational demands. PEFT has emerged as an efficient method for adapting to various downstream tasks, minimizing the number of additional parameters introduced or the computational resources required, thereby enabling the fine-tuned model's performance to approach or even surpass that of full-parameter fine-tuning methods. This survey provides a systematic overview of the latest advancements in PEFT, encompassing introductions to classic pre-trained large models, classification and principle explanation of PEFT algorithms, applications of PEFT methods, and prospects for future research directions in PEFT. This survey not only offers readers a comprehensive and systematic organization of PEFT work but also inspires researchers in various fields to identify potential research directions in PEFT research, accelerating the research process of PEFT methods.

%% file: main.bbl
\begin{thebibliography}{100}
\providecommand{\url}[1]{#1}
\csname url@samestyle\endcsname
\providecommand{\newblock}{\relax}
\providecommand{\bibinfo}[2]{#2}
\providecommand{\BIBentrySTDinterwordspacing}{\spaceskip=0pt\relax}
\providecommand{\BIBentryALTinterwordstretchfactor}{4}
\providecommand{\BIBentryALTinterwordspacing}{\spaceskip=\fontdimen2\font plus
\BIBentryALTinterwordstretchfactor\fontdimen3\font minus \fontdimen4\font\relax}
\providecommand{\BIBforeignlanguage}[2]{{%
\expandafter\ifx\csname l@#1\endcsname\relax
\typeout{** WARNING: IEEEtran.bst: No hyphenation pattern has been}%
\typeout{** loaded for the language `#1'. Using the pattern for}%
\typeout{** the default language instead.}%
\else
\language=\csname l@#1\endcsname
\fi
#2}}
\providecommand{\BIBdecl}{\relax}
\BIBdecl

\bibitem{wu2023next}
S.~Wu, H.~Fei, L.~Qu, W.~Ji, and T.-S. Chua, ``Next-gpt: Any-to-any multimodal llm,'' \emph{arXiv preprint arXiv:2309.05519}, 2023.

\bibitem{li2024pre}
J.~Li, T.~Tang, W.~X. Zhao, J.-Y. Nie, and J.-R. Wen, ``Pre-trained language models for text generation: A survey,'' \emph{ACM Computing Surveys}, vol.~56, no.~9, pp. 1--39, 2024.

\bibitem{zhu2023multilingual}
W.~Zhu, H.~Liu, Q.~Dong, J.~Xu, S.~Huang, L.~Kong, J.~Chen, and L.~Li, ``Multilingual machine translation with large language models: Empirical results and analysis,'' \emph{arXiv preprint arXiv:2304.04675}, 2023.

\bibitem{wang2023document}
L.~Wang, C.~Lyu, T.~Ji, Z.~Zhang, D.~Yu, S.~Shi, and Z.~Tu, ``Document-level machine translation with large language models,'' \emph{arXiv preprint arXiv:2304.02210}, 2023.

\bibitem{zheng2023judging}
L.~Zheng, W.-L. Chiang, Y.~Sheng, S.~Zhuang, Z.~Wu, Y.~Zhuang, Z.~Lin, Z.~Li, D.~Li, E.~Xing \emph{et~al.}, ``Judging llm-as-a-judge with mt-bench and chatbot arena,'' \emph{Advances in Neural Information Processing Systems}, vol.~36, pp. 46\,595--46\,623, 2023.

\bibitem{kim2023chatgpt}
J.~K. Kim, M.~Chua, M.~Rickard, and A.~Lorenzo, ``Chatgpt and large language model (llm) chatbots: The current state of acceptability and a proposal for guidelines on utilization in academic medicine,'' \emph{Journal of Pediatric Urology}, vol.~19, no.~5, pp. 598--604, 2023.

\bibitem{dan2023educhat}
Y.~Dan, Z.~Lei, Y.~Gu, Y.~Li, J.~Yin, J.~Lin, L.~Ye, Z.~Tie, Y.~Zhou, Y.~Wang \emph{et~al.}, ``Educhat: A large-scale language model-based chatbot system for intelligent education,'' \emph{arXiv preprint arXiv:2308.02773}, 2023.

\bibitem{zhang2019pretraining}
H.~Zhang, J.~Xu, and J.~Wang, ``Pretraining-based natural language generation for text summarization,'' \emph{arXiv preprint arXiv:1902.09243}, 2019.

\bibitem{zhang2023enhancing}
B.~Zhang, H.~Yang, T.~Zhou, M.~Ali~Babar, and X.-Y. Liu, ``Enhancing financial sentiment analysis via retrieval augmented large language models,'' in \emph{Proceedings of the fourth ACM international conference on AI in finance}, 2023, pp. 349--356.

\bibitem{pan2024conv}
Z.~Pan, H.~Luo, M.~Li, and H.~Liu, ``Conv-coa: Improving open-domain question answering in large language models via conversational chain-of-action,'' \emph{arXiv preprint arXiv:2405.17822}, 2024.

\bibitem{yao2024survey}
Y.~Yao, J.~Duan, K.~Xu, Y.~Cai, Z.~Sun, and Y.~Zhang, ``A survey on large language model (llm) security and privacy: The good, the bad, and the ugly,'' \emph{High-Confidence Computing}, p. 100211, 2024.

\bibitem{huang2024survey}
K.~Huang, F.~Mo, H.~Li, Y.~Li, Y.~Zhang, W.~Yi, Y.~Mao, J.~Liu, Y.~Xu, J.~Xu \emph{et~al.}, ``A survey on large language models with multilingualism: Recent advances and new frontiers,'' \emph{arXiv preprint arXiv:2405.10936}, 2024.

\bibitem{huang2022towards}
J.~Huang and K.~C.-C. Chang, ``Towards reasoning in large language models: A survey,'' \emph{arXiv preprint arXiv:2212.10403}, 2022.

\bibitem{saparov2022language}
A.~Saparov and H.~He, ``Language models are greedy reasoners: A systematic formal analysis of chain-of-thought,'' \emph{arXiv preprint arXiv:2210.01240}, 2022.

\bibitem{houlsby2019parameter}
N.~Houlsby, A.~Giurgiu, S.~Jastrzebski, B.~Morrone, Q.~De~Laroussilhe, A.~Gesmundo, M.~Attariyan, and S.~Gelly, ``Parameter-efficient transfer learning for nlp,'' in \emph{International conference on machine learning}.\hskip 1em plus 0.5em minus 0.4em\relax PMLR, 2019, pp. 2790--2799.

\bibitem{li2021prefix}
X.~L. Li and P.~Liang, ``Prefix-tuning: Optimizing continuous prompts for generation,'' \emph{arXiv preprint arXiv:2101.00190}, 2021.

\bibitem{lester2021power}
B.~Lester, R.~Al-Rfou, and N.~Constant, ``The power of scale for parameter-efficient prompt tuning,'' \emph{arXiv preprint arXiv:2104.08691}, 2021.

\bibitem{chen2020recall}
S.~Chen, Y.~Hou, Y.~Cui, W.~Che, T.~Liu, and X.~Yu, ``Recall and learn: Fine-tuning deep pretrained language models with less forgetting,'' \emph{arXiv preprint arXiv:2004.12651}, 2020.

\bibitem{hu2021lora}
E.~J. Hu, Y.~Shen, P.~Wallis, Z.~Allen-Zhu, Y.~Li, S.~Wang, L.~Wang, and W.~Chen, ``Lora: Low-rank adaptation of large language models,'' \emph{arXiv preprint arXiv:2106.09685}, 2021.

\bibitem{zhang2023adalora}
Q.~Zhang, M.~Chen, A.~Bukharin, N.~Karampatziakis, P.~He, Y.~Cheng, W.~Chen, and T.~Zhao, ``Adalora: Adaptive budget allocation for parameter-efficient fine-tuning,'' \emph{arXiv preprint arXiv:2303.10512}, 2023.

\bibitem{liu-etal-2022-p}
X.~Liu, Y.~Zheng, Z.~Du, M.~Ding, Y.~Qian, Z.~Yang, and J.~Tang, ``Gpt understands, too,'' \emph{AI Open}, vol.~5, pp. 208--215, 2024.

\bibitem{zaken2021bitfit}
E.~B. Zaken, S.~Ravfogel, and Y.~Goldberg, ``Bitfit: Simple parameter-efficient fine-tuning for transformer-based masked language-models,'' \emph{arXiv preprint arXiv:2106.10199}, 2021.

\bibitem{ding2022delta}
N.~Ding, Y.~Qin, G.~Yang, F.~Wei, Z.~Yang, Y.~Su, S.~Hu, Y.~Chen, C.-M. Chan, W.~Chen \emph{et~al.}, ``Delta tuning: A comprehensive study of parameter efficient methods for pre-trained language models,'' \emph{arXiv preprint arXiv:2203.06904}, 2022.

\bibitem{lialin2023scaling}
V.~Lialin, V.~Deshpande, and A.~Rumshisky, ``Scaling down to scale up: A guide to parameter-efficient fine-tuning,'' \emph{arXiv preprint arXiv:2303.15647}, 2023.

\bibitem{xu2023parameter}
L.~Xu, H.~Xie, S.-Z.~J. Qin, X.~Tao, and F.~L. Wang, ``Parameter-efficient fine-tuning methods for pretrained language models: A critical review and assessment,'' \emph{arXiv preprint arXiv:2312.12148}, 2023.

\bibitem{xin2024parameter}
Y.~Xin, S.~Luo, H.~Zhou, J.~Du, X.~Liu, Y.~Fan, Q.~Li, and Y.~Du, ``Parameter-efficient fine-tuning for pre-trained vision models: A survey,'' \emph{arXiv preprint arXiv:2402.02242}, 2024.

\bibitem{han2024parameter}
Z.~Han, C.~Gao, J.~Liu, S.~Q. Zhang \emph{et~al.}, ``Parameter-efficient fine-tuning for large models: A comprehensive survey,'' \emph{arXiv preprint arXiv:2403.14608}, 2024.

\bibitem{vaswani2017attention}
A.~Vaswani, N.~Shazeer, N.~Parmar, J.~Uszkoreit, L.~Jones, A.~N. Gomez, {\L}.~Kaiser, and I.~Polosukhin, ``Attention is all you need,'' \emph{Advances in neural information processing systems}, vol.~30, 2017.

\bibitem{devlin2018bert}
J.~Devlin, M.-W. Chang, K.~Lee, and K.~Toutanova, ``Bert: Pre-training of deep bidirectional transformers for language understanding,'' \emph{arXiv preprint arXiv:1810.04805}, 2018.

\bibitem{achiam2023gpt}
J.~Achiam, S.~Adler, S.~Agarwal, L.~Ahmad, I.~Akkaya, F.~L. Aleman, D.~Almeida, J.~Altenschmidt, S.~Altman, S.~Anadkat \emph{et~al.}, ``Gpt-4 technical report,'' \emph{arXiv preprint arXiv:2303.08774}, 2023.

\bibitem{wei2022emergent}
J.~Wei, Y.~Tay, R.~Bommasani, C.~Raffel, B.~Zoph, S.~Borgeaud, D.~Yogatama, M.~Bosma, D.~Zhou, D.~Metzler \emph{et~al.}, ``Emergent abilities of large language models,'' \emph{arXiv preprint arXiv:2206.07682}, 2022.

\bibitem{sanh2021multitask}
V.~Sanh, A.~Webson, C.~Raffel, S.~H. Bach, L.~Sutawika, Z.~Alyafeai, A.~Chaffin, A.~Stiegler, T.~L. Scao, A.~Raja \emph{et~al.}, ``Multitask prompted training enables zero-shot task generalization,'' \emph{arXiv preprint arXiv:2110.08207}, 2021.

\bibitem{brown2020language}
T.~Brown, B.~Mann, N.~Ryder, M.~Subbiah, J.~D. Kaplan, P.~Dhariwal, A.~Neelakantan, P.~Shyam, G.~Sastry, A.~Askell \emph{et~al.}, ``Language models are few-shot learners,'' \emph{Advances in neural information processing systems}, vol.~33, pp. 1877--1901, 2020.

\bibitem{ouyang2022training}
L.~Ouyang, J.~Wu, X.~Jiang, D.~Almeida, C.~Wainwright, P.~Mishkin, C.~Zhang, S.~Agarwal, K.~Slama, A.~Ray \emph{et~al.}, ``Training language models to follow instructions with human feedback,'' \emph{Advances in neural information processing systems}, vol.~35, pp. 27\,730--27\,744, 2022.

\bibitem{chung2024scaling}
H.~W. Chung, L.~Hou, S.~Longpre, B.~Zoph, Y.~Tay, W.~Fedus, Y.~Li, X.~Wang, M.~Dehghani, S.~Brahma \emph{et~al.}, ``Scaling instruction-finetuned language models,'' \emph{Journal of Machine Learning Research}, vol.~25, no.~70, pp. 1--53, 2024.

\bibitem{kalla2023study}
D.~Kalla, N.~Smith, F.~Samaah, and S.~Kuraku, ``Study and analysis of chat gpt and its impact on different fields of study,'' \emph{International journal of innovative science and research technology}, vol.~8, no.~3, 2023.

\bibitem{team2023gemini}
G.~Team, R.~Anil, S.~Borgeaud, J.-B. Alayrac, J.~Yu, R.~Soricut, J.~Schalkwyk, A.~M. Dai, A.~Hauth, K.~Millican \emph{et~al.}, ``Gemini: a family of highly capable multimodal models,'' \emph{arXiv preprint arXiv:2312.11805}, 2023.

\bibitem{liu2024deepseekv3}
A.~Liu, B.~Feng, B.~Xue, B.~Wang, B.~Wu, C.~Lu, C.~Zhao, C.~Deng, C.~Zhang, C.~Ruan \emph{et~al.}, ``Deepseek-v3 technical report,'' \emph{arXiv preprint arXiv:2412.19437}, 2024.

\bibitem{wang2023power}
J.~Wang, ``The power of ai-assisted diagnosis,'' \emph{EAI Endorsed Transactions on e-Learning}, vol.~8, no.~4, 2023.

\bibitem{biswas2023role}
S.~S. Biswas, ``Role of chat gpt in public health,'' \emph{Annals of biomedical engineering}, vol.~51, no.~5, pp. 868--869, 2023.

\bibitem{li2024legalagentbench}
H.~Li, J.~Chen, J.~Yang, Q.~Ai, W.~Jia, Y.~Liu, K.~Lin, Y.~Wu, G.~Yuan, Y.~Hu \emph{et~al.}, ``Legalagentbench: Evaluating llm agents in legal domain,'' \emph{arXiv preprint arXiv:2412.17259}, 2024.

\bibitem{xing2024designing}
F.~Xing, ``Designing heterogeneous llm agents for financial sentiment analysis,'' \emph{ACM Transactions on Management Information Systems}, 2024.

\bibitem{ahn2024large}
J.~Ahn, R.~Verma, R.~Lou, D.~Liu, R.~Zhang, and W.~Yin, ``Large language models for mathematical reasoning: Progresses and challenges,'' \emph{arXiv preprint arXiv:2402.00157}, 2024.

\bibitem{wang2023scibench}
X.~Wang, Z.~Hu, P.~Lu, Y.~Zhu, J.~Zhang, S.~Subramaniam, A.~R. Loomba, S.~Zhang, Y.~Sun, and W.~Wang, ``Scibench: Evaluating college-level scientific problem-solving abilities of large language models,'' \emph{arXiv preprint arXiv:2307.10635}, 2023.

\bibitem{bommasani2021opportunities}
R.~Bommasani, D.~A. Hudson, E.~Adeli, R.~Altman, S.~Arora, S.~von Arx, M.~S. Bernstein, J.~Bohg, A.~Bosselut, E.~Brunskill \emph{et~al.}, ``On the opportunities and risks of foundation models,'' \emph{arXiv preprint arXiv:2108.07258}, 2021.

\bibitem{bender2021dangers}
E.~M. Bender, T.~Gebru, A.~McMillan-Major, and S.~Shmitchell, ``On the dangers of stochastic parrots: Can language models be too big?'' in \emph{Proceedings of the 2021 ACM conference on fairness, accountability, and transparency}, 2021, pp. 610--623.

\bibitem{lin2021truthfulqa}
S.~Lin, J.~Hilton, and O.~Evans, ``Truthfulqa: Measuring how models mimic human falsehoods,'' \emph{arXiv preprint arXiv:2109.07958}, 2021.

\bibitem{kaplan2020scaling}
J.~Kaplan, S.~McCandlish, T.~Henighan, T.~B. Brown, B.~Chess, R.~Child, S.~Gray, A.~Radford, J.~Wu, and D.~Amodei, ``Scaling laws for neural language models,'' \emph{arXiv preprint arXiv:2001.08361}, 2020.

\bibitem{hoffmann2022training}
J.~Hoffmann, S.~Borgeaud, A.~Mensch, E.~Buchatskaya, T.~Cai, E.~Rutherford, D.~d.~L. Casas, L.~A. Hendricks, J.~Welbl, A.~Clark \emph{et~al.}, ``Training compute-optimal large language models,'' \emph{arXiv preprint arXiv:2203.15556}, 2022.

\bibitem{radford2018improving}
A.~Radford, K.~Narasimhan, T.~Salimans, I.~Sutskever \emph{et~al.}, ``Improving language understanding by generative pre-training,'' \emph{OpenAI}, 2018.

\bibitem{radford2019language}
A.~Radford, J.~Wu, R.~Child, D.~Luan, D.~Amodei, I.~Sutskever \emph{et~al.}, ``Language models are unsupervised multitask learners,'' \emph{OpenAI blog}, vol.~1, no.~8, p.~9, 2019.

\bibitem{chen2021evaluating}
M.~Chen, J.~Tworek, H.~Jun, Q.~Yuan, H.~P. d.~O. Pinto, J.~Kaplan, H.~Edwards, Y.~Burda, N.~Joseph, G.~Brockman \emph{et~al.}, ``Evaluating large language models trained on code,'' \emph{arXiv preprint arXiv:2107.03374}, 2021.

\bibitem{nakano2021webgpt}
R.~Nakano, J.~Hilton, S.~Balaji, J.~Wu, L.~Ouyang, C.~Kim, C.~Hesse, S.~Jain, V.~Kosaraju, W.~Saunders \emph{et~al.}, ``Webgpt: Browser-assisted question-answering with human feedback,'' \emph{arXiv preprint arXiv:2112.09332}, 2021.

\bibitem{kojima2022large}
T.~Kojima, S.~S. Gu, M.~Reid, Y.~Matsuo, and Y.~Iwasawa, ``Large language models are zero-shot reasoners,'' \emph{Advances in neural information processing systems}, vol.~35, pp. 22\,199--22\,213, 2022.

\bibitem{touvron2023llama}
H.~Touvron, T.~Lavril, G.~Izacard, X.~Martinet, M.-A. Lachaux, T.~Lacroix, B.~Rozi{\`e}re, N.~Goyal, E.~Hambro, F.~Azhar \emph{et~al.}, ``Llama: Open and efficient foundation language models,'' \emph{arXiv preprint arXiv:2302.13971}, 2023.

\bibitem{vavekanand2024llama}
R.~Vavekanand and K.~Sam, ``Llama 3.1: An in-depth analysis of the next-generation large language model,'' 2024.

\bibitem{openai2024openaio1card}
A.~Jaech, A.~Kalai, A.~Lerer, A.~Richardson, A.~El-Kishky, A.~Low, A.~Helyar, A.~Madry, A.~Beutel, A.~Carney \emph{et~al.}, ``Openai o1 system card,'' \emph{arXiv preprint arXiv:2412.16720}, 2024.

\bibitem{bi2024deepseek}
X.~Bi, D.~Chen, G.~Chen, S.~Chen, D.~Dai, C.~Deng, H.~Ding, K.~Dong, Q.~Du, Z.~Fu \emph{et~al.}, ``Deepseek llm: Scaling open-source language models with longtermism,'' \emph{arXiv preprint arXiv:2401.02954}, 2024.

\bibitem{dai2024deepseekmoe}
D.~Dai, C.~Deng, C.~Zhao, R.~Xu, H.~Gao, D.~Chen, J.~Li, W.~Zeng, X.~Yu, Y.~Wu \emph{et~al.}, ``Deepseekmoe: Towards ultimate expert specialization in mixture-of-experts language models,'' \emph{arXiv preprint arXiv:2401.06066}, 2024.

\bibitem{liu2024deepseekv2}
A.~Liu, B.~Feng, B.~Wang, B.~Wang, B.~Liu, C.~Zhao, C.~Dengr, C.~Ruan, D.~Dai, D.~Guo \emph{et~al.}, ``Deepseek-v2: A strong, economical, and efficient mixture-of-experts language model,'' \emph{arXiv preprint arXiv:2405.04434}, 2024.

\bibitem{li2021chimera}
S.~Li and T.~Hoefler, ``Chimera: efficiently training large-scale neural networks with bidirectional pipelines,'' in \emph{Proceedings of the International Conference for High Performance Computing, Networking, Storage and Analysis}, 2021, pp. 1--14.

\bibitem{guo2025deepseek}
D.~Guo, D.~Yang, H.~Zhang, J.~Song, R.~Zhang, R.~Xu, Q.~Zhu, S.~Ma, P.~Wang, X.~Bi \emph{et~al.}, ``Deepseek-r1: Incentivizing reasoning capability in llms via reinforcement learning,'' \emph{arXiv preprint arXiv:2501.12948}, 2025.

\bibitem{shao2024deepseekmath}
Z.~Shao, P.~Wang, Q.~Zhu, R.~Xu, J.~Song, X.~Bi, H.~Zhang, M.~Zhang, Y.~Li, Y.~Wu \emph{et~al.}, ``Deepseekmath: Pushing the limits of mathematical reasoning in open language models,'' \emph{arXiv preprint arXiv:2402.03300}, 2024.

\bibitem{claude}
Anthropic, ``Claude,'' Available: \url{https://www.anthropic.com/claude}, Online, accessed: Feb. 11, 2025.

\bibitem{anil2023gemini}
R.~Anil, S.~Borgeaud, Y.~Wu, J.-B. Alayrac, J.~Yu, R.~Soricut, J.~Schalkwyk, A.~M. Dai, A.~Hauth, K.~Millican \emph{et~al.}, ``Gemini: A family of highly capable multimodal models,'' \emph{arXiv preprint arXiv:2312.11805}, vol.~1, 2023.

\bibitem{jiang2023mistral}
A.~Q. Jiang, A.~Sablayrolles, A.~Mensch, C.~Bamford, D.~S. Chaplot, D.~d.~l. Casas, F.~Bressand, G.~Lengyel, G.~Lample, L.~Saulnier \emph{et~al.}, ``Mistral 7b,'' \emph{arXiv preprint arXiv:2310.06825}, 2023.

\bibitem{riquelme2021scaling}
C.~Riquelme, J.~Puigcerver, B.~Mustafa, M.~Neumann, R.~Jenatton, A.~Susano~Pinto, D.~Keysers, and N.~Houlsby, ``Scaling vision with sparse mixture of experts,'' \emph{Advances in Neural Information Processing Systems}, vol.~34, pp. 8583--8595, 2021.

\bibitem{chowdhery2023palm}
A.~Chowdhery, S.~Narang, J.~Devlin, M.~Bosma, G.~Mishra, A.~Roberts, P.~Barham, H.~W. Chung, C.~Sutton, S.~Gehrmann \emph{et~al.}, ``Palm: Scaling language modeling with pathways,'' \emph{Journal of Machine Learning Research}, vol.~24, no. 240, pp. 1--113, 2023.

\bibitem{tay2022transcending}
Y.~Tay, J.~Wei, H.~W. Chung, V.~Q. Tran, D.~R. So, S.~Shakeri, X.~Garcia, H.~S. Zheng, J.~Rao, A.~Chowdhery \emph{et~al.}, ``Transcending scaling laws with 0.1\% extra compute,'' \emph{arXiv preprint arXiv:2210.11399}, 2022.

\bibitem{radford2021learning}
A.~Radford, J.~W. Kim, C.~Hallacy, A.~Ramesh, G.~Goh, S.~Agarwal, G.~Sastry, A.~Askell, P.~Mishkin, J.~Clark \emph{et~al.}, ``Learning transferable visual models from natural language supervision,'' in \emph{International conference on machine learning}.\hskip 1em plus 0.5em minus 0.4em\relax PMLR, 2021, pp. 8748--8763.

\bibitem{cherti2023reproducible}
M.~Cherti, R.~Beaumont, R.~Wightman, M.~Wortsman, G.~Ilharco, C.~Gordon, C.~Schuhmann, L.~Schmidt, and J.~Jitsev, ``Reproducible scaling laws for contrastive language-image learning,'' in \emph{Proceedings of the IEEE/CVF Conference on Computer Vision and Pattern Recognition}, 2023, pp. 2818--2829.

\bibitem{sun2023eva}
Q.~Sun, Y.~Fang, L.~Wu, X.~Wang, and Y.~Cao, ``Eva-clip: Improved training techniques for clip at scale,'' \emph{arXiv preprint arXiv:2303.15389}, 2023.

\bibitem{wang2022ofa}
P.~Wang, A.~Yang, R.~Men, J.~Lin, S.~Bai, Z.~Li, J.~Ma, C.~Zhou, J.~Zhou, and H.~Yang, ``Ofa: Unifying architectures, tasks, and modalities through a simple sequence-to-sequence learning framework,'' in \emph{International Conference on Machine Learning}.\hskip 1em plus 0.5em minus 0.4em\relax PMLR, 2022, pp. 23\,318--23\,340.

\bibitem{cho2021unifying}
J.~Cho, J.~Lei, H.~Tan, and M.~Bansal, ``Unifying vision-and-language tasks via text generation,'' in \emph{International Conference on Machine Learning}.\hskip 1em plus 0.5em minus 0.4em\relax PMLR, 2021, pp. 1931--1942.

\bibitem{zhai2022scaling}
X.~Zhai, A.~Kolesnikov, N.~Houlsby, and L.~Beyer, ``Scaling vision transformers,'' in \emph{Proceedings of the IEEE/CVF conference on computer vision and pattern recognition}, 2022, pp. 12\,104--12\,113.

\bibitem{raffel2020exploring}
C.~Raffel, N.~Shazeer, A.~Roberts, K.~Lee, S.~Narang, M.~Matena, Y.~Zhou, W.~Li, and P.~J. Liu, ``Exploring the limits of transfer learning with a unified text-to-text transformer,'' \emph{Journal of machine learning research}, vol.~21, no. 140, pp. 1--67, 2020.

\bibitem{liu2019multi}
X.~Liu, P.~He, W.~Chen, and J.~Gao, ``Multi-task deep neural networks for natural language understanding,'' \emph{arXiv preprint arXiv:1901.11504}, 2019.

\bibitem{aghajanyan2021muppet}
A.~Aghajanyan, A.~Gupta, A.~Shrivastava, X.~Chen, L.~Zettlemoyer, and S.~Gupta, ``Muppet: Massive multi-task representations with pre-finetuning,'' \emph{arXiv preprint arXiv:2101.11038}, 2021.

\bibitem{aribandi2021ext5}
V.~Aribandi, Y.~Tay, T.~Schuster, J.~Rao, H.~S. Zheng, S.~V. Mehta, H.~Zhuang, V.~Q. Tran, D.~Bahri, J.~Ni \emph{et~al.}, ``Ext5: Towards extreme multi-task scaling for transfer learning,'' \emph{arXiv preprint arXiv:2111.10952}, 2021.

\bibitem{khashabi2020unifiedqa}
D.~Khashabi, S.~Min, T.~Khot, A.~Sabharwal, O.~Tafjord, P.~Clark, and H.~Hajishirzi, ``Unifiedqa: Crossing format boundaries with a single qa system,'' \emph{arXiv preprint arXiv:2005.00700}, 2020.

\bibitem{mccann1806natural}
B.~McCann, N.~S. Keskar, C.~Xiong, and R.~Socher, ``The natural language decathlon: Multitask learning as question answering. arxiv 2018,'' \emph{arXiv preprint arXiv:1806.08730}, 2018.

\bibitem{keskar2019unifying}
N.~S. Keskar, B.~McCann, C.~Xiong, and R.~Socher, ``Unifying question answering, text classification, and regression via span extraction,'' \emph{arXiv preprint arXiv:1904.09286}, 2019.

\bibitem{mishra2021cross}
S.~Mishra, D.~Khashabi, C.~Baral, and H.~Hajishirzi, ``Cross-task generalization via natural language crowdsourcing instructions,'' \emph{arXiv preprint arXiv:2104.08773}, 2021.

\bibitem{wei2021finetuned}
J.~Wei, M.~Bosma, V.~Y. Zhao, K.~Guu, A.~W. Yu, B.~Lester, N.~Du, A.~M. Dai, and Q.~V. Le, ``Finetuned language models are zero-shot learners,'' \emph{arXiv preprint arXiv:2109.01652}, 2021.

\bibitem{bach2022promptsource}
S.~H. Bach, V.~Sanh, Z.-X. Yong, A.~Webson, C.~Raffel, N.~V. Nayak, A.~Sharma, T.~Kim, M.~S. Bari, T.~Fevry \emph{et~al.}, ``Promptsource: An integrated development environment and repository for natural language prompts,'' \emph{arXiv preprint arXiv:2202.01279}, 2022.

\bibitem{min2021metaicl}
S.~Min, M.~Lewis, L.~Zettlemoyer, and H.~Hajishirzi, ``Metaicl: Learning to learn in context,'' \emph{arXiv preprint arXiv:2110.15943}, 2021.

\bibitem{wang2022benchmarking}
Y.~Wang, S.~Mishra, P.~Alipoormolabashi, Y.~Kordi, A.~Mirzaei, A.~Arunkumar, A.~Ashok, A.~S. Dhanasekaran, A.~Naik, D.~Stap \emph{et~al.}, ``Benchmarking generalization via in-context instructions on 1,600+ language tasks,'' \emph{arXiv preprint arXiv:2204.07705}, vol.~2, 2022.

\bibitem{iyer2022opt}
S.~Iyer, X.~V. Lin, R.~Pasunuru, T.~Mihaylov, D.~Simig, P.~Yu, K.~Shuster, T.~Wang, Q.~Liu, P.~S. Koura \emph{et~al.}, ``Opt-iml: Scaling language model instruction meta learning through the lens of generalization,'' \emph{arXiv preprint arXiv:2212.12017}, 2022.

\bibitem{muennighoff2022crosslingual}
N.~Muennighoff, T.~Wang, L.~Sutawika, A.~Roberts, S.~Biderman, T.~L. Scao, M.~S. Bari, S.~Shen, Z.-X. Yong, H.~Schoelkopf \emph{et~al.}, ``Crosslingual generalization through multitask finetuning,'' \emph{arXiv preprint arXiv:2211.01786}, 2022.

\bibitem{chung2022scaling}
H.~W. Chung, L.~Hou, S.~Longpre, B.~Zoph, Y.~Tay, W.~Fedus, Y.~Li, X.~Wang, M.~Dehghani, S.~Brahma \emph{et~al.}, ``Scaling instruction-finetuned language models,'' \emph{arXiv preprint arXiv:2210.11416}, 2022.

\bibitem{wang2022self}
Y.~Wang, Y.~Kordi, S.~Mishra, A.~Liu, N.~A. Smith, D.~Khashabi, and H.~Hajishirzi, ``Self-instruct: Aligning language models with self-generated instructions,'' \emph{arXiv preprint arXiv:2212.10560}, 2022.

\bibitem{honovich2022unnatural}
O.~Honovich, T.~Scialom, O.~Levy, and T.~Schick, ``Unnatural instructions: Tuning language models with (almost) no human labor,'' \emph{arXiv preprint arXiv:2212.09689}, 2022.

\bibitem{ye2022guess}
S.~Ye, D.~Kim, J.~Jang, J.~Shin, and M.~Seo, ``Guess the instruction! making language models stronger zero-shot learners,'' \emph{arXiv preprint arXiv:2210.02969}, 2022.

\bibitem{gupta2022instructdial}
P.~Gupta, C.~Jiao, Y.-T. Yeh, S.~Mehri, M.~Eskenazi, and J.~P. Bigham, ``Instructdial: Improving zero and few-shot generalization in dialogue through instruction tuning,'' \emph{arXiv preprint arXiv:2205.12673}, 2022.

\bibitem{glaese2022improving}
A.~Glaese, N.~McAleese, M.~Trbacz, J.~Aslanides, V.~Firoiu, T.~Ewalds, M.~Rauh, L.~Weidinger, M.~Chadwick, P.~Thacker \emph{et~al.}, ``Improving alignment of dialogue agents via targeted human judgements,'' \emph{arXiv preprint arXiv:2209.14375}, 2022.

\bibitem{bai2022constitutional}
Y.~Bai, S.~Kadavath, S.~Kundu, A.~Askell, J.~Kernion, A.~Jones, A.~Chen, A.~Goldie, A.~Mirhoseini, C.~McKinnon \emph{et~al.}, ``Constitutional ai: Harmlessness from ai feedback,'' \emph{arXiv preprint arXiv:2212.08073}, 2022.

\bibitem{liu2022few}
H.~Liu, D.~Tam, M.~Muqeeth, J.~Mohta, T.~Huang, M.~Bansal, and C.~A. Raffel, ``Few-shot parameter-efficient fine-tuning is better and cheaper than in-context learning,'' \emph{Advances in Neural Information Processing Systems}, vol.~35, pp. 1950--1965, 2022.

\bibitem{vu2021spot}
T.~Vu, B.~Lester, N.~Constant, R.~Al-Rfou, and D.~Cer, ``Spot: Better frozen model adaptation through soft prompt transfer,'' \emph{arXiv preprint arXiv:2110.07904}, 2021.

\bibitem{singhal2023large}
K.~Singhal, S.~Azizi, T.~Tu, S.~S. Mahdavi, J.~Wei, H.~W. Chung, N.~Scales, A.~Tanwani, H.~Cole-Lewis, S.~Pfohl \emph{et~al.}, ``Large language models encode clinical knowledge,'' \emph{Nature}, vol. 620, no. 7972, pp. 172--180, 2023.

\bibitem{knox2008tamer}
W.~B. Knox and P.~Stone, ``Tamer: Training an agent manually via evaluative reinforcement,'' in \emph{2008 7th IEEE international conference on development and learning}.\hskip 1em plus 0.5em minus 0.4em\relax IEEE, 2008, pp. 292--297.

\bibitem{christiano2017deep}
P.~F. Christiano, J.~Leike, T.~Brown, M.~Martic, S.~Legg, and D.~Amodei, ``Deep reinforcement learning from human preferences,'' \emph{Advances in neural information processing systems}, vol.~30, 2017.

\bibitem{lee2023rlaif}
H.~Lee, S.~Phatale, H.~Mansoor, K.~Lu, T.~Mesnard, C.~Bishop, V.~Carbune, and A.~Rastogi, ``Rlaif: Scaling reinforcement learning from human feedback with ai feedback,'' \emph{arXiv preprint arXiv:2309.00267}, 2023.

\bibitem{sutton1995generalization}
R.~S. Sutton, ``Generalization in reinforcement learning: Successful examples using sparse coarse coding,'' \emph{Advances in neural information processing systems}, vol.~8, 1995.

\bibitem{fan2020theoretical}
J.~Fan, Z.~Wang, Y.~Xie, and Z.~Yang, ``A theoretical analysis of deep q-learning,'' in \emph{Learning for dynamics and control}.\hskip 1em plus 0.5em minus 0.4em\relax PMLR, 2020, pp. 486--489.

\bibitem{schulman2017proximal}
J.~Schulman, F.~Wolski, P.~Dhariwal, A.~Radford, and O.~Klimov, ``Proximal policy optimization algorithms,'' \emph{arXiv preprint arXiv:1707.06347}, 2017.

\bibitem{rafailov2024direct}
R.~Rafailov, A.~Sharma, E.~Mitchell, C.~D. Manning, S.~Ermon, and C.~Finn, ``Direct preference optimization: Your language model is secretly a reward model,'' \emph{Advances in Neural Information Processing Systems}, vol.~36, 2024.

\bibitem{yuan2021wudaocorpora}
S.~Yuan, H.~Zhao, Z.~Du, M.~Ding, X.~Liu, Y.~Cen, X.~Zou, Z.~Yang, and J.~Tang, ``Wudaocorpora: {A} super large-scale chinese corpora for pre-training language models,'' \emph{{AI} Open}, vol.~2, pp. 65--68, 2021.

\bibitem{Zhu_2015_ICCV}
Y.~Zhu, R.~Kiros, R.~S. Zemel, R.~Salakhutdinov, R.~Urtasun, A.~Torralba, and S.~Fidler, ``Aligning books and movies: Towards story-like visual explanations by watching movies and reading books,'' in \emph{{ICCV}}.\hskip 1em plus 0.5em minus 0.4em\relax {IEEE} Computer Society, 2015, pp. 19--27.

\bibitem{rae2019compressive}
J.~W. Rae, A.~Potapenko, S.~M. Jayakumar, C.~Hillier, and T.~P. Lillicrap, ``Compressive transformers for long-range sequence modelling,'' in \emph{{ICLR}}.\hskip 1em plus 0.5em minus 0.4em\relax OpenReview.net, 2020.

\bibitem{kocetkov2022stack}
D.~Kocetkov, R.~Li, L.~B. Allal, J.~Li, C.~Mou, Y.~Jernite, M.~Mitchell, C.~M. Ferrandis, S.~Hughes, T.~Wolf, D.~Bahdanau, L.~von Werra, and H.~de~Vries, ``The stack: 3 {TB} of permissively licensed source code,'' \emph{Trans. Mach. Learn. Res.}, vol. 2023, 2023.

\bibitem{Gokaslan2019OpenWeb}
A.~Gokaslan and V.~Cohen, ``Openwebtext corpus,'' \url{http://Skylion007.github.io/OpenWebTextCorpus}, 2019.

\bibitem{baumgartner2020pushshift}
J.~Baumgartner, S.~Zannettou, B.~Keegan, M.~Squire, and J.~Blackburn, ``The pushshift reddit dataset,'' in \emph{{ICWSM}}.\hskip 1em plus 0.5em minus 0.4em\relax {AAAI} Press, 2020, pp. 830--839.

\bibitem{gao2020pile}
L.~Gao, S.~Biderman, S.~Black, L.~Golding, T.~Hoppe, C.~Foster, J.~Phang, H.~He, A.~Thite, N.~Nabeshima \emph{et~al.}, ``The pile: An 800gb dataset of diverse text for language modeling,'' \emph{arXiv preprint arXiv:2101.00027}, 2020.

\bibitem{lo2019s2orc}
K.~Lo, L.~L. Wang, M.~Neumann, R.~Kinney, and D.~S. Weld, ``{S2ORC:} the semantic scholar open research corpus,'' in \emph{{ACL}}.\hskip 1em plus 0.5em minus 0.4em\relax Association for Computational Linguistics, 2020, pp. 4969--4983.

\bibitem{eisele2010multiun}
A.~Eisele and Y.~Chen, ``Multiun: A multilingual corpus from united nation documents.'' in \emph{LREC}, 2010.

\bibitem{duvsek2020evaluating}
O.~Du{\v{s}}ek, J.~Novikova, and V.~Rieser, ``Evaluating the state-of-the-art of end-to-end natural language generation: The e2e nlg challenge,'' \emph{Computer Speech \& Language}, vol.~59, pp. 123--156, 2020.

\bibitem{zhongSeq2SQL2017}
V.~Zhong, C.~Xiong, and R.~Socher, ``Seq2sql: Generating structured queries from natural language using reinforcement learning,'' \emph{arXiv preprint arXiv:1709.00103}, 2017.

\bibitem{web_nlg}
C.~Gardent, A.~Shimorina, S.~Narayan, and L.~Perez-Beltrachini, ``Creating training corpora for nlg micro-planning,'' in \emph{55th Annual Meeting of the Association for Computational Linguistics, ACL 2017}.\hskip 1em plus 0.5em minus 0.4em\relax Association for Computational Linguistics (ACL), 2017, pp. 179--188.

\bibitem{gliwa-etal-2019-samsum}
B.~Gliwa, I.~Mochol, M.~Biesek, and A.~Wawer, ``Samsum corpus: A human-annotated dialogue dataset for abstractive summarization,'' \emph{arXiv preprint arXiv:1911.12237}, 2019.

\bibitem{wang2023openchat}
G.~Wang, S.~Cheng, X.~Zhan, X.~Li, S.~Song, and Y.~Liu, ``Openchat: Advancing open-source language models with mixed-quality data,'' in \emph{{ICLR}}.\hskip 1em plus 0.5em minus 0.4em\relax OpenReview.net, 2024.

\bibitem{Narayan2018DontGM}
S.~Narayan, S.~B. Cohen, and M.~Lapata, ``Don't give me the details, just the summary! topic-aware convolutional neural networks for extreme summarization,'' in \emph{{EMNLP}}.\hskip 1em plus 0.5em minus 0.4em\relax Association for Computational Linguistics, 2018, pp. 1797--1807.

\bibitem{radev2020dart}
L.~Nan, D.~R. Radev, R.~Zhang, A.~Rau, A.~Sivaprasad, C.~Hsieh, X.~Tang, A.~Vyas, N.~Verma, P.~Krishna, Y.~Liu, N.~Irwanto, J.~Pan, F.~Rahman, A.~Zaidi, M.~Mutuma, Y.~Tarabar, A.~Gupta, T.~Yu, Y.~C. Tan, X.~V. Lin, C.~Xiong, R.~Socher, and N.~F. Rajani, ``{DART:} open-domain structured data record to text generation,'' in \emph{{NAACL-HLT}}.\hskip 1em plus 0.5em minus 0.4em\relax Association for Computational Linguistics, 2021, pp. 432--447.

\bibitem{bai2022training}
Y.~Bai, A.~Jones, K.~Ndousse, A.~Askell, A.~Chen, N.~DasSarma, D.~Drain, S.~Fort, D.~Ganguli, T.~Henighan \emph{et~al.}, ``Training a helpful and harmless assistant with reinforcement learning from human feedback,'' \emph{arXiv preprint arXiv:2204.05862}, 2022.

\bibitem{ji2024beavertails}
J.~Ji, M.~Liu, J.~Dai, X.~Pan, C.~Zhang, C.~Bian, B.~Chen, R.~Sun, Y.~Wang, and Y.~Yang, ``Beavertails: Towards improved safety alignment of llm via a human-preference dataset,'' \emph{Advances in Neural Information Processing Systems}, vol.~36, pp. 24\,678--24\,704, 2023.

\bibitem{yang2018hotpotqa}
Z.~Yang, P.~Qi, S.~Zhang, Y.~Bengio, W.~W. Cohen, R.~Salakhutdinov, and C.~D. Manning, ``Hotpotqa: {A} dataset for diverse, explainable multi-hop question answering,'' in \emph{{EMNLP}}.\hskip 1em plus 0.5em minus 0.4em\relax Association for Computational Linguistics, 2018, pp. 2369--2380.

\bibitem{pmlr-v162-ethayarajh22a}
K.~Ethayarajh, Y.~Choi, and S.~Swayamdipta, ``Understanding dataset difficulty with $\mathcal{V}$-usable information,'' in \emph{Proceedings of the 39th International Conference on Machine Learning}, ser. Proceedings of Machine Learning Research, K.~Chaudhuri, S.~Jegelka, L.~Song, C.~Szepesvari, G.~Niu, and S.~Sabato, Eds., vol. 162.\hskip 1em plus 0.5em minus 0.4em\relax PMLR, 17--23 Jul 2022, pp. 5988--6008.

\bibitem{hendrycks2021measuring}
D.~Hendrycks, C.~Burns, S.~Kadavath, A.~Arora, S.~Basart, E.~Tang \emph{et~al.}, ``Measuring mathematical problem solving with the math dataset,'' in \emph{Thirty-fifth Conference on Neural Information Processing Systems Datasets and Benchmarks Track (Round 2)}, 2021, pp. 1--11.

\bibitem{rein2024gpqa}
D.~Rein, B.~L. Hou, A.~C. Stickland, J.~Petty, R.~Y. Pang, J.~Dirani, J.~Michael, and S.~R. Bowman, ``Gpqa: A graduate-level google-proof q\&a benchmark,'' in \emph{First Conference on Language Modeling}, 2024.

\bibitem{sprague2024musrtestinglimitschainofthought}
Z.~Sprague, X.~Ye, K.~Bostrom, S.~Chaudhuri, and G.~Durrett, ``Musr: Testing the limits of chain-of-thought with multistep soft reasoning,'' \emph{arXiv preprint arXiv:2310.16049}, 2023.

\bibitem{hendryckstest2021}
D.~Hendrycks, C.~Burns, S.~Basart, A.~Zou, M.~Mazeika, D.~Song, and J.~Steinhardt, ``Measuring massive multitask language understanding,'' in \emph{{ICLR}}.\hskip 1em plus 0.5em minus 0.4em\relax OpenReview.net, 2021.

\bibitem{allenai:arc}
P.~Clark, I.~Cowhey, O.~Etzioni, T.~Khot, A.~Sabharwal, C.~Schoenick, and O.~Tafjord, ``Think you have solved question answering? try arc, the ai2 reasoning challenge,'' \emph{arXiv preprint arXiv:1803.05457v1}, 2018.

\bibitem{zellers2019hellaswag}
R.~Zellers, A.~Holtzman, Y.~Bisk, A.~Farhadi, and Y.~Choi, ``Hellaswag: Can a machine really finish your sentence?'' in \emph{Proceedings of the 57th Annual Meeting of the Association for Computational Linguistics}, 2019.

\bibitem{wang2018glue}
A.~Wang, A.~Singh, J.~Michael, F.~Hill, O.~Levy, and S.~R. Bowman, ``{GLUE:} {A} multi-task benchmark and analysis platform for natural language understanding,'' in \emph{{ICLR} (Poster)}.\hskip 1em plus 0.5em minus 0.4em\relax OpenReview.net, 2019.

\bibitem{wang2019superglue}
A.~Wang, Y.~Pruksachatkun, N.~Nangia, A.~Singh, J.~Michael, F.~Hill, O.~Levy, and S.~R. Bowman, ``Superglue: {A} stickier benchmark for general-purpose language understanding systems,'' in \emph{NeurIPS}, 2019, pp. 3261--3275.

\bibitem{cobbe2021gsm8k}
K.~Cobbe, V.~Kosaraju, M.~Bavarian, M.~Chen, H.~Jun, L.~Kaiser, M.~Plappert, J.~Tworek, J.~Hilton, R.~Nakano, C.~Hesse, and J.~Schulman, ``Training verifiers to solve math word problems,'' \emph{arXiv preprint arXiv:2110.14168}, 2021.

\bibitem{chen2023theoremqa}
W.~Chen, M.~Yin, M.~Ku, P.~Lu, Y.~Wan, X.~Ma, J.~Xu, X.~Wang, and T.~Xia, ``Theoremqa: {A} theorem-driven question answering dataset,'' in \emph{{EMNLP}}.\hskip 1em plus 0.5em minus 0.4em\relax Association for Computational Linguistics, 2023, pp. 7889--7901.

\bibitem{austin2021program}
J.~Austin, A.~Odena, M.~Nye, M.~Bosma, H.~Michalewski, D.~Dohan, E.~Jiang, C.~Cai, M.~Terry, Q.~Le \emph{et~al.}, ``Program synthesis with large language models,'' \emph{arXiv preprint arXiv:2108.07732}, 2021.

\bibitem{zhong2023agieval}
W.~Zhong, R.~Cui, Y.~Guo, Y.~Liang, S.~Lu, Y.~Wang, A.~Saied, W.~Chen, and N.~Duan, ``Agieval: {A} human-centric benchmark for evaluating foundation models,'' in \emph{{NAACL-HLT} (Findings)}.\hskip 1em plus 0.5em minus 0.4em\relax Association for Computational Linguistics, 2024, pp. 2299--2314.

\bibitem{Zhang2023EvaluatingTP}
X.~Zhang, C.~Li, Y.~Zong, Z.~Ying, L.~He, and X.~Qiu, ``Evaluating the performance of large language models on gaokao benchmark,'' \emph{arXiv preprint arXiv:2305.12474}, 2023.

\bibitem{suzgun2022challenging}
M.~Suzgun, N.~Scales, N.~Sch{\"{a}}rli, S.~Gehrmann, Y.~Tay, H.~W. Chung, A.~Chowdhery, Q.~V. Le, E.~H. Chi, D.~Zhou, and J.~Wei, ``Challenging big-bench tasks and whether chain-of-thought can solve them,'' in \emph{{ACL} (Findings)}.\hskip 1em plus 0.5em minus 0.4em\relax Association for Computational Linguistics, 2023, pp. 13\,003--13\,051.

\bibitem{lin2020exploring}
Z.~Lin, A.~Madotto, and P.~Fung, ``Exploring versatile generative language model via parameter-efficient transfer learning,'' \emph{arXiv preprint arXiv:2004.03829}, 2020.

\bibitem{ruckle2020adapterdrop}
A.~R{\"u}ckl{\'e}, G.~Geigle, M.~Glockner, T.~Beck, J.~Pfeiffer, N.~Reimers, and I.~Gurevych, ``Adapterdrop: On the efficiency of adapters in transformers,'' \emph{arXiv preprint arXiv:2010.11918}, 2020.

\bibitem{zhao2022tiny}
H.~Zhao, H.~Tan, and H.~Mei, ``Tiny-attention adapter: Contexts are more important than the number of parameters,'' \emph{arXiv preprint arXiv:2211.01979}, 2022.

\bibitem{he2021towards}
J.~He, C.~Zhou, X.~Ma, T.~Berg-Kirkpatrick, and G.~Neubig, ``Towards a unified view of parameter-efficient transfer learning,'' \emph{arXiv preprint arXiv:2110.04366}, 2021.

\bibitem{zhu2021counter}
Y.~Zhu, J.~Feng, C.~Zhao, M.~Wang, and L.~Li, ``Counter-interference adapter for multilingual machine translation,'' \emph{arXiv preprint arXiv:2104.08154}, 2021.

\bibitem{lei2024conditional}
T.~Lei, J.~Bai, S.~Brahma, J.~Ainslie, K.~Lee, Y.~Zhou, N.~Du, V.~Zhao, Y.~Wu, B.~Li \emph{et~al.}, ``Conditional adapters: Parameter-efficient transfer learning with fast inference,'' \emph{Advances in Neural Information Processing Systems}, vol.~36, 2024.

\bibitem{chen2023hadamard}
Y.~Chen, Q.~Fu, G.~Fan, L.~Du, J.-G. Lou, S.~Han, D.~Zhang, Z.~Li, and Y.~Xiao, ``Hadamard adapter: An extreme parameter-efficient adapter tuning method for pre-trained language models,'' in \emph{Proceedings of the 32nd ACM International Conference on Information and Knowledge Management}, 2023, pp. 276--285.

\bibitem{karimi2021compacter}
R.~Karimi~Mahabadi, J.~Henderson, and S.~Ruder, ``Compacter: Efficient low-rank hypercomplex adapter layers,'' \emph{Advances in Neural Information Processing Systems}, vol.~34, pp. 1022--1035, 2021.

\bibitem{he2022sparseadapter}
S.~He, L.~Ding, D.~Dong, M.~Zhang, and D.~Tao, ``Sparseadapter: An easy approach for improving the parameter-efficiency of adapters,'' \emph{arXiv preprint arXiv:2210.04284}, 2022.

\bibitem{liu2021gpt}
X.~Liu, Y.~Zheng, Z.~Du, M.~Ding, Y.~Qian, Z.~Yang, and J.~Tang, ``Gpt understands, too,'' \emph{arXiv preprint arXiv:2103.10385}, 2021.

\bibitem{liu2021p}
X.~Liu, K.~Ji, Y.~Fu, W.~L. Tam, Z.~Du, Z.~Yang, and J.~Tang, ``P-tuning v2: Prompt tuning can be comparable to fine-tuning universally across scales and tasks,'' \emph{arXiv preprint arXiv:2110.07602}, 2021.

\bibitem{choi2023smop}
J.-Y. Choi, J.~Kim, J.-H. Park, W.-L. Mok, and S.~Lee, ``Smop: Towards efficient and effective prompt tuning with sparse mixture-of-prompts,'' in \emph{The 2023 Conference on Empirical Methods in Natural Language Processing}, 2023.

\bibitem{zhang2023towards}
Z.-R. Zhang, C.~Tan, H.~Xu, C.~Wang, J.~Huang, and S.~Huang, ``Towards adaptive prefix tuning for parameter-efficient language model fine-tuning,'' \emph{arXiv preprint arXiv:2305.15212}, 2023.

\bibitem{wu2022idpg}
Z.~Wu, S.~Wang, J.~Gu, R.~Hou, Y.~Dong, V.~Vydiswaran, and H.~Ma, ``Idpg: An instance-dependent prompt generation method,'' \emph{arXiv preprint arXiv:2204.04497}, 2022.

\bibitem{liu2022late}
X.~Liu, T.~Sun, X.~Huang, and X.~Qiu, ``Late prompt tuning: A late prompt could be better than many prompts,'' \emph{arXiv preprint arXiv:2210.11292}, 2022.

\bibitem{zhu2023spt}
W.~Zhu and M.~Tan, ``Spt: Learning to selectively insert prompts for better prompt tuning,'' in \emph{The 2023 Conference on Empirical Methods in Natural Language Processing}, 2023.

\bibitem{wang2023aprompt}
Q.~Wang, Y.~Mao, J.~Wang, H.~Yu, S.~Nie, S.~Wang, F.~Feng, L.~Huang, X.~Quan, Z.~Xu \emph{et~al.}, ``Aprompt: Attention prompt tuning for efficient adaptation of pre-trained language models,'' in \emph{Proceedings of the 2023 Conference on Empirical Methods in Natural Language Processing}, 2023, pp. 9147--9160.

\bibitem{shi2023dept}
Z.~Shi and A.~Lipani, ``Dept: Decomposed prompt tuning for parameter-efficient fine-tuning,'' \emph{arXiv preprint arXiv:2309.05173}, 2023.

\bibitem{wu2024infoprompt}
J.~Wu, T.~Yu, R.~Wang, Z.~Song, R.~Zhang, H.~Zhao, C.~Lu, S.~Li, and R.~Henao, ``Infoprompt: Information-theoretic soft prompt tuning for natural language understanding,'' \emph{Advances in Neural Information Processing Systems}, vol.~36, 2024.

\bibitem{ma2022xprompt}
F.~Ma, C.~Zhang, L.~Ren, J.~Wang, Q.~Wang, W.~Wu, X.~Quan, and D.~Song, ``Xprompt: Exploring the extreme of prompt tuning,'' \emph{arXiv preprint arXiv:2210.04457}, 2022.

\bibitem{chen2023ptp}
L.~Chen, H.~Huang, and M.~Cheng, ``Ptp: Boosting stability and performance of prompt tuning with perturbation-based regularizer,'' \emph{arXiv preprint arXiv:2305.02423}, 2023.

\bibitem{zadouri2023pushing}
T.~Zadouri, A.~{\"U}st{\"u}n, A.~Ahmadian, B.~Ermi{\c{s}}, A.~Locatelli, and S.~Hooker, ``Pushing mixture of experts to the limit: Extremely parameter efficient moe for instruction tuning,'' \emph{arXiv preprint arXiv:2309.05444}, 2023.

\bibitem{lian2022scaling}
D.~Lian, D.~Zhou, J.~Feng, and X.~Wang, ``Scaling \& shifting your features: A new baseline for efficient model tuning,'' \emph{Advances in Neural Information Processing Systems}, vol.~35, pp. 109--123, 2022.

\bibitem{yang2023parameter}
X.~Yang, J.~Y. Huang, W.~Zhou, and M.~Chen, ``Parameter-efficient tuning with special token adaptation,'' in \emph{Proceedings of the 17th Conference of the European Chapter of the Association for Computational Linguistics}, 2023, pp. 865--872.

\bibitem{lu2023inference}
X.~Lu, F.~Brahman, P.~West, J.~Jung, K.~Chandu, A.~Ravichander, P.~Ammanabrolu, L.~Jiang, S.~Ramnath, N.~Dziri \emph{et~al.}, ``Inference-time policy adapters (ipa): Tailoring extreme-scale lms without fine-tuning,'' in \emph{Proceedings of the 2023 Conference on Empirical Methods in Natural Language Processing}, 2023, pp. 6863--6883.

\bibitem{sung2022lst}
Y.-L. Sung, J.~Cho, and M.~Bansal, ``Lst: Ladder side-tuning for parameter and memory efficient transfer learning,'' \emph{Advances in Neural Information Processing Systems}, vol.~35, pp. 12\,991--13\,005, 2022.

\bibitem{cao2022attention}
J.~Cao, C.~S. Prakash, and W.~Hamza, ``Attention fusion: a light yet efficient late fusion mechanism for task adaptation in nlu,'' in \emph{Findings of the Association for Computational Linguistics: NAACL 2022}, 2022, pp. 857--866.

\bibitem{aghajanyan2020intrinsic}
A.~Aghajanyan, L.~Zettlemoyer, and S.~Gupta, ``Intrinsic dimensionality explains the effectiveness of language model fine-tuning,'' \emph{arXiv preprint arXiv:2012.13255}, 2020.

\bibitem{edalati2022krona}
A.~Edalati, M.~Tahaei, I.~Kobyzev, V.~P. Nia, J.~J. Clark, and M.~Rezagholizadeh, ``Krona: Parameter efficient tuning with kronecker adapter,'' \emph{arXiv preprint arXiv:2212.10650}, 2022.

\bibitem{valipour2023dylora}
M.~Valipour, M.~Rezagholizadeh, I.~Kobyzev, and A.~Ghodsi, ``Dylora: Parameter-efficient tuning of pre-trained models using dynamic search-free low-rank adaptation,'' in \emph{Proceedings of the 17th Conference of the European Chapter of the Association for Computational Linguistics}, 2023, pp. 3274--3287.

\bibitem{zhang2023increlora}
F.~Zhang, L.~Li, J.~Chen, Z.~Jiang, B.~Wang, and Y.~Qian, ``Increlora: Incremental parameter allocation method for parameter-efficient fine-tuning,'' \emph{arXiv preprint arXiv:2308.12043}, 2023.

\bibitem{ding2023sparse}
N.~Ding, X.~Lv, Q.~Wang, Y.~Chen, B.~Zhou, Z.~Liu, and M.~Sun, ``Sparse low-rank adaptation of pre-trained language models,'' in \emph{Proceedings of the 2023 Conference on Empirical Methods in Natural Language Processing}, 2023, pp. 4133--4145.

\bibitem{hayou2024lora+}
S.~Hayou, N.~Ghosh, and B.~Yu, ``Lora+: Efficient low rank adaptation of large models,'' \emph{arXiv preprint arXiv:2402.12354}, 2024.

\bibitem{zhang2023lora}
L.~Zhang, L.~Zhang, S.~Shi, X.~Chu, and B.~Li, ``Lora-fa: Memory-efficient low-rank adaptation for large language models fine-tuning,'' \emph{arXiv preprint arXiv:2308.03303}, 2023.

\bibitem{liu2024dora}
S.-Y. Liu, C.-Y. Wang, H.~Yin, P.~Molchanov, Y.-C.~F. Wang, K.-T. Cheng, and M.-H. Chen, ``Dora: Weight-decomposed low-rank adaptation,'' \emph{arXiv preprint arXiv:2402.09353}, 2024.

\bibitem{yang2023bayesian}
A.~X. Yang, M.~Robeyns, X.~Wang, and L.~Aitchison, ``Bayesian low-rank adaptation for large language models,'' in \emph{The Twelfth International Conference on Learning Representations}, 2023.

\bibitem{chen2022empowering}
Y.~Chen, D.~Hazarika, M.~Namazifar, Y.~Liu, D.~Jin, and D.~Hakkani-Tur, ``Empowering parameter-efficient transfer learning by recognizing the kernel structure in self-attention,'' in \emph{Findings of the Association for Computational Linguistics: NAACL 2022}, 2022, pp. 1375--1388.

\bibitem{meng2024periodiclora}
X.~Meng, D.~Dai, W.~Luo, Z.~Yang, S.~Wu, X.~Wang, P.~Wang, Q.~Dong, L.~Chen, and Z.~Sui, ``Periodiclora: Breaking the low-rank bottleneck in lora optimization,'' \emph{arXiv preprint arXiv:2402.16141}, 2024.

\bibitem{tian2024hydralora}
C.~Tian, Z.~Shi, Z.~Guo, L.~Li, and C.~Xu, ``Hydralora: An asymmetric lora architecture for efficient fine-tuning,'' \emph{arXiv preprint arXiv:2404.19245}, 2024.

\bibitem{liu2024aflora}
Z.~Liu, S.~Kundu, A.~Li, J.~Wan, L.~Jiang, and P.~A. Beerel, ``Aflora: Adaptive freezing of low rank adaptation in parameter efficient fine-tuning of large models,'' \emph{arXiv preprint arXiv:2403.13269}, 2024.

\bibitem{wu2024lora}
Y.~Wu, Y.~Xiang, S.~Huo, Y.~Gong, and P.~Liang, ``Lora-sp: Streamlined partial parameter adaptation for resource-efficient fine-tuning of large language models,'' \emph{arXiv preprint arXiv:2403.08822}, 2024.

\bibitem{chen2024superlora}
X.~Chen, J.~Liu, Y.~Wang, M.~Brand, G.~Wang, T.~Koike-Akino \emph{et~al.}, ``Superlora: Parameter-efficient unified adaptation of multi-layer attention modules,'' \emph{arXiv preprint arXiv:2403.11887}, 2024.

\bibitem{guo2020parameter}
D.~Guo, A.~M. Rush, and Y.~Kim, ``Parameter-efficient transfer learning with diff pruning,'' \emph{arXiv preprint arXiv:2012.07463}, 2020.

\bibitem{lawton2023neural}
N.~Lawton, A.~Kumar, G.~Thattai, A.~Galstyan, and G.~V. Steeg, ``Neural architecture search for parameter-efficient fine-tuning of large pre-trained language models,'' \emph{arXiv preprint arXiv:2305.16597}, 2023.

\bibitem{liao2023parameter}
B.~Liao, Y.~Meng, and C.~Monz, ``Parameter-efficient fine-tuning without introducing new latency,'' \emph{arXiv preprint arXiv:2305.16742}, 2023.

\bibitem{sung2021training}
Y.-L. Sung, V.~Nair, and C.~A. Raffel, ``Training neural networks with fixed sparse masks,'' \emph{Advances in Neural Information Processing Systems}, vol.~34, pp. 24\,193--24\,205, 2021.

\bibitem{das2023unified}
S.~S.~S. Das, R.~H. Zhang, P.~Shi, W.~Yin, and R.~Zhang, ``Unified low-resource sequence labeling by sample-aware dynamic sparse finetuning,'' \emph{arXiv preprint arXiv:2311.03748}, 2023.

\bibitem{ansell2021composable}
A.~Ansell, E.~M. Ponti, A.~Korhonen, and I.~Vuli{\'c}, ``Composable sparse fine-tuning for cross-lingual transfer,'' \emph{arXiv preprint arXiv:2110.07560}, 2021.

\bibitem{fu2023effectiveness}
Z.~Fu, H.~Yang, A.~M.-C. So, W.~Lam, L.~Bing, and N.~Collier, ``On the effectiveness of parameter-efficient fine-tuning,'' in \emph{Proceedings of the AAAI Conference on Artificial Intelligence}, vol.~37, no.~11, 2023, pp. 12\,799--12\,807.

\bibitem{xu2021raise}
R.~Xu, F.~Luo, Z.~Zhang, C.~Tan, B.~Chang, S.~Huang, and F.~Huang, ``Raise a child in large language model: Towards effective and generalizable fine-tuning,'' \emph{arXiv preprint arXiv:2109.05687}, 2021.

\bibitem{zhao2020masking}
M.~Zhao, T.~Lin, F.~Mi, M.~Jaggi, and H.~Sch{\"u}tze, ``Masking as an efficient alternative to finetuning for pretrained language models,'' \emph{arXiv preprint arXiv:2004.12406}, 2020.

\bibitem{zhang2023pruning}
M.~Zhang, C.~Shen, Z.~Yang, L.~Ou, X.~Yu, B.~Zhuang \emph{et~al.}, ``Pruning meets low-rank parameter-efficient fine-tuning,'' \emph{arXiv preprint arXiv:2305.18403}, 2023.

\bibitem{vucetic2022efficient}
D.~Vucetic, M.~Tayaranian, M.~Ziaeefard, J.~J. Clark, B.~H. Meyer, and W.~J. Gross, ``Efficient fine-tuning of bert models on the edge,'' in \emph{2022 IEEE International Symposium on Circuits and Systems (ISCAS)}.\hskip 1em plus 0.5em minus 0.4em\relax IEEE, 2022, pp. 1838--1842.

\bibitem{gheini2021cross}
M.~Gheini, X.~Ren, and J.~May, ``Cross-attention is all you need: Adapting pretrained transformers for machine translation,'' \emph{arXiv preprint arXiv:2104.08771}, 2021.

\bibitem{he2023sensitivity}
H.~He, J.~Cai, J.~Zhang, D.~Tao, and B.~Zhuang, ``Sensitivity-aware visual parameter-efficient fine-tuning,'' in \emph{Proceedings of the IEEE/CVF International Conference on Computer Vision}, 2023, pp. 11\,825--11\,835.

\bibitem{mao2021unipelt}
Y.~Mao, L.~Mathias, R.~Hou, A.~Almahairi, H.~Ma, J.~Han, W.-t. Yih, and M.~Khabsa, ``Unipelt: A unified framework for parameter-efficient language model tuning,'' \emph{arXiv preprint arXiv:2110.07577}, 2021.

\bibitem{chen2023parameter}
J.~Chen, A.~Zhang, X.~Shi, M.~Li, A.~Smola, and D.~Yang, ``Parameter-efficient fine-tuning design spaces,'' \emph{arXiv preprint arXiv:2301.01821}, 2023.

\bibitem{hu2023llm}
Z.~Hu, L.~Wang, Y.~Lan, W.~Xu, E.-P. Lim, L.~Bing, X.~Xu, S.~Poria, and R.~K.-W. Lee, ``Llm-adapters: An adapter family for parameter-efficient fine-tuning of large language models,'' \emph{arXiv preprint arXiv:2304.01933}, 2023.

\bibitem{zhang2022neural}
Y.~Zhang, K.~Zhou, and Z.~Liu, ``Neural prompt search,'' \emph{arXiv preprint arXiv:2206.04673}, 2022.

\bibitem{zhou2024autopeft}
H.~Zhou, X.~Wan, I.~Vuli{\'c}, and A.~Korhonen, ``Autopeft: Automatic configuration search for parameter-efficient fine-tuning,'' \emph{Transactions of the Association for Computational Linguistics}, vol.~12, pp. 525--542, 2024.

\bibitem{hu2022sparse}
S.~Hu, Z.~Zhang, N.~Ding, Y.~Wang, Y.~Wang, Z.~Liu, and M.~Sun, ``Sparse structure search for delta tuning,'' \emph{Advances in Neural Information Processing Systems}, vol.~35, pp. 9853--9865, 2022.

\bibitem{zeng2023one}
G.~Zeng, P.~Zhang, and W.~Lu, ``One network, many masks: Towards more parameter-efficient transfer learning,'' \emph{arXiv preprint arXiv:2305.17682}, 2023.

\bibitem{jie2023revisiting}
S.~Jie, H.~Wang, and Z.-H. Deng, ``Revisiting the parameter efficiency of adapters from the perspective of precision redundancy,'' in \emph{Proceedings of the IEEE/CVF International Conference on Computer Vision}, 2023, pp. 17\,217--17\,226.

\bibitem{kim2024memory}
J.~Kim, J.~H. Lee, S.~Kim, J.~Park, K.~M. Yoo, S.~J. Kwon, and D.~Lee, ``Memory-efficient fine-tuning of compressed large language models via sub-4-bit integer quantization,'' \emph{Advances in Neural Information Processing Systems}, vol.~36, 2024.

\bibitem{dettmers2024qlora}
T.~Dettmers, A.~Pagnoni, A.~Holtzman, and L.~Zettlemoyer, ``Qlora: Efficient finetuning of quantized llms,'' \emph{Advances in Neural Information Processing Systems}, vol.~36, 2024.

\bibitem{guo2023lq}
H.~Guo, P.~Greengard, E.~P. Xing, and Y.~Kim, ``Lq-lora: Low-rank plus quantized matrix decomposition for efficient language model finetuning,'' \emph{arXiv preprint arXiv:2311.12023}, 2023.

\bibitem{xu2023qa}
Y.~Xu, L.~Xie, X.~Gu, X.~Chen, H.~Chang, H.~Zhang, Z.~Chen, X.~ZHANG, and Q.~Tian, ``Qa-lora: Quantization-aware low-rank adaptation of large language models,'' in \emph{The Twelfth International Conference on Learning Representations}, 2023.

\bibitem{rajabzadeh2024qdylora}
H.~Rajabzadeh, M.~Valipour, T.~Zhu, M.~Tahaei, H.~J. Kwon, A.~Ghodsi, B.~Chen, and M.~Rezagholizadeh, ``Qdylora: Quantized dynamic low-rank adaptation for efficient large language model tuning,'' \emph{arXiv preprint arXiv:2402.10462}, 2024.

\bibitem{li2023loftq}
Y.~Li, Y.~Yu, C.~Liang, N.~Karampatziakis, P.~He, W.~Chen, and T.~Zhao, ``Loftq: Lora-fine-tuning-aware quantization for large language models,'' in \emph{The Twelfth International Conference on Learning Representations}, 2023.

\bibitem{liu2024bitdelta}
J.~Liu, G.~Xiao, K.~Li, J.~D. Lee, S.~Han, T.~Dao, and T.~Cai, ``Bitdelta: Your fine-tune may only be worth one bit,'' \emph{arXiv preprint arXiv:2402.10193}, 2024.

\bibitem{pfeiffer2020adapterfusion}
J.~Pfeiffer, A.~Kamath, A.~R{\"u}ckl{\'e}, K.~Cho, and I.~Gurevych, ``Adapterfusion: Non-destructive task composition for transfer learning,'' \emph{arXiv preprint arXiv:2005.00247}, 2020.

\bibitem{wang2022adamix}
Y.~Wang, S.~Mukherjee, X.~Liu, J.~Gao, A.~H. Awadallah, and J.~Gao, ``Adamix: Mixture-of-adapter for parameter-efficient tuning of large language models,'' \emph{arXiv preprint arXiv:2205.12410}, vol.~1, no.~2, p.~4, 2022.

\bibitem{zhao2023prototype}
H.~Zhao, J.~Fu, and Z.~He, ``Prototype-based hyperadapter for sample-efficient multi-task tuning,'' \emph{arXiv preprint arXiv:2310.11670}, 2023.

\bibitem{chronopoulou2023adaptersoup}
A.~Chronopoulou, M.~E. Peters, A.~Fraser, and J.~Dodge, ``Adaptersoup: Weight averaging to improve generalization of pretrained language models,'' \emph{arXiv preprint arXiv:2302.07027}, 2023.

\bibitem{he2023mera}
S.~He, R.-Z. Fan, L.~Ding, L.~Shen, T.~Zhou, and D.~Tao, ``Mera: Merging pretrained adapters for few-shot learning,'' \emph{arXiv preprint arXiv:2308.15982}, 2023.

\bibitem{mahabadi2021parameter}
R.~K. Mahabadi, S.~Ruder, M.~Dehghani, and J.~Henderson, ``Parameter-efficient multi-task fine-tuning for transformers via shared hypernetworks,'' \emph{arXiv preprint arXiv:2106.04489}, 2021.

\bibitem{vu2022spot}
T.~Vu, B.~Lester, N.~Constant, R.~Al-Rfou, and D.~Cer, ``Spot: Better frozen model adaptation through soft prompt transfer,'' in \emph{Proceedings of the 60th Annual Meeting of the Association for Computational Linguistics (Volume 1: Long Papers)}, 2022, pp. 5039--5059.

\bibitem{asai2022attempt}
A.~Asai, M.~Salehi, M.~E. Peters, and H.~Hajishirzi, ``Attempt: Parameter-efficient multi-task tuning via attentional mixtures of soft prompts,'' in \emph{Proceedings of the 2022 Conference on Empirical Methods in Natural Language Processing}, 2022, pp. 6655--6672.

\bibitem{wang2022multitask}
Z.~Wang, R.~Panda, L.~Karlinsky, R.~Feris, H.~Sun, and Y.~Kim, ``Multitask prompt tuning enables parameter-efficient transfer learning,'' in \emph{The Eleventh International Conference on Learning Representations}, 2022.

\bibitem{qin2021exploring}
Y.~Qin, X.~Wang, Y.~Su, Y.~Lin, N.~Ding, J.~Yi, W.~Chen, Z.~Liu, J.~Li, L.~Hou \emph{et~al.}, ``Exploring universal intrinsic task subspace via prompt tuning,'' \emph{arXiv preprint arXiv:2110.07867}, 2021.

\bibitem{su2021transferability}
Y.~Su, X.~Wang, Y.~Qin, C.-M. Chan, Y.~Lin, H.~Wang, K.~Wen, Z.~Liu, P.~Li, J.~Li \emph{et~al.}, ``On transferability of prompt tuning for natural language processing,'' \emph{arXiv preprint arXiv:2111.06719}, 2021.

\bibitem{huang2023lorahub}
C.~Huang, Q.~Liu, B.~Y. Lin, T.~Pang, C.~Du, and M.~Lin, ``Lorahub: Efficient cross-task generalization via dynamic lora composition,'' \emph{arXiv preprint arXiv:2307.13269}, 2023.

\bibitem{liu2023moelora}
Q.~Liu, X.~Wu, X.~Zhao, Y.~Zhu, D.~Xu, F.~Tian, and Y.~Zheng, ``Moelora: An moe-based parameter efficient fine-tuning method for multi-task medical applications,'' \emph{arXiv preprint arXiv:2310.18339}, 2023.

\bibitem{tang2023parameter}
A.~Tang, L.~Shen, Y.~Luo, Y.~Zhan, H.~Hu, B.~Du, Y.~Chen, and D.~Tao, ``Parameter efficient multi-task model fusion with partial linearization,'' \emph{arXiv preprint arXiv:2310.04742}, 2023.

\bibitem{agiza2024mtlora}
A.~Agiza, M.~Neseem, and S.~Reda, ``Mtlora: A low-rank adaptation approach for efficient multi-task learning,'' \emph{arXiv preprint arXiv:2403.20320}, 2024.

\bibitem{bahng2022exploring}
H.~Bahng, A.~Jahanian, S.~Sankaranarayanan, and P.~Isola, ``Exploring visual prompts for adapting large-scale models,'' \emph{arXiv preprint arXiv:2203.17274}, 2022.

\bibitem{jia2022visual}
M.~Jia, L.~Tang, B.-C. Chen, C.~Cardie, S.~Belongie, B.~Hariharan, and S.-N. Lim, ``Visual prompt tuning,'' in \emph{European Conference on Computer Vision}.\hskip 1em plus 0.5em minus 0.4em\relax Springer, 2022, pp. 709--727.

\bibitem{jie2022convolutional}
S.~Jie and Z.-H. Deng, ``Convolutional bypasses are better vision transformer adapters,'' \emph{arXiv preprint arXiv:2207.07039}, 2022.

\bibitem{chen2022adaptformer}
S.~Chen, C.~Ge, Z.~Tong, J.~Wang, Y.~Song, J.~Wang, and P.~Luo, ``Adaptformer: Adapting vision transformers for scalable visual recognition,'' \emph{Advances in Neural Information Processing Systems}, vol.~35, pp. 16\,664--16\,678, 2022.

\bibitem{huang2023diversity}
Q.~Huang, X.~Dong, D.~Chen, W.~Zhang, F.~Wang, G.~Hua, and N.~Yu, ``Diversity-aware meta visual prompting,'' in \emph{Proceedings of the IEEE/CVF Conference on Computer Vision and Pattern Recognition}, 2023, pp. 10\,878--10\,887.

\bibitem{chen2023understanding}
A.~Chen, Y.~Yao, P.-Y. Chen, Y.~Zhang, and S.~Liu, ``Understanding and improving visual prompting: A label-mapping perspective,'' in \emph{Proceedings of the IEEE/CVF Conference on Computer Vision and Pattern Recognition}, 2023, pp. 19\,133--19\,143.

\bibitem{wu2024evp}
J.~Wu, X.~Li, C.~Wei, H.~Wang, A.~Yuille, Y.~Zhou, and C.~Xie, ``Unleashing the power of visual prompting at the pixel level,'' \emph{TMLR}, 2024.

\bibitem{tu2023visual}
C.-H. Tu, Z.~Mai, and W.-L. Chao, ``Visual query tuning: Towards effective usage of intermediate representations for parameter and memory efficient transfer learning,'' in \emph{Proceedings of the IEEE/CVF Conference on Computer Vision and Pattern Recognition}, 2023, pp. 7725--7735.

\bibitem{jie2023fact}
S.~Jie and Z.-H. Deng, ``Fact: Factor-tuning for lightweight adaptation on vision transformer,'' in \emph{Proceedings of the AAAI conference on artificial intelligence}, vol.~37, no.~1, 2023, pp. 1060--1068.

\bibitem{fu2024dtl}
M.~Fu, K.~Zhu, and J.~Wu, ``Dtl: Disentangled transfer learning for visual recognition,'' in \emph{Proceedings of the AAAI Conference on Artificial Intelligence}, vol.~38, no.~11, 2024, pp. 12\,082--12\,090.

\bibitem{wang2024lion}
H.~Wang, J.~Chang, Y.~Zhai, X.~Luo, J.~Sun, Z.~Lin, and Q.~Tian, ``Lion: Implicit vision prompt tuning,'' in \emph{Proceedings of the AAAI Conference on Artificial Intelligence}, vol.~38, no.~6, 2024, pp. 5372--5380.

\bibitem{liu2022polyhistor}
Y.-C. Liu, C.-Y. Ma, J.~Tian, Z.~He, and Z.~Kira, ``Polyhistor: Parameter-efficient multi-task adaptation for dense vision tasks,'' \emph{Advances in Neural Information Processing Systems}, vol.~35, pp. 36\,889--36\,901, 2022.

\bibitem{chen2022vision}
Z.~Chen, Y.~Duan, W.~Wang, J.~He, T.~Lu, J.~Dai, and Y.~Qiao, ``Vision transformer adapter for dense predictions,'' \emph{arXiv preprint arXiv:2205.08534}, 2022.

\bibitem{xu2023side}
M.~Xu, Z.~Zhang, F.~Wei, H.~Hu, and X.~Bai, ``Side adapter network for open-vocabulary semantic segmentation,'' in \emph{Proceedings of the IEEE/CVF Conference on Computer Vision and Pattern Recognition}, 2023, pp. 2945--2954.

\bibitem{yin20231}
D.~Yin, Y.~Yang, Z.~Wang, H.~Yu, K.~Wei, and X.~Sun, ``1\% vs 100\%: Parameter-efficient low rank adapter for dense predictions,'' in \emph{Proceedings of the IEEE/CVF Conference on Computer Vision and Pattern Recognition}, 2023, pp. 20\,116--20\,126.

\bibitem{ruiz2023dreambooth}
N.~Ruiz, Y.~Li, V.~Jampani, Y.~Pritch, M.~Rubinstein, and K.~Aberman, ``Dreambooth: Fine tuning text-to-image diffusion models for subject-driven generation,'' in \emph{Proceedings of the IEEE/CVF Conference on Computer Vision and Pattern Recognition}, 2023, pp. 22\,500--22\,510.

\bibitem{gal2022image}
R.~Gal, Y.~Alaluf, Y.~Atzmon, O.~Patashnik, A.~H. Bermano, G.~Chechik, and D.~Cohen-Or, ``An image is worth one word: Personalizing text-to-image generation using textual inversion,'' \emph{arXiv preprint arXiv:2208.01618}, 2022.

\bibitem{dong2022dreamartist}
Z.~Dong, P.~Wei, and L.~Lin, ``Dreamartist: Towards controllable one-shot text-to-image generation via positive-negative prompt-tuning,'' \emph{arXiv preprint arXiv:2211.11337}, 2022.

\bibitem{voynov2023p+}
A.~Voynov, Q.~Chu, D.~Cohen-Or, and K.~Aberman, ``$ p+ $: Extended textual conditioning in text-to-image generation,'' \emph{arXiv preprint arXiv:2303.09522}, 2023.

\bibitem{xie2023difffit}
E.~Xie, L.~Yao, H.~Shi, Z.~Liu, D.~Zhou, Z.~Liu, J.~Li, and Z.~Li, ``Difffit: Unlocking transferability of large diffusion models via simple parameter-efficient fine-tuning,'' in \emph{Proceedings of the IEEE/CVF International Conference on Computer Vision}, 2023, pp. 4230--4239.

\bibitem{liu2023cones}
Z.~Liu, R.~Feng, K.~Zhu, Y.~Zhang, K.~Zheng, Y.~Liu, D.~Zhao, J.~Zhou, and Y.~Cao, ``Cones: Concept neurons in diffusion models for customized generation,'' \emph{arXiv preprint arXiv:2303.05125}, 2023.

\bibitem{han2023svdiff}
L.~Han, Y.~Li, H.~Zhang, P.~Milanfar, D.~Metaxas, and F.~Yang, ``Svdiff: Compact parameter space for diffusion fine-tuning,'' in \emph{Proceedings of the IEEE/CVF International Conference on Computer Vision}, 2023, pp. 7323--7334.

\bibitem{yeh2023navigating}
S.-Y. Yeh, Y.-G. Hsieh, Z.~Gao, B.~B. Yang, G.~Oh, and Y.~Gong, ``Navigating text-to-image customization: From lycoris fine-tuning to model evaluation,'' \emph{arXiv preprint arXiv:2309.14859}, 2023.

\bibitem{marjit2024diffusekrona}
S.~Marjit, H.~Singh, N.~Mathur, S.~Paul, C.-M. Yu, and P.-Y. Chen, ``Diffusekrona: A parameter efficient fine-tuning method for personalized diffusion model,'' \emph{arXiv preprint arXiv:2402.17412}, 2024.

\bibitem{qiu2024controlling}
Z.~Qiu, W.~Liu, H.~Feng, Y.~Xue, Y.~Feng, Z.~Liu, D.~Zhang, A.~Weller, and B.~Sch{\"o}lkopf, ``Controlling text-to-image diffusion by orthogonal finetuning,'' \emph{Advances in Neural Information Processing Systems}, vol.~36, 2024.

\bibitem{voynov2023sketch}
A.~Voynov, K.~Aberman, and D.~Cohen-Or, ``Sketch-guided text-to-image diffusion models,'' in \emph{ACM SIGGRAPH 2023 Conference Proceedings}, 2023, pp. 1--11.

\bibitem{zhang2023adding}
L.~Zhang, A.~Rao, and M.~Agrawala, ``Adding conditional control to text-to-image diffusion models,'' in \emph{Proceedings of the IEEE/CVF International Conference on Computer Vision}, 2023, pp. 3836--3847.

\bibitem{mou2023t2i}
C.~Mou, X.~Wang, L.~Xie, Y.~Wu, J.~Zhang, Z.~Qi, Y.~Shan, and X.~Qie, ``T2i-adapter: Learning adapters to dig out more controllable ability for text-to-image diffusion models,'' \emph{arXiv preprint arXiv:2302.08453}, 2023.

\bibitem{zhao2024uni}
S.~Zhao, D.~Chen, Y.-C. Chen, J.~Bao, S.~Hao, L.~Yuan, and K.-Y.~K. Wong, ``Uni-controlnet: All-in-one control to text-to-image diffusion models,'' \emph{Advances in Neural Information Processing Systems}, vol.~36, 2024.

\bibitem{ye2023ip}
H.~Ye, J.~Zhang, S.~Liu, X.~Han, and W.~Yang, ``Ip-adapter: Text compatible image prompt adapter for text-to-image diffusion models,'' \emph{arXiv preprint arXiv:2308.06721}, 2023.

\bibitem{li2023blip}
J.~Li, D.~Li, S.~Savarese, and S.~Hoi, ``Blip-2: Bootstrapping language-image pre-training with frozen image encoders and large language models,'' \emph{arXiv preprint arXiv:2301.12597}, 2023.

\bibitem{liu2023improved}
H.~Liu, C.~Li, Y.~Li, and Y.~J. Lee, ``Improved baselines with visual instruction tuning,'' \emph{arXiv preprint arXiv:2310.03744}, 2023.

\bibitem{liu2024visual}
H.~Liu, C.~Li, Q.~Wu, and Y.~J. Lee, ``Visual instruction tuning,'' \emph{Advances in neural information processing systems}, vol.~36, 2024.

\bibitem{alayrac2022flamingo}
J.-B. Alayrac, J.~Donahue, P.~Luc, A.~Miech, I.~Barr, Y.~Hasson, K.~Lenc, A.~Mensch, K.~Millican, M.~Reynolds \emph{et~al.}, ``Flamingo: a visual language model for few-shot learning,'' \emph{Advances in neural information processing systems}, vol.~35, pp. 23\,716--23\,736, 2022.

\bibitem{zhang2023llama}
R.~Zhang, J.~Han, C.~Liu, P.~Gao, A.~Zhou, X.~Hu, S.~Yan, P.~Lu, H.~Li, and Y.~Qiao, ``Llama-adapter: Efficient fine-tuning of language models with zero-init attention,'' \emph{arXiv preprint arXiv:2303.16199}, 2023.

\bibitem{wang2023cogvlm}
W.~Wang, Q.~Lv, W.~Yu, W.~Hong, J.~Qi, Y.~Wang, J.~Ji, Z.~Yang, L.~Zhao, X.~Song \emph{et~al.}, ``Cogvlm: Visual expert for pretrained language models,'' \emph{arXiv preprint arXiv:2311.03079}, 2023.

\bibitem{dai2024instructblip}
W.~Dai, J.~Li, D.~Li, A.~M.~H. Tiong, J.~Zhao, W.~Wang, B.~Li, P.~N. Fung, and S.~Hoi, ``Instructblip: Towards general-purpose vision-language models with instruction tuning,'' \emph{Advances in Neural Information Processing Systems}, vol.~36, 2024.

\bibitem{yang2024multi}
B.~Yang, H.~Tian, J.~Ren, H.~Zhang, J.~Klein, T.~F. Bissyand{\'e}, C.~L. Goues, and S.~Jin, ``Multi-objective fine-tuning for enhanced program repair with llms,'' \emph{arXiv preprint arXiv:2404.12636}, 2024.

\bibitem{zavras2024mind}
A.~Zavras, D.~Michail, B.~Demir, and I.~Papoutsis, ``Mind the modality gap: Towards a remote sensing vision-language model via cross-modal alignment,'' \emph{arXiv preprint arXiv:2402.09816}, 2024.

\bibitem{xu2024automatic}
S.~Xu and X.~Wen, ``Automatic design of adapter architectures for enhanced parameter-efficient fine-tuning,'' in \emph{ICASSP 2024-2024 IEEE International Conference on Acoustics, Speech and Signal Processing (ICASSP)}.\hskip 1em plus 0.5em minus 0.4em\relax IEEE, 2024, pp. 12\,536--12\,540.

\bibitem{wei2024online}
X.~Wei, G.~Li, and R.~Marculescu, ``Online-lora: Task-free online continual learning via low rank adaptation,'' \emph{arXiv preprint arXiv:2411.05663}, 2024.

\bibitem{jiang2021can}
Z.~Jiang, J.~Araki, H.~Ding, and G.~Neubig, ``How can we know when language models know? on the calibration of language models for question answering,'' \emph{Transactions of the Association for Computational Linguistics}, vol.~9, pp. 962--977, 2021.

\bibitem{tian2023just}
K.~Tian, E.~Mitchell, A.~Zhou, A.~Sharma, R.~Rafailov, H.~Yao, C.~Finn, and C.~D. Manning, ``Just ask for calibration: Strategies for eliciting calibrated confidence scores from language models fine-tuned with human feedback,'' \emph{arXiv preprint arXiv:2305.14975}, 2023.

\bibitem{lee2024survey}
J.~Lee, N.~Stevens, S.~C. Han, and M.~Song, ``A survey of large language models in finance (finllms),'' \emph{arXiv preprint arXiv:2402.02315}, 2024.

\bibitem{huang2024crispr}
K.~Huang, Y.~Qu, H.~Cousins, W.~A. Johnson, D.~Yin, M.~Shah, D.~Zhou, R.~Altman, M.~Wang, and L.~Cong, ``Crispr-gpt: An llm agent for automated design of gene-editing experiments,'' \emph{arXiv preprint arXiv:2404.18021}, 2024.

\bibitem{abadi2016deep}
M.~Abadi, A.~Chu, I.~Goodfellow, H.~B. McMahan, I.~Mironov, K.~Talwar, and L.~Zhang, ``Deep learning with differential privacy,'' in \emph{Proceedings of the 2016 ACM SIGSAC conference on computer and communications security}, 2016, pp. 308--318.

\bibitem{li2021large}
X.~Li, F.~Tramer, P.~Liang, and T.~Hashimoto, ``Large language models can be strong differentially private learners,'' \emph{arXiv preprint arXiv:2110.05679}, 2021.

\end{thebibliography}
